\documentclass[10pt,twoside]{article}
\usepackage{amsmath}
\usepackage[utf8]{inputenc}
\usepackage[T1]{fontenc}
\usepackage{url}
\usepackage{dsfont}
\usepackage{graphicx}
\usepackage{yhmath}
\usepackage[table]{xcolor}
\usepackage{mathrsfs} % permet d'avoir le D cursive
\usepackage{amssymb}
\usepackage{url}
\usepackage{makecell}
\usepackage{arydshln}
\usepackage{multirow}
\usepackage{arydshln}
\usepackage{mathtools}
\usepackage{stmaryrd}  % pour intervalle entiers
 \usepackage{amsthm,amssymb}
\usepackage{cancel}

\input epsf
\usepackage[colorlinks=true, linkcolor=blue, urlcolor=magenta, citecolor=blue]{hyperref}
\usepackage{cleveref}

\DeclareMathOperator*{\argmin}{argmin}

\newtheorem{definition}{Definition}[section]
\newtheorem{theorem}{Theorem}[section]
\newtheorem{proposition}{Proposition}[section]
\newtheorem{corollary}{Corollary }[section]
\newtheorem{lemma}{Lemma }[section]

\newtheorem{remark}{Remark}[section]

\newcommand{\norm}[1]{\left\|#1\right\|}
\numberwithin{equation}{section} % Pour rénitialiser le compteur des équations apres chaque section

\newcommand{\E}{\ensuremath{\mathbb{E}}}
\newcommand{\R}{\ensuremath{\mathbb{R}}}
\newcommand{\Z}{\ensuremath{\mathbb{Z}}}

\newcommand{\N}{\ensuremath{\mathbb{N}}}

\definecolor{grisclair}{gray}{0.9}
\font\dsrom=dsrom10 scaled 1200
\def \ind{\textrm{\dsrom{1}}}
\newcommand{\mk}{ { \mathcal{K}} }
\newcommand{\mx}{ { \mathcal{X}} }
\newcommand{\my}{ { \mathcal{Y}} }
\newcommand{\sgn}{ { \text{sgn} } }
\newcommand{\mf}{ { \mathcal{F}} }

\usepackage{array}
\usepackage{float}
\usepackage{multirow}
\usepackage{lscape}
\usepackage{layout}

\usepackage{tikz,bm} 
\usepackage{circuitikz}

\begin{document}

\title{\bf Adaptive deep nonparametric regression from dependent data under covariate  shift}
 \maketitle \vspace{-1.0cm}

\begin{center}
      William Kengne
     % \footnote{Developed within the ANR BREAKRISK: ANR-17-CE26-0001-01 and the  CY Initiative of Excellence (grant "Investissements d'Avenir" ANR-16-IDEX-0008), Project "EcoDep" PSI-AAP2020-0000000013 } 
   and 
     Ehud Mossa Ockegna 
     %\footnote{Supported by the MME-DII center of excellence (ANR-11-LABEX-0023-01)} 
 \end{center}

  \begin{center}
  { \it 
  Université Jean Monnet, ICJ UMR5208, CNRS, Ecole Centrale de Lyon, INSA Lyon, Universite
Claude Bernard Lyon 1, 42000 Saint-Étienne, France\\
  E-mail:   william.kengne@univ-st-etienne.fr  ; ehud.mossa.ockegna@univ-st-etienne.fr\\
  }
\end{center}

 \pagestyle{myheadings}
%\markboth{}{Kengne and Ockegna}

\markboth{Adaptive deep nonparametric regression under covariate  shift}{Kengne and Mossa}

\textbf{Abstract:} 
 Covariate shift often occurs because, in many real applications, the source and the target observations may be generated from different distributions. In this case, the standard metric under the source distribution is not appropriate. 
 This paper considers deep neural network estimators for nonparametric quantile and Huber regression under covariate  shift and from dependent observations.
 We deal with a generalized Bernstein-type inequality that is satisfied by many classical models, including i.i.d. observations, $\phi$-mixing, strong mixing, and $\mathcal{C}$-mixing processes.
To perform the covariate shift phenomenon, we propose a sparse-penalized deep neural network (SPDNN) estimator that takes into account the discrepancy between the source and target distributions of the data.
When the density ratio (between the source and target distributions of the covariate) is unknown, a two steps pre-training procedure is carried out: the first step is devoted to the construction of a least squares SPDNN estimator of the density ratio; which is used in the second step to perform a pre-training reweighted SPDNN estimator of the regression function.
For both the quantile and the Huber regression, non-asymptotic error bounds of the proposed SPDNN estimators are established in the class of H\"older smooth  functions. 
These estimators can adaptively attain (up to a logarithmic factor) the minimax optimal convergence rate from i.i.d. data as well as from several classical time series models.


\section{Introduction}

Many results on the theoretical guarantees of deep neural network (DNN) algorithms have been established; see, for example, \cite{bauer2019deep}, \cite{schmidt2020nonparametric}, \cite{kim2021fast}, \cite{padilla2022quantile}, \cite{ohn2022nonconvex}, \cite{jiao2023deep} for some contributions from independent data, and \cite{ma2022theoretical}, \cite{kohler2023rate}, \cite{alquier2025minimax}, \cite{kurisu2025adaptive}, \cite{kengne2025deep} for some works from dependent observations.
But most of these results in the literature on DNN estimators assume that the source (training) and the target (testing) datasets have the same distribution.
Such a condition fails in many real-world applications.
For example, in healthcare applications, the distribution mismatch may due to the inconsistency in medical equipment and the scanning protocols, see for instance \cite{taori2020measuring}, \cite{koh2021wilds}, \cite{guan2021domain}.
Another example is natural language processing with vocabulary evolution in the study population.
 The phenomenon of distribution shift appears when there exists a divergence between the distributions of the training and the testing datasets. Covariate shift is a particular case of distribution shift, and it appears when the marginal distributions of the covariates of the training and the target data are different.
This question has recently been addressed for nonparametric regression from independent and identically distributed (i.i.d.) observations, see for instance \cite{feng2023towards}, \cite {ma2023optimally}, \cite {feng2024deep}.

\medskip

  In this contribution, we consider a stationary and ergodic process $\{Z_t = (X_t, Y_t), t \in \mathbb{Z}\}$ with values in $\mathcal{Z} = \mathcal{X} \times \mathcal{Y} \subset \mathbb{R}^d \times \mathbb{R}$ (with $d \in \mathbb{N}$), and satisfying:
\begin{equation}
\label{regression_model}
    Y_t = h_0(X_t) + \xi_t , \;  t \in \mathbb{Z},
\end{equation} 
where $\mx, \my$ are respectively the feature and the response spaces, $h_0 : \mathcal{X} \rightarrow \mathbb{R}$ is the unknown (measurable) regression function, and $\left( \xi_t \right)_{t \in \mathbb{Z}}$ is a centered iid error process.
We focus on the estimation of $h_0$ from the Huber and the quantile regression in the covariate shift framework, and based on a training sample $\mathbb{D}:=\{ (X_1,Y_1),\cdots,(X_n,Y_n) \}$.
%
%with a loss function $\ell_\tau$.

 \medskip

 Assume that the training and testing data are trajectories of the stationary process $(X_t, Y_t)_{t \in \Z}$ where the distribution of $(X_0, Y_0)$ is $\mathbb{P}_{X_0,Y_0}$ and $\mathbb{Q}_{X_0,Y_0}$ respectively for the training and the testing observations.
 As pointed out above, in many real applications, $\mathbb{P}_{X_0,Y_0}$ is not equal to $\mathbb{Q}_{X_0,Y_0}$. 
 For the distribution $\mathbb{P}_{X_0,Y_0}$, denote by $\mathbb{P}_{X_0}$ and $\mathbb{P}_{Y_0|X_0}$, respectively, the marginal distribution of $X_0$ and the conditional distribution of $Y_0$ given $X_0$. The same notations are set for $\mathbb{Q}_{X_0,Y_0}$.
 In the setting considered here, $\mathbb{P}_{X_0}$ and $\mathbb{Q}_{X_0}$ are different, but $\mathbb{P}_{Y_0|X_0} = \mathbb{Q}_{Y_0|X_0}$, which implies that the conditional distribution of $\xi_0$ given $X_0$ remains invariant whatever $X_0$ is generated from $\mathbb{P}_{X_0}$ or $\mathbb{Q}_{X_0}$. This is known as the covariate shift phenomenon.
 When it occurs, it is appropriate to assess the performance of estimators from the $L_2(\mathbb{Q}_{X_0})$ norm, instead of the $L_2(\mathbb{P}_{X_0})$ norm as in the standard approach.
 In the sequel, denote by $p_{X_0}$ and $q_{X_0}$, the probability density functions of the distributions $\mathbb{P}_{X_0}$ and $\mathbb{Q}_{X_0}$ respectively. 
 The density ratio, given by:
 \begin{equation}\label{def_dens_ratio}
  r (x) := \dfrac{q_{X_0}(x)}{p_{X_0}(x)}, \text{ for all } x \in \mx,
 \end{equation}
 can be used to quantify the discrepancy between $\mathbb{P}_{X_0}$ and $\mathbb{Q}_{X_0}$.

 \medskip

 The literature on deep nonparametric regression under covariate shift is not vast.
 This question has recently been addressed by \cite{duan2022convergence}, \cite{feng2024deep}, \cite{wang2026deep}.
 \cite{duan2022convergence} propose ReLU (rectiﬁed linear unit) DNN estimators for nonparametric regression with a sub-Gaussian noise under covariate shift. 
 The convergence rates of the proposed estimators are established on the class of H\"older smooth functions.
 \cite{feng2024deep} consider this question for nonparametric quantile regression.
 Based on the ReLU DNN, they construct a reweighted and pre-training reweighted estimators, and derive their convergence rates on the class of H\"older functions.
 The pre-training reweighted DNN estimator is performed from a two-stage procedure when the density ratio is unknown. 
Let us highlight the great merit of this approach, which allows these authors to obtain estimators that can achieve (up to a logarithmic factor) the minimax optimal rate when the density ratio is unbounded and has moments of any order; whereas the optimality of the standard unweighted estimator is not ensured.
Nonparametric regression with repeated measurements under covariate shift has been considered by \cite{wang2026deep}.
They proposed ReLU DNN estimators for the density ratio (when it is unknown) and the regression function, and established their convergence rates on the class of H\"older functions. 
These estimators can attain (up to a logarithmic factor) the minimax optimal rate.
However, the results in  \cite{duan2022convergence}, \cite{feng2024deep}, \cite{wang2026deep} are established under independent assumptions of the observations.

 \medskip

 This work is an extension of \cite{feng2024deep} in three directions: (i) dependent observations, (ii) Huber and quantile nonparametric regression, and (iii) sparse-penalized regularization which leads to adaptive estimators.
\begin{enumerate}
    \item[(i)] \textbf{Dependent framework}. We unify the theory for several class dependence structures, including $\phi$-mixing, strong mixing, and $\mathcal{C}$-mixing processes. This is handled through a generalized Bernstein-type inequality, which is based on the so-called the ``effective number of observations'' $\varphi(n)$.
    So, the convergence rates of the proposed estimators are expressed in terms of this value $\varphi(n)$.
    \item[(ii)] \textbf{Huber and quantile regression}. Besides the quantile regression as in \cite{feng2024deep}, we also consider the Huber regression. This induces other issues, including the calibration of the parameter of the Huber loss function, see for instance, Lemma \ref{lemma_decomposition_estimator} below.
    This loss function is widely used for robust estimation of the presence of outliers or for problems with heavy-tailed data.
    Moreover, for deep nonparametric regression with heavy-tailed errors, the Huber estimator can achieve the minimax optimal rate, whereas the least square estimators cannot attain this rate, except when the error is sub-Gaussian, see \cite{fan2024noise}.
    \item[(iii)] \textbf{Sparse-penalized regularization}. We perform the sparse-penalization approach. The regularization is carried out through a penalty term, and the estimators proposed can adaptively attain the minimax convergence rate.
    \cite{feng2024deep} focus on the sparsity constrained regularization, based on a parameter $\mathcal{S}$. 
    Besides the difficulties in its practical implementation (due to the discrete nature of such a constraint), the appropriate level $\mathcal{S}$ needed to reach the optimal rate depends on the smoothness of the target function, which is unknown. Such estimators are nonadaptive. 
\end{enumerate}

\medskip

We consider the Huber and the quantile nonparametric regression under covariate shift from dependent observations in the model (\ref{regression_model}) and in addition to the issues pointed above, our main contributions include:
\begin{enumerate}
    \item We propose sparse-penalized DNN (SPDNN) unweighted and reweighted estimators of $h_0$, and establish their non-asymptotic $L_2(\mathbb{Q}_{X_0})$ error bounds on the class of H\"older smooth functions.
    For this purpose, some preliminary bounds for both the Huber and the quantile regression have been established, see, for instance Lemma \ref{lemma_decomposition_estimator}.
     In the case of unbounded density ratio, a truncated version is considered in the reweighted SPDNN estimator.
    When the density ratio is uniformly bounded, these estimators can adaptively achieve (up to a logarithmic factor) the minimax optimal rate of order $\mathcal{O}(n^{-\frac{2s}{2s+d}})$ (where $n$ is the sample size and $s>0$ the smoothness parameter of the regression function) in many classical models.
    The reweighted estimator remains minimax optimal even if the density ratio is not uniformly bounded, but has moments of any order.  
    \item  When the density ratio is unknown, a two-step pre-training procedure based on the approach of \cite{feng2024deep} is performed.
    The first step focuses on a least squares SPDNN estimator of the density ratio $r$ and its $L_2(\mathbb{P}_{X_0})$ error bound on the H\"older class.
    This is used in the second step to construct a pre-training reweighted SPDNN estimator of the regression function $h_0$.
    This two steps pre-training procedure is carried out in both the bounded and unbounded density ratio. In this latter case, a truncated estimator of $r$ is considered.
    Non-asymptotic $L_2(\mathbb{Q}_{X_0})$ error bounds of these pre-training reweighted SPDNN estimators of $h_0$ are derived.
    For several dependence structures, including $\phi$-mixing, strong mixing, and $\mathcal{C}$-mixing, these estimators can adaptively attain the minimax optimal rate.
\end{enumerate}
Figure \ref{Etimators} summarizes our main contributions with an illustration of the estimators proposed.

{

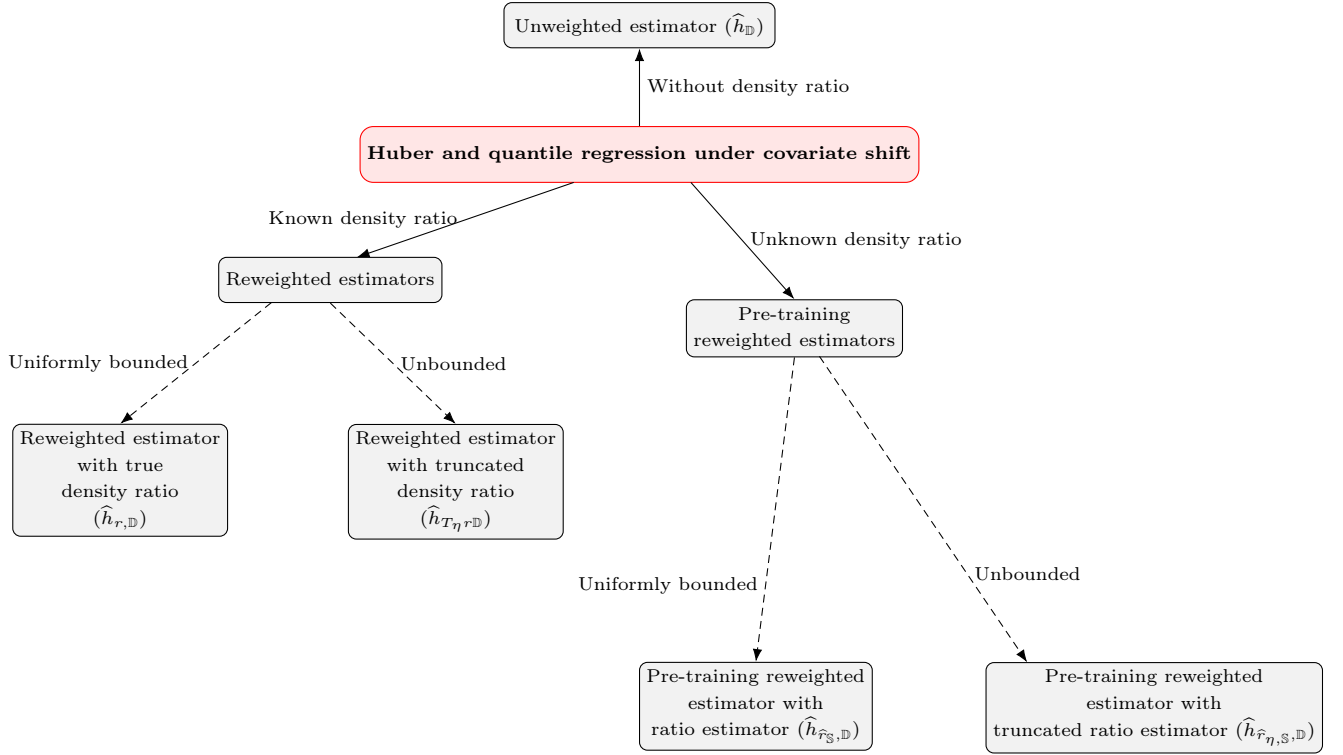
\begin{figure}[h]
\scriptsize
    \centering
\begin{tikzpicture}
  \node[draw=red, fill=red!10, rectangle, rounded corners=5pt, minimum width=58pt, minimum height=21pt] (node1) at (0.87,2.97) {\textbf{Huber and quantile regression under covariate shift}};
  \node[draw, fill=black!5, rectangle, rounded corners=3pt, minimum width=102pt, minimum height=17pt] (node5) at (0.87,4.68) {Unweighted estimator ($\widehat{h}_{\mathbb{D}}$)};
  \draw[arrows=-Latex] (node1.north) -- (node5) node[midway, right] {Without density ratio};
  \draw[arrows=-Latex] (0,2.6) -- (-2.87,1.61) node[midway, left] {Known density ratio};
  \node[draw, fill=black!5, rectangle, rounded corners=3pt, minimum height=17pt] (node2) at (-3.22,1.31) {Reweighted estimators  };
  \node[draw, fill=black!5, rectangle, rounded corners=3pt, minimum height=16pt, minimum width=30pt, align=center] (node4) at (2.92,0.67) {Pre-training \\ reweighted estimators };
  \draw[arrows=-Latex] (1.55,2.6) -- (node4.north) node[midway, right] {Unknown density ratio};
  \node[draw, fill=black!5, rectangle, rounded corners=3pt, minimum height=16pt, minimum width=30pt, align=center] (node6) at (-6,-1.34) {Reweighted estimator \\with true \\ density ratio \\ ($\widehat{h}_{r, \mathbb{D}}$)};
  \draw[arrows=-Latex, densely dashed] (-4,1.01) -- (node6.north) node[midway, left] {Uniformly bounded};
  \node[draw, fill=black!5, rectangle, rounded corners=3pt, minimum height=16pt, minimum width=30pt, align=center] (node3) at (-1.56,-1.36) { Reweighted estimator \\with truncated \\density ratio \\ ($\widehat{h}_{ T_\eta r \mathbb{D}}$)};
  \draw[arrows=-Latex, densely dashed] (node2.south) -- (node3.north) node[midway, right] {Unbounded};
  \node[draw, fill=black!5,rectangle, rounded corners=3pt, minimum height=16pt, minimum width=30pt, align=center] (node7) at (2.41,-4.33) {Pre-training reweighted \\ estimator with \\ ratio estimator ($\widehat{h}_{\widehat{r}_{\mathbb{S}}, \mathbb{D}}$)};
  \node[draw, fill=black!5, rectangle, rounded corners=3pt, minimum height=16pt, minimum width=30pt, align=center] (node8) at (7.68,-4.35) { Pre-training reweighted \\ estimator with \\ truncated ratio estimator ($\widehat{h}_{\widehat{r}_{\eta , \mathbb{S}}, \mathbb{D}}$)};
  \draw[arrows=-Latex, densely dashed] (node4.south) -- (node7.north) node[left, near end] {Uniformly bounded};
  \draw[arrows=-Latex, densely dashed] (3.25,0.3) -- (6,-3.75) node[right, pos=0.71] {Unbounded};
\end{tikzpicture}
    
    \caption{Our main contribution: settings considered with the associated SPDNN estimators.}
    \label{Etimators}
\end{figure}

}

 \medskip

 The rest of the paper is organized as follows. Section \ref{section2} presents some notations, assumptions and the deep regression framework.
 Unweighted and reweighted SPDNN estimators are studied in Section \ref{sec_unw_rew_SPDNN}, whereas Section \ref{sec_pretraining_reweighted} focuses on the two steps pre-training reweighted estimators.
  Some applications and examples are considered in Section \ref{sect_exampl}.
 Conclusion and outlook is given in Section \ref{sec_conclusion}, and Section \ref{sec_proofs} is devoted to the proofs of the main results.

\section{Regression framework and deep neural networks }
\label{section2}
\subsection{Some notations} 
Let $E_1$ and $E_2$ be two subsets of separable Banach spaces equipped, respectively, with a norm $\|\cdot\|_{E_1}, \|\cdot\|_{E_2}$.
Let us set some notations for the sequel.
\begin{itemize}
    \item For any function $h : E_1 \to E_2$ and $U \subset E_1$, we set,
    \begin{equation}
    \label{infinite_norm}
        \|h\|_{\infty} = \sup_{x \in E_1} \|h(x)\|_{E_2}, ~  \|h\|_{\infty, U} = \sup_{x \in U} \|h(x)\|_{E_2}.     
    \end{equation}

    \item Denote  by $\mathcal{F}(E_1, E_2)$, the set of measurable functions from $E_1$ to $E_2$.
    \item  Let  $h \in \mathcal{F}(E_1, E_2)$ and $\epsilon > 0$. Set $B(h, \epsilon)$, the ball of radius $\epsilon$ of $\mathcal{F}(E_1, E_2)$ centered at $h$; that is,
    \[
    B(h, \epsilon) = \{ f \in \mathcal{F}(E_1, E_2), \ \|f - h\|_{\infty} \leq \epsilon \}.
    \]
    \item For any $\epsilon > 0$ and  $\mathcal{H} \subset \mathcal{F}(E_1, E_2)$, the $\epsilon$-covering number of $\mathcal{H}$, denoted $\mathcal{N}(\mathcal{H}, \epsilon)$, is the minimal number of balls of radius $\epsilon$ needed to cover $\mathcal{H}$; that is,
    \begin{equation}
    \label{covering_number}
        \mathcal{N}(\mathcal{H}, \epsilon) = \inf \left\{ m \geq 1 : \exists h_1, \ldots, h_m \in \mathcal{H} \ \text{such that} \ 
        \mathcal{H} \subset \bigcup_{i=1}^{m} B(h_i, \epsilon) \right\}.
    \end{equation}

    \item For all $u, v \in \mathbb{R}$, set $u \vee v = \max(u, v)$ and $u \wedge v = \min(u, v)$.
    \item For two sequences of real numbers $(u_n)_n$ and $(v_n)_n$, we write $u_n \lesssim v_n$ or $v_n \gtrsim u_n$ if there exists a constant $C > 0$ such that $u_n \leq C v_n$ for all $n \in \mathbb{N}$. 
    Also, we write $u_n \asymp v_n$ if $u_n \lesssim v_n$ and $u_n \gtrsim v_n$.
   %
   % \item For any finite set $A$, $|A|$ denotes the cardinal of $A$.
\end{itemize}
\subsection{Huber and quantile nonparametric regression under covariate shift}
Recall the model,
\begin{equation}
\label{regression_model2}
    Y_t = h_0(X_t) + \xi_t , \;  t \in \mathbb{Z}
\end{equation} 
where $\mathcal{X} \subset \mathbb{R}^d \; (d \geq 1)$, $\mathcal{Y} \subset \mathbb{R}$, are respectively the input and the output spaces, $h_0 : \mathcal{X} \rightarrow \mathbb{R}$ is the unknown regression function, assumed to be measurable, $(X_t, Y_t)_{t \in \Z}$ is a stationary and ergodic process,
and $\left( \xi_t \right)_{t \in \mathbb{Z}}$ is a centered iid error process satisfying $\mathbb{E} \left( \lvert \xi_0 \rvert \right) < \infty$.
We consider a learning procedure based on a loss function $\ell: \R \times \my \rightarrow [0,\infty)$ with $\ell(y, y') = \ell_\tau(y-y')$ for all $(y, y') \in \R \times \my$.
Two frameworks:
\begin{itemize}
\item \textbf{Huber regression}:  $\ell_\tau$ is the Huber loss with parameter $\tau >0$,  defined  for all $x \in \mathbb{R}$  by:
\[ \ell_\tau(x) = \frac{1}{2} x^2 \ind_{ \{ \lvert x \rvert \leq \tau \} } + \left( \tau \lvert x \rvert - \frac{1}{2} \tau^2 \right)  \ind_{ \{ \lvert x \rvert > \tau \} }. \]  In this case, it is assumed that for all $x \in \mx$, the distribution of $\xi_0|X_0=x$ is symmetric around 0 and there exists $C_\xi >0$ such that 
  $\E(|\xi_0| ~ | X_0=x) \leq C_\xi < \infty$ for all $x \in \mx$.
\item \textbf{Quantile regression}: For all $x \in \mx$ and $\tau \in (0,1)$, the conditional $\tau$-th quantile of $Y_0$ given $X_0 = x$ is $h_0(x)$, such that $P\big(Y_0 - h_0(X_0) \leq 0| X_0=x \big) =  P(\xi_0 \leq 0 | X_0=x) = \tau $.  In this case, $\ell_\tau$ is the $\tau$-th quantile loss defined for all $x \in \mathbb{R}$ by: \[  \ell_\tau(x) = x \left( \tau - \ind_{\{ x < 0\}} \right). \] 
\end{itemize}
We focus on the estimation of $h_0$ from the Huber and the quantile regression under covariate shift, and based on a training sample $\mathbb{D}:=\{ (X_1,Y_1),\cdots,(X_n,Y_n) \}$, which is a trajectory of the process $(X_t, Y_t)_{t \in \Z}$.
For the training and the testing observations, the distribution of $(X_0, Y_0)$ is $\mathbb{P}_{X_0,Y_0}$ and $\mathbb{Q}_{X_0,Y_0}$, with the marginal distribution of $X_0$, $\mathbb{P}_{X_0}$ and $\mathbb{Q}_{X_0}$ respectively.
It is assumed that these distributions have density functions $p_{X_0}$ and $q_{X_0}$ with respect to the Lebesgue measure.
Also, the distribution of $\xi_0 | X_0$ is assumed to be the same, whether $X_0$ is distributed as $\mathbb{P}_{X_0}$ or $\mathbb{Q}_{X_0}$.
This implies that the conditional distribution of $Y_0$ given $X_0$ is the same over the source and the target distributions.
Denote by $F_{Y_0|X_0}$, its cumulative distribution function.

\medskip

In the setting considered for both the Huber and the quantile regression, the regression function $h_0$ satisfies,
\begin{equation}\label{h0_target_function}
  h_0 \in \argmin_{h \in \mathcal{F} }  R_\tau(h) , ~ \text{ where }  R_\tau(h) = \E_{(X_0, Y_0) \sim \mathbb{P}_{X_0,Y_0}} \big[ \ell_\tau \big(Y_0 - h(X_0)  \big) \big],     
\end{equation} 
where $\mathcal{F} := \{ h : \mathcal{X} \to \mathcal{Y}, \text{ measurable} \}$ and $\tau >0$.
Also, for the Huber regression, one can get from the proof of Proposition 3.1 in \cite{fan2024noise} that for any $ h \in \mathcal{F}$,
\begin{equation}\label{local_quad_Huber}
 R_\tau(h) - R_\tau(h_0) \leq \frac{1}{2} \| h - h_0 \|^2_{2, \mathbb{P}_{X_0}},    
\end{equation}
where for all $\kappa \geq 1$,
\[  \| h - h_0 \|^\kappa_{\kappa, \mathbb{P}_{X_0}} = \int  \| h(x) - h_0(x) \|^\kappa dP_{X_0}(x)  .\]

In the case of quantile regression and under \textbf{(A4)} (see below), we get (see, for instance, the proof Theorem 3.13 in \cite{wang2024optimal}),
\begin{equation}\label{local_quad_quantile}
 R_\tau(h) - R_\tau(h_0) \leq \frac{\tilde{c}}{2} \| h - h_0 \|^2_{2, \mathbb{P}_{X_0}},    
\end{equation}
where $\tilde{c}$ is given in \textbf{(A4)}.

\subsection{Some assumptions}
Consider the regression problem above and throughout the sequel, we assume that:
\begin{itemize}
    \item  $\mathcal{X} \subset \mathbb{R}^d$ is a compact set.
    \item For some constant $\mathcal{K} >0$, 
     \[
 \| h_0 \|_{\infty, \mathcal{X}} \leq \mathcal{K} < \infty, 
   \]
 where $h_0$ is the regression function.
  \item There exits $\mathcal{K_{\ell_\tau}}$ such that, the loss function $\ell_\tau$ is $\mathcal{K}_{\ell_\tau}$-Lipchitz. 
\end{itemize}
In the cases of the Huber and the quantile losses $\ell_\tau$, the latter assumption is satisfied with $\mathcal{K}_{\ell_\tau} = \tau$ and $\mathcal{K}_{\ell_\tau} = \max \{ \tau, 1- \tau\}$ respectively.
 Let us set some additional assumptions:

\begin{itemize}
    \item[\textbf{(A1)}:] The process $\mathcal{Z} = \{Z_t = (X_t, Y_t), t \in \mathbb{Z}\}$ is stationary and ergodic. Moreover, let $g : \mathcal{Z} \rightarrow \mathbb{R}$ be a bounded measurable function such that $\mathbb{E} \left[ g(Z_0)\right] = 0, \mathbb{E} \left[ \left( g(Z_0) \right)^2\right] \leq \gamma^2$ and $\| g\|_{\infty} \leq A$ for some $A > 0 $ and $\gamma \geq 0$. Assume that, for all $\varepsilon > 0$, there exits $n_0 \geq 1$ independent of $\varepsilon$ and $\varphi(n) \geq 1$ such that for all $n \geq n_0$, we get
\[
 \mathbb{P} \left\{ \dfrac{1}{n} \sum_{i = 1}^n g(Z_i) \geq \varepsilon \right\} \leq C \exp \left( - \dfrac{\varepsilon^2 \varphi(n)}{c_{\gamma} \gamma^2 + c_A \varepsilon A} \right), 
\]
where $\varphi : \mathbb{N} \rightarrow \mathbb{N}$ is no decreasing function, satisfaying $\varphi(n) \leq n, C > 0$ is a constant independent of $n$ and $c_{\gamma}, c_A$ are positive constants. 
\item[\textbf{(A2)}:]  The density ratio $r(x) = q_{X_0}(x)/p_{X_0}(x)$ is uniformly upper-bounded, that is, $\Gamma := \sup_{x \in \mathcal{X}} r(x) < \infty$.
 \item[\textbf{(A3)}:] There exists some constant $\mu > 0$ such that, the $\mu$-th moment of the density ratio $r$  with respect to $\mathbb{P}_{X_0}$ is finite, that is, $U_\mu := \mathbb{E}_{X_0 \sim \mathbb{P}_{X_0}} [r^{\mu}(X_0)] < \infty$.
 \item[\textbf{(A4)}:]  There exist $\eta_1, \eta_2 , \kappa >0$ such that for any $x \in \mx$, $\delta, y \in \R$ satisfying  $|\delta| \leq \eta_1$ and $|y- h_{0}(x)| \leq \eta_2$ it holds,
\[
\left|F_{Y_0|X_0=x}(y + \delta) - F_{Y_0|X_0=x}(y)\right| \geq \kappa|\delta|, \quad \textit{almost surely}.
\]
Furthermore,  $\sup_{t \in \mathbb{R}} p_{Y_0|X_0=x}(t) \leq \tilde{c} < \infty$ for some constant $\tilde{c} >0$.
 \item[\textbf{(A5)}:] For all $x \in \mx$, $p_{X_0}(x) >0$ and the density ratio $r$ is bounded away from zero, that is, $\Upsilon := \inf_{x \in \mathcal{X}} r(x) > 0$.
\end{itemize}
 
\medskip
\noindent
Assumption \textbf{(A1)} is a generalized Bernstein-type inequality and has also been considered by \cite{hang2016learning}. 
 $\varphi(n)$ is often called the ``effective number of observations'', which refers to a heuristic of the fact that the statistical properties from dependent observations are similar to a suitably number of independent observations, see \cite{zikeba2010effective}.
Several classes of stationary processes, such as $\phi$-mixing, strong mixing, and $\mathcal{C}$-mixing processes satisfy this assumption, see \cite{hang2016learning} and Section \ref{sect_exampl} below. 
This includes many useful autoregressive processes which are $\phi$-mixing  or strong mixing, see for instance \cite{Doukhan1994,chen2000geometric}.
In the literature on covariate shift, assumption \textbf{(A2)} is often used, see \cite{feng2024deep}, \cite{duan2022convergence}, \cite{ma2023optimally}, \cite{feng2023towards}, \cite{cortes2010learning}. 
This condition indicates that the difference between $q_{X_0}$ and $p_{X_0}$ is not too large; and one can easily get 
 $ \| h - h_0 \|^{2}_{2, \mathbb{Q}_{X_0}}  \leq \Gamma  \|  h - h_0 \|^{2}_{2,\mathbb{P}_{X_0}}  $ for all $h \in \mathcal{F}$.
 Let $\mu \geq 1$. Under \textbf{(A2)}, we have $\mathbb{E}_{X_0 \sim \mathbb{P}_{X_0}}[r^\mu (X_0)] = \mathbb{E}_{X_0 \sim \mathbb{Q}_{X_0}}[r^{\mu -1 }(X_0)] \le \Gamma^{\mu -1}$.
 Hence \textbf{(A3)} is satisfied with $U_{\mu} = \Gamma^{\mu -1}$.
 This assumption \textbf{(A3)} which is weaker than assumption \textbf{(A2)}, is introduced to relax the conditions on the density ratio.
 These assumptions \textbf{(A2)} and \textbf{(A3)}, also used by \cite{feng2023towards}, \cite{feng2024deep}, \cite{ma2023optimally} are related to the Rényi divergence between $p_{X_0}$ and $q_{X_0}$, see \cite{cortes2010learning}. 
 Assumption \textbf{(A4)} is common in quantile regression see, for instance, \cite{padilla2022quantile}, \cite{shen2024nonparametric}, \cite{feng2024deep}. 
 If the distribution of $Y_0|X_0$ has a density that is bounded away from zero on any compact set, then the assumption \textbf{(A4)} is satisfied, see \cite{madrid2022risk}.
\textbf{(A5)} ensures that the density ratio $r$ is well-defined and is uniformly lower bounded.

\subsection{Deep neural networks}
\label{section3}
In the sequel, we consider the class of $(L, \mathbf{p})$ architecture DNN based on an activation function $\sigma : \R \Rightarrow \R$, where $L\in \N$ is the number of hidden layers or depth, and $\mathbf{p} = (p_0, p_1, \dots, p_{L+1}) \in \mathbb{N}^{L+2}$ is the vector of widths. 
These functions are written as: 

\begin{equation} \label{eq:dnn_function}
    h: \mathbb{R}^{p_0} \to \mathbb{R}^{p_{L+1}}, \mathbf{x} \mapsto h(\mathbf{x}) = A_{L+1} \circ \sigma_L \circ A_L \circ \sigma_{L-1} \circ \cdots \circ \sigma_1 \circ A_1(\mathbf{x}),
\end{equation}
where $A_j: \mathbb{R}^{p_{j-1}} \to \mathbb{R}^{p_j}$ is a linear affine map, defined by $A_j(\mathbf{x}) := W_j \mathbf{x} + \mathbf{b}_j$, for given $p_j \times p_{j-1}$ weight matrix $W_j$ and shift vector $\mathbf{b}_j \in \mathbb{R}^{p_j}$ and $\sigma_j: \mathbb{R}^{p_j} \to \mathbb{R}^{p_j}$ is a nonlinear element-wise activation map, defined for all $\mathbf{z}=(z_1, \dots, z_{p_j})' \in \R^{p_j}$ by $\sigma_j(\mathbf{z}) = (\sigma(z_1), \dots, \sigma(z_{p_j}))'$, and $'$ denotes the transpose. 
The parameter of a DNN $h$ of the form (\ref{eq:dnn_function}) is: 
\begin{equation} \label{eq:theta_h}
    \theta(h) = (\text{vec}(W_1)', \mathbf{b}_1', \dots, \text{vec}(W_L)', \mathbf{b}_L')',
\end{equation} 
where $\text{vec}(W)$ is the vector obtained by concatenating the column vectors of the matrix $W$.
Denote by $\mathcal{H}_{\sigma, \mathbf{p}_0, p_{L+1}}$ the class of DNN predictors with $p_0$-dimensional input and $p_{L+1}$-dimensional output based on the activation function $\sigma$. In the framework of the regression considered above, $p_0=d$ and $p_{L+1}=1$.
For a DNN $h$ as in (\ref{eq:dnn_function}), let us denote by $\text{depth}(h)$ and $\text{width}(h)$ its depth and width, respectively; that is, $\text{depth}(h)=L$ and $\text{width}(h)=\max_{1 \leq j \leq L} p_j$.
For any constants $L, N, B, F > 0$, set:
\begin{equation} \label{eq:H_sigma_LNB}
\mathcal{H}_\sigma(L, N, B, F) = \left\{h \in \mathcal{H}_{\sigma, d, 1}: \text{depth}(h) \leq L, \text{width}(h) \leq N, \norm{\theta(h)}_\infty \leq B, \norm{h}_{\infty, \mathcal{X}} \leq F \right\}.
\end{equation}
Define the class of sparsity constrained DNN with sparsity level $S > 0$, by:
\begin{equation} \label{eq:H_sigma_LNBF_S}
\mathcal{H}_\sigma(L, N, B, F, S) = \left\{h \in \mathcal{H}_\sigma(L, N, B, F): \norm{\theta(h)}_0 \leq S \right\},
\end{equation}
where $\norm{\mathbf{x}}_0 = \sum_{i=1}^P \mathbb{I}(x_i \neq 0)$, $\norm{\mathbf{x}}_\infty = \max_{1 \leq i \leq P} |x_i|$ for all $\mathbf{x}=(x_1, \dots, x_P)' \in \mathbb{R}^P$ ($p \in \mathbb{N}$). 

\medskip
In the sequel, we will consider the class of DNN above, based on piecewise linear or locally quadratic activation functions, see, for instance   \cite{ohn2019smooth}, \cite{ohn2022nonconvex}.
\begin{definition}
Let $f : \mathbb{R} \to \mathbb{R}$.
\begin{enumerate}
    \item $f$ is continuous piecewise linear (or ``piecewise linear'' for simplicity) if it is continuous and there exists $K$ $(K \in \mathbb{N})$ break points $a_1, \ldots, a_K \in \mathbb{R}$ with $a_1 \leq a_2 \leq \cdots \leq a_K$ such that, for any $k = 1, \cdots, K$, $f'(a_k^-) \neq f'(a_k^+)$ and $f$ is linear on $(-\infty, a_1], [a_1, a_2], \cdots [a_K, \infty)$.
    
    \item $f$ is locally quadratic if there exists an interval $(c,d)$ on which $f$ is three times continuously differentiable with bounded derivatives and there exists $\eta \in (c,d)$ such that $f'(\eta) \neq 0$ and $f''(\eta) \neq 0$.
\end{enumerate}
\end{definition}

\noindent
In the following, we consider the DNN class $\mathcal{H}_{\sigma, d, 1}$ and assume that the activation function $\sigma$ is $C_\sigma$-Lipschitz (for some $C_\sigma > 0$), either piecewise linear or locally quadratic, and fixes a non empty interior segment $I \subset [0, 1]$ (i.e. $\sigma(z) = z$ for all $z \in I$).  
The ReLU activation function ($\sigma(z) = \max(z, 0)$ for all $z \in \R$) and several other classical activation functions satisfy this assumption, see \cite{kengne2025excess}.

\section{Unweighted and reweighted SPDNN estimators and $L_2(\mathbb{Q}_{X_0})$ error bounds}\label{sec_unw_rew_SPDNN}
Let $L_n, N_n, B_n, F>0$. Consider the DNN class $\mathcal{H}_\sigma(L_n, N_n, B_n, F)$.
The unweighted and reweighted SPDNN estimators are based on a sparse penalty term $J_n$ defined for all $h \in \mathcal{H}_\sigma(L_n, N_n, B_n, F)$ by:
\begin{equation}\label{eq:sparse_penalty}
J_n(h) = \sum_{j=1}^{p} \pi_{\lambda_n, \tau_n}(|\theta_j(h)|),
\end{equation}
where $\theta_j(h)$ is the $j$-th component of $\theta(h) = (\theta_1(h), \ldots, \theta_p(h))'$, 
which is the parameter of $h$ (recall that $'$ denotes the transpose) and $\pi_{\lambda_n, \tau_n} : [0, \infty) \to [0, \infty)$ is a positive function based on the tuning parameters $\lambda_n, \tau_n > 0$.
For the function $\pi_{\lambda_n, \tau_n}$, it is assumed that (see also \cite{kurisu2025adaptive}):
\begin{enumerate}
    \item[(i)] $\pi_{\lambda_n, \tau_n}(0) = 0$ and $\pi_{\lambda_n, \tau_n}(\cdot)$ is non-decreasing.
    \item[(ii)] $\pi_{\lambda_n, \tau_n}(x) = \lambda_n$ if $x > \tau_n$.
\end{enumerate}

\medskip
\noindent
An example of such penalty is the clipped $L_1$ penalty (see \cite{zhang2010analysis}) given as in (\ref{eq:sparse_penalty}) with
\[  \pi_{\lambda_n, \tau_n}(x)  = \lambda_n \left( \dfrac{x}{\tau_n} \wedge 1 \right) \text{ for all } x \geq 0 .\]
This example satisfies the conditions (i) and (ii) above.
Other examples are the smoothly clipped absolute deviation (SCAD) penalty studied by \cite{fan2001variable}, the minimax concave penalty \cite{zhang2010nearly} or the seamless $L_0$ penalty, see \cite{dicker2013variable}; which are specific cases of the penalty $J_n$ above, see \cite{kurisu2025adaptive}.

\medskip
Consider the Huber and the quantile nonparametric regressions (\ref{regression_model2}).
Based on the penalty term $J_n$ and the training sample $\mathbb{D}  = \left\{ (X_i,Y_i), i = 1, \cdots , n \right\} $, the unweighted sparse-penalized DNN (SPDNN) estimator of $h_0$ is given by: 
\begin{equation}
\label{unweighted_estimator}
    \widehat{h}_{\mathbb{D}} \in \argmin_{h \in \mathcal{H}_\sigma(L_n, N_n, B_n, F)} \left\{  \dfrac{1}{n} \sum_{i = 1}^{n} \ell_\tau \big(Y_i - h(X_i) \big) + J_n(h) \right\},
\end{equation}
where $\ell_\tau$ is either the Huber or the quantile loss. 
Note that, for any $h \in \mathcal{H}_\sigma(L_n, N_n, B_n, F)$, the empirical risk $ \frac{1}{n} \sum_{i = 1}^{n} \ell_\tau \big(Y_i - h(X_i) \big)$ in (\ref{unweighted_estimator}) is an unbiased estimator of the risk $R_\tau(h) = \E_{(X_0, Y_0) \sim \mathbb{P}_{X_0,Y_0}} \big[ \ell_\tau \big(Y_0 - h(X_0)  \big)]$.
But under covariate shift, we would like to construct an estimator whose the risk on the target data and the $L_2(\mathbb{Q}_{X_0})$ error are small.
So, when the density ratio $r$ is known, consider the reweighted SPDNN estimator defined by: 
\begin{equation}
\label{reweighted_estimator}
    \widehat{h}_{r ,\mathbb{D}} \in \argmin_{h \in \mathcal{H}_\sigma(L_n, N_n, B_n, F)} \left\{  \dfrac{1}{n} \sum_{i = 1}^{n} r  (X_i)  \ell_\tau (Y_i - h(X_i)) + J_n(h) \right\}.
\end{equation}
Therefore, the estimator $\widehat{h}_{r ,\mathbb{D}}$ is computed from the reweighted empirical risk $\frac{1}{n} \sum_{i = 1}^{n} r  (X_i)  \ell_\tau \big(Y_i - h(X_i) \big) $, which is an unbiased estimator of the target risk 
$\widetilde{R}_\tau(h) = \E_{(X_0, Y_0) \sim \mathbb{Q}_{X_0,Y_0}} \big[ \ell_\tau \big(Y_0 - h(X_0)  \big)]$.
In fact, on can easily get that,
$ \E_{(X_0, Y_0) \sim \mathbb{P}_{X_0,Y_0}} \big[ r(X_0)\ell_\tau \big(Y_0 - h(X_0)  \big)]  = \E_{(X_0, Y_0) \sim \mathbb{Q}_{X_0,Y_0}} \big[ \ell_\tau \big(Y_0 - h(X_0)  \big)]$. 
Similar approach has been considered, among other, by \cite{ma2023optimally}, \cite{feng2023towards}, \cite{feng2024deep}. 

\subsection{Some preliminary results}
In the sequel, for a network architecture $L_n, N_n,B_n, F_n >0$, we set 
\begin{equation}\label{H_sigma_n}
\mathcal{H}_{\sigma,n} := \mathcal{H}_{\sigma} \left( L_n, N_n,B_n, F_n \right).    
\end{equation}
Let $h_{\mathcal{H}_{\sigma,n}}$ be a best approximation of $h_0$ within the class $\mathcal{H}_{\sigma,n}$, that is 
\begin{equation}\label{def_h_H_sigma}
    h_{\mathcal{H}_{\sigma,n}} \in \argmin_{h \in \mathcal{H}_{\sigma,n}} \| h - h_0 \|_{\infty, \mathcal{X}}.
\end{equation}
For any estimator $h_n \in \mathcal{H}_{\sigma,n}$, the following lemma establishes a decomposition of the $L_2(\mathbb{P}_{X_0})$ error of $h_n$ into an estimation error (with respect to $h_{\mathcal{H}_{\sigma,n}}$) and the $L_2$ approximation error.
\begin{lemma}
\label{lemma_decomposition_estimator}
    Consider model \eqref{regression_model2},  a DNN class $ \mathcal{H}_{\sigma,n} := \mathcal{H}_\sigma(L_n, N_n, B_n, F)$ for some $L_n, N_n, B_n, F >0$ and assume \textbf{(A4)}. 
    For both the Huber regression with $\tau > 6 \left( \mathcal{K} \lor  F \lor  C_{\xi} \right)$ and the quantile regression, 
    we have for all $ h \in  \mathcal{H}_{\sigma,n}$,
\begin{equation}
\label{unweighted_decomposition}
    \norm{h - h_0 }^2_{2,\mathbb{P}_{X_0}} \leq C_1   \mathbb{E}_{(X_0, Y_0) \sim \mathbb{P}_{X_0,Y_0}} \left[ \ell_{\tau} \left( Y_0 - h(X_0) \right) - \ell_{\tau} \left( Y_0 - h_{\mathcal{H}_{\sigma,n}} (X_0) \right)\right]   + C_2 \inf_{h \in \mathcal{H}_{\sigma,n}} \norm{h - h_0}^2_{\infty, \mathcal{X}},
\end{equation}
for some constants $C_1, C_2 >0$ and where $h_{\mathcal{H}_{\sigma,n}}$ is given in (\ref{def_h_H_sigma}). 
\end{lemma}
\noindent
So, from this Lemma \ref{lemma_decomposition_estimator}, to derive an upper bound of $ \mathbb{E}_{\mathbb{D}} \left[ \|\widehat{h}_{\mathbb{D}} - h_{0}\|^2_{2,\mathbb{P}_{X_0}} \right]$ ($\mathbb{E}_{\mathbb{D}}$ denotes the expectation taken over the training sample $\mathbb{E}_{\mathbb{D}}$), it suffices to bound the statistical error
$\mathbb{E}_{\mathbb{D}}\, \mathbb{E}_{(X_0, Y_0) \sim \mathbb{P}_{X_0,Y_0}}
\big[
\ell_{\tau}\!\left( Y_0 - \widehat{h}_{\mathbb{D}}(X_0) \right)
-\\
\ell_{\tau}\!\left( Y_0 - h_{\mathcal{H}_{\sigma,n}}(X_0) \right)
\big]$
and the approximation error term
$\inf_{h \in \mathcal{H}_{\sigma,n}} \|h - h_0\|^2_{\infty, \mathcal{X}}$.
The latter can be done using Theorem 3.2 in \cite{kengne2025excess}.

\medskip
The next lemma establishes a decomposition of the $L_2(\mathbb{Q}_{X_0})$ error into an estimation error with respect to the reweighted risk and the $L_2$ approximation error.
\begin{lemma}
\label{lemma_decomposition_estimator_reweighted}
     Consider model \eqref{regression_model2},  a DNN class $ \mathcal{H}_{\sigma,n} := \mathcal{H}_\sigma(L_n, N_n, B_n, F)$ for some $L_n, N_n, B_n, F >0$ and assume \textbf{(A4)}. 
    For both the Huber regression with $\tau > 6 \left( \mathcal{K} \lor  F \lor  C_{\xi} \right)$ and the quantile regression, 
    we have for all $ h \in  \mathcal{H}_{\sigma,n}$,
   \[
      \norm{h- h_0}_{2, \mathbb{Q}_{X_0}}^2   \leq \mathbb{E}_{(X_0, Y_0) \sim \mathbb{P}_{X_0, Y_0}}  \left[ r(X_0) \left( \ell_\tau \big( Y_0 - h(X_0)  \big) - \ell_\tau \big( Y_0 - h_{\mathcal{H}_{\sigma,n}}(X_0) \big) \right) \right]   +C_2 \inf_{h \in \mathcal{H}_{\sigma,n}} \| h - h_0\|^2_{\infty, \mx},
   \]
 for some constants $C_1, C_2 >0$, and where $h_{\mathcal{H}_{\sigma,n}}$ is given in (\ref{def_h_H_sigma}). 
\end{lemma}

\subsection{$L_2(\mathbb{Q}_{X_0})$ error bounds of the unweighted and the reweighted estimators}\label{sub_sect_unwei_rewei_theoretical}
In this subsection, we consider the unweighted and the reweighted SPDNN estimators $\widehat{h}_{\mathbb{D}}$ and $\widehat{h}_{r ,\mathbb{D}}$ defined above under covariate shift and focus on bounds of $\left\|\widehat{h}_{\mathbb{D}}-h_0\right\|_{2, \mathbb{Q}_{X_0}}^2$ and $\left\|\widehat{h}_{r,\mathbb{D}}-h_0\right\|_{2, \mathbb{Q}_{X_0}}^2$, where
\begin{equation}
\label{error}
\left\|h-h_0\right\|_{2, \mathbb{Q}_{X_0}}^2
=\mathbb{E}_{X_0\sim \mathbb{Q}_{X_0}}\!\left[\big(h(X_0)-h_0(X_0)\big)^2\right]
=\int_{\mathcal{X}} \big(h(x)-h_0(x)\big)^2\,q_{X_0}(x)\,dx,
\end{equation}
for all predictors $h$ and $q_{X_0}$ is the density (with respect to the Lebesgue measure)  of $\mathbb{Q}_{X_0}$.
To this end, it is assumed that $h_0$ belongs to the class of H\"older smooth functions.
Let $r \in \mathbb{N}$, $D \subset \mathbb{R}^r$, $\beta, A > 0$. 
We will consider the ball of $\beta$-H\"older functions with radius $A$ defined by:
\begin{equation}\label{def_Holder_class}
    \mathcal{C}^{\beta}(D,A) = 
\left\{
h : D \to \mathbb{R} \; : \;
\sum_{\alpha : |\alpha|_1 < \beta} \|\partial^{\alpha} h\|_{\infty}
+ 
\sum_{\alpha : |\alpha|_1 = \lfloor \beta \rfloor} 
\sup_{\substack{x,y \in D \\ x \neq y}}
\frac{|\partial^{\alpha} h(x) - \partial^{\alpha} h(y)|}{|x - y|^{\beta - \lfloor \beta \rfloor}}
\leq A
\right\},
\end{equation}
with $\alpha = (\alpha_1, \ldots, \alpha_r) \in \mathbb{N}^r$, 
$|\alpha|_1 := \sum_{i=1}^r \alpha_i$ and 
$\partial^{\alpha} = \partial^{\alpha_1} \cdots \partial^{\alpha_r}$.

\medskip
 The following theorem establishes non-asymptotic $L_2(\mathbb{Q}_{X_0})$ error bounds of $\widehat{h}_{\mathbb{D}}$ under the uniformly bounded and finite second-order moment conditions on the density ratio.
\begin{theorem}
\label{unweighted_SPDNN_result}
     Consider model \eqref{regression_model2} as either the Huber regression with $\tau > 6 \left( \mathcal{K} \lor  F \lor  C_{\xi} \right)$ or the quantile regression. Assume that \textbf{(A1)} and \textbf{(A4)} hold,  and that $h_0 \in \mathcal{C}^s(\mathcal{X}, \mathcal{K})$ for some $s, \mathcal{K} > 0$ the class of H\"older smooth functions. Consider the DNNs class $\mathcal{H}_\sigma(L_n, N_n, B_n, F_n)$ with the network architecture parameters $(L_n, N_n, B_n, F_n)$ satisfying
     \begin{equation}\label{cond_network_architecture}
    L_n \asymp \log \Big( \varphi (n) \Big), N_n \gtrsim \Big( \varphi (n) \Big)^{\frac{d}{2s+d}}, B_n \gtrsim \Big( \varphi (n) \Big)^{\frac{4(s+d)}{2 s+d}}, F_n = F \geq \mathcal{K} + 1.
     \end{equation}   
     Let $\widehat{h}_{\mathbb{D}}$ be the unweighted SPDNN estimator defined in (\ref{unweighted_estimator}) where the tuning parameters $\lambda_n$ and $\tau_n$ of the penalty term fulfill (\ref{cond_tuning_par}).
     Then,
\begin{enumerate}
    \item Under \textbf{(A2)}, it holds that
     \begin{equation}\label{theo_unif_bounded}
\mathbb{E}_{\mathbb{D}} \left[ \norm{\widehat{h}_{\mathbb{D}} - h_0 }^2_{2,\mathbb{Q}_{X_0}} \right]  \lesssim   \dfrac{\Big(\log \varphi(n) \Big)^{\nu}}{\Big( \varphi (n) \Big)^{\frac{2s}{2s + d}}}, ~ \text{ for all } \nu > 3.
\end{equation}
 \item Under \textbf{(A3)} with $\mu =2$, we have
  \begin{equation}\label{theo_finite_sd_moment}
\mathbb{E}_{\mathbb{D}} \left[ \norm{\widehat{h}_{\mathbb{D}} - h_0 }^2_{2,\mathbb{Q}_{X_0}} \right]  \lesssim  \dfrac{\Big(\log \varphi(n) \Big)^{\nu}}{\Big( \varphi (n) \Big)^{\frac{s}{2s + d}}}, ~ \text{ for all } \nu > 3/2.
\end{equation}
\end{enumerate}
\end{theorem}

\noindent
In the i.i.d. case,  $\varphi(n) = n$ (see Section \ref{sect_exampl}) and the bounds in Theorem \ref{unweighted_SPDNN_result} coincide (up to logarithmic factors), with that obtained in Theorem 8 and 9 of \cite{feng2024deep} in the context of nonparametric quantile regression.
Theorem \ref{unweighted_SPDNN_result} above extends such results to dependent observations and to Huber regression.
For i.i.d. data, $\phi$-mixing, exponential $\alpha$-mixing and $\mathcal{C}$-mixing processes, which satisfy assumption \textbf{(A1)} with $\varphi(n) = n/(\log n)^{\nu_0}$ for some $\nu_0 \geq 0$ (see Section \ref{sect_exampl}), the upper bounds in (\ref{theo_unif_bounded}) match (up to a logarithmic factor) with the minimax lower bound in \cite{stone1982optimal}.
So, when the covariate shift occurs, the unweighted SPDNN estimator $\widehat{h}_{\mathbb{D}} $ remains optimal in the minimax sense if the density ratio $r$ is uniformly bounded.
This condition implies that the shift between $\mathbb{P}_{X_0}$ and $\mathbb{Q}_{X_0}$ is not too large, which leads to the optimality of $\widehat{h}_{\mathbb{D}} $.
Under the weaker condition of finite second-order moment of $r$, the bound in (\ref{theo_finite_sd_moment}) is suboptimal.
As we will see later, this can be overcome with the reweighted estimator based on the truncated density ratio.

\medskip

 Now, let us focus on the $L_2(\mathbb{Q}_{X_0})$ error bound of $\widehat{h}_{r, \mathbb{D}}$. 
 Define the reweighted risk by, 
\begin{equation}\label{def_reweighted_risk}
 R_{\mathbb{P},r} (h) := \mathbb{E}_{(X_0,Y_0) \sim \mathbb{P}_{X_0,Y_0}} \left[ r(X_0) \Big( \ell_{\tau} \left( Y_0- h(X_0) \right) \Big) \right] .
\end{equation}
We have the following proposition.
\begin{proposition}
\label{argmin_h0}
    Consider model \eqref{regression_model2} and assume \textbf{(A4)}. For both the Huber regression with $\tau > 6 \left( \mathcal{K} \lor  F \lor  C_{\xi} \right)$ and the quantile regression, the regression function defined in \eqref{regression_model2} satisfies,
    \[ h_0 \in \argmin_{h \in \mathcal{F} } R_{\mathbb{P},r} (h), \]
 where $\mathcal{F} := \{ h : \mathcal{X} \to \mathcal{Y}, \text{ measurable} \}$.   
\end{proposition}
\noindent
Proposition \ref{argmin_h0} states that, for both the nonparametric quantile and Huber regressions, the regression function $h_0$ is a target function with respect to the reweighted risk. 
This allows us to define the reweighted excess risk of a predictor $h$ as follows:
\begin{equation}\label{def_reweighted_excess_risk}
     \mathcal{E}_{\mathbb{P},r} (h)  := R_{\mathbb{P},r} (h) - R_{\mathbb{P},r} (h_0) = \mathbb{E}_{(X_0,Y_0) \sim \mathbb{P}_{X_0,Y_0}} \left[  r(X_0)  \Big( \ell_\tau (Y_0 - h(X_0) ) - \ell_\tau (Y_0 - h_0 (X_0))\Big) \right]  .
\end{equation}
 The next theorem establishes an oracle inequality for the reweighted excess risk of $\widehat{h}_{r, \mathbb{D}}$. 

\begin{theorem}
\label{oracle_reweighted}
    Assume that \textbf{(A1)} and \textbf{(A2)} with $1 < \Gamma \lesssim \varphi(n)^{1 - \nu_4}$ hold. Let $L_n \lesssim \log\varphi(n)$, $N_n \lesssim \left( \varphi (n)\right)^{\nu_1}$, $1 \leq B_n \lesssim \left( \varphi (n)\right)^{\nu_2}$, $F_n = F > 0$, for some $\nu_1 > 0$, $\nu_2 > 0$. Let $\nu_3 > 2,  0 < \nu_4 \leq 1 $ and the regression function is such that $\|h_0\|_\infty \leq \mathcal{K}$ for some constant $\mathcal{K} > 0$ . Then, there exists $n_0 \in \mathbb{N}$ such that, for all $n \geq n_0$, the reweighted SPDNN estimator defined in \eqref{reweighted_estimator}, with 
    \begin{equation}\label{cond_tuning_par}
   \lambda_n \asymp \frac{(\log \varphi(n))^{\nu_3}}{\varphi(n)^{\nu_4}}, \tau_n \leq \frac{1}{16 \Gamma \mathcal{K}_{\ell_{\tau}}(L_n + 1)((N_n + 1)B_n)^{L_n+1} \varphi(n)}   
    \end{equation}
    satisfies,
\begin{equation}\label{oracle_reweighted_eq}
\mathbb{E}\left[\mathcal{E}_{\mathbb{P}, r}(\widehat{h}_{r, \mathbb{D}})\right] \leq 2\left(\inf_{h \in \mathcal{H}_{\sigma}(L_n,N_n,B_n,F)} \left[\mathcal{E}_{\mathbb{P}, r}(h) + J_n(h)\right]\right) + \frac{\Xi \Gamma }{\varphi(n)}.    
\end{equation}
for some constant $\Xi > 0$, where $\mk_{{\ell}_\tau}$ is the Lipschitz constant of $\ell_\tau$. 
\end{theorem}
\begin{remark}
Under Assumption \textbf{(A2)}, $\Gamma < \infty$. But in the sequel, we will need the inequality (\ref{oracle_reweighted_eq}) in a setting where $\Gamma$ depends on the sample size $n$. 
The condition $1 < \Gamma \lesssim \varphi(n)^{1 - \nu_4}$ in Theorem \ref{oracle_reweighted} is set to take into account this case.
\end{remark}

  \medskip
 The following corollary provides a bound of the reweighted excess risk of $\widehat{h}_{r, \mathbb{D}}$.
\begin{corollary}
    \label{excess_risk_reweighted}
    Assume that \textbf{(A1)} and \textbf{(A2)}  with $1 < \Gamma < \infty$ hold, and that $h_0 \in \mathcal{C}^s(\mathcal{X}, \mathcal{K})$ for some $s, \mathcal{K} > 0$. Let us consider the DNNs class $\mathcal{H}_\sigma(L_n, N_n, B_n, F_n)$   with the network architecture $(L_n, N_n, B_n, F_n)$ satisfying the condition in (\ref{cond_network_architecture}). 
 Then, the reweighted SPDNN estimator $\widehat{h}_{r,\mathbb{D}}$ defined in (\ref{reweighted_estimator}) where the tuning parameters $\lambda_n$ and $\tau_n$ fulfill the condition in (\ref{cond_tuning_par}), satisfies 
  \begin{equation}\label{corrollary_rewei_excess_risk}
      \mathbb{E}[\mathcal{E}_{\mathbb{P},r}(\widehat{h}_{r,\mathbb{D}})] \lesssim \dfrac{\left( \log \varphi(n) \right)^{\nu}}{\left( \varphi (n) \right)^{\frac{2 s}{2 s+d}}}, 
  \end{equation}
   for all $\nu > 3$, and where $\mathcal{E}_{\mathbb{P},r}$ denotes the reweighted excess risk defined in (\ref{def_reweighted_excess_risk}).
\end{corollary}

\noindent
The bound in (\ref{corrollary_rewei_excess_risk}) coincides with that obtained when there is no covariate shift, see Corollary 3.4 in \cite{kengne2025general}.
So, under the assumption \textbf{(A2)}, the estimator $\widehat{h}_{r,\mathbb{D}}$ displays the same reweighted excess risk bound as when there is no shift.
  
\medskip
The next theorem establishes an upper bound of the $L_2(\mathbb{Q}_{X_0})$ error of the reweighted estimator when the ratio $r$ is uniformly bounded.
\begin{theorem}
\label{reweighted_estimator_result}
 Consider model \eqref{regression_model2} as either the Huber regression with $\tau > 6 \left( \mathcal{K} \lor  F \lor  C_{\xi} \right)$ or the quantile regression.  Assume that \textbf{(A1)},  \textbf{(A2)} and \textbf{(A4)} hold, and that $h_0 \in \mathcal{C}^s(\mathcal{X}, \mathcal{K})$ for some $s, \mathcal{K} > 0$.
 Let us consider the DNNs class $\mathcal{H}_\sigma(L_n, N_n, B_n, F_n)$ with the network architecture $(L_n, N_n, B_n, F_n)$ satisfying the condition in (\ref{cond_network_architecture}). 
  Then, the reweighted SPDNN estimator $\widehat{h}_{r,\mathbb{D}}$ defined in (\ref{reweighted_estimator}) where the tuning parameters $\lambda_n$ and $\tau_n$ fulfill the condition in (\ref{cond_tuning_par}), satisfies
    \begin{equation}\label{theo_L2_Q_rewei_uniform}
\E_{\mathbb{D}}\left[ \norm{\widehat{h}_{r, \mathbb{D}} - h_0}_{2, \mathbb{Q}_{X_0}}^2  \right] \lesssim \dfrac{\Big(\log \varphi(n) \Big)^{\nu}}{\Big( \varphi (n) \Big)^{\frac{2s}{2s + d}}} ,   
    \end{equation} 
for all $\nu > 3$.
\end{theorem}
\noindent
As in the comment above, the reweighted estimator $\widehat{h}_{r,\mathbb{D}}$ can achieves (up to a logarithmic factor), the minimax optimal rate for i.i.d. observations, $\phi$-mixing, exponential $\alpha$-mixing processes, and geometrically $C$-mixing processes.
Moreover, these processes can fulfill \textbf{(A1)} with $\varphi(n) = n/(\log n)^{\nu_0}$ for some $\nu_0 \geq 0$ (see Section \ref{sect_exampl}). 
The network architecture $L_n, N_n, B_n, F_n$ can be chosen independently of the smoothness parameter $s$ such that (\ref{cond_network_architecture}) is satisfied.
Which shows that, $\widehat{h}_{r,\mathbb{D}}$ can adaptively attain (up to a logarithmic factor) the minimax optimal rates on the class of H\"older smooth functions.

\medskip
 
Let us investigate the case where the density ratio $r$ can be unbounded, but has a finite second-order moment.
To this end, consider the truncated density ratio of a threshold $\eta >0$ defined by,
\begin{equation}\label{def_T_eta}
    T_\eta r(x) = \begin{cases} r(x), & \text{if } r(x) \leq \eta, \\ \eta, & \text{otherwise.} \end{cases} 
\end{equation} 
Thus, the reweighted estimator based on the truncated density ratio is defined by,
\begin{equation}\label{def_rewei_truccated_est}
    \widehat{h}_{T_\eta r,\mathbb{D}} \in \argmin_{h \in \mathcal{H}_{\sigma}(L_n, N_n, B_n, F)} \left\{  \dfrac{1}{n} \sum_{i = 1}^{n} T_\eta r (X_i)  \ell_\tau (Y_i - h(X_i)) + J_n(h) \right\}.
\end{equation}
The following theorem derives a non-asymptotic $L_2(\mathbb{Q}_{X_0})$ error bound for the reweighted estimator $\widehat{h}_{T_\eta r, \mathbb{D}}$ on the class of H\"older smooth functions. 
\begin{theorem}
\label{reweighted_estimator_truncated_ratio_result}
     Consider model \eqref{regression_model2} as either the Huber regression with $\tau > 6 \left( \mathcal{K} \lor  F \lor  C_{\xi} \right)$ or the quantile regression.  Assume that \textbf{(A1)}, \textbf{(A3)} with $\mu=1+\delta$ for some $\delta >0$ and \textbf{(A4)} hold, and that $h_0 \in \mathcal{C}^s(\mathcal{X}, \mathcal{K})$ for some $s, \mathcal{K} > 0$.  
     Consider the DNNs class $\mathcal{H}_\sigma(L_n, N_n, B_n, F_n)$ with the network architecture $(L_n, N_n, B_n, F_n)$ satisfying the condition in (\ref{cond_network_architecture}).
    Then, the reweighted estimator with a truncated density ratio defined in (\ref{def_rewei_truccated_est}) with $\eta \asymp \Big( \varphi(n) \Big)^{\frac{2 s}{(1 + \delta)(2 s +d)}} $ where $\lambda_n$ and $\tau_n$ fulfill the condition in (\ref{cond_tuning_par}) with $\nu_4 = \frac{d}{2s + d}$, satisfies,
   \begin{equation}\label{theo_reweighted_truncated_bound}
           \E_{\mathbb{D}}\left[ \norm{\widehat{h}_{T_\eta r, \mathbb{D}} - h_0}_{2, \mathbb{Q}_{X_0}}^2  \right]  \lesssim \dfrac{\Big(\log \varphi(n) \Big)^{\nu}}{\Big( \varphi (n) \Big)^{\frac{\delta}{ 1+ \delta} \times \frac{2s}{2s + d}}}, 
   \end{equation}
for all $\nu > 3$.
\end{theorem}
Therefore, when $\delta \to +\infty$ (that is, $\mu \to +\infty$ in \textbf{(A3)} i.e. $r$ has moments of any order), the bound in (\ref{theo_reweighted_truncated_bound}) coincides with that of the reweighted estimator under the uniformly bounded condition on the density ratio $r$ obtained in (\ref{theo_L2_Q_rewei_uniform}).
Which shows that the reweighted estimator based on the truncated density ratio can be optimal (up to a logarithmic factor) in the minimax sense, for example, for the class of mixing processes listed above.

\section{Pre-training reweighted SPDNN estimators and error bounds}\label{sec_pretraining_reweighted}

\subsection{Pre-training reweighted estimators}
When the density ratio $r$ is unknown, a natural approach is to  seek its estimator.
In this section, we perform a two-step reweighted estimation procedure, where the first step focuses on the estimate of $r$, which is considered in the second step to construct a reweighted estimator of $h_0$. Such an approach has also been considered in \cite{feng2024deep}.

\medskip
\noindent
We will carried out a  least squares estimation of $r$, which is a minimizer of $u \mapsto \frac{1}{2}\mathbb{E}_{X_0 \sim \mathbb{P}_{X_0}} \left[(u(X_0)  -  r (X_0))^2\right]$ for $u \in \mathcal{F} = \{ h : \mathcal{X} \to \mathcal{Y}, \text{ measurable} \} $.
From a simple calculation, we get
\begin{equation*} %\label{rho}
r  \in  \argmin_{u \in \mathcal{F} } \frac{1}{2}\mathbb{E}_{X_0 \sim \mathbb{P}_{X_0}} [u(X_0)^2] - \mathbb{E}_{X_0 \sim \mathbb{Q}_{X_0}} [u(X_0)].
\end{equation*}
Suppose that we get the extra unlabeled samples $S_{\mathbb{P}} := \{X_i^P\}_{i=1}^m$ and $S_{\mathbb{Q}} := \{X_i^Q\}_{i=1}^m$ drawn from $\mathbb{P}_{X_0}$ and $\mathbb{Q}_{X_0}$ respectively.
So, a sparse-penalized DNN estimator of $r$ is given by,
\begin{equation}
\label{rho_estimator}
\widehat{r }_{\mathbb{S}} \in \argmin_{u \in \mathcal{H}_\sigma(L_m, N_m, B_m, F_m)}  \left\{ \dfrac{1}{2m} \sum_{i=1}^m u(X_i^P)^2 - \frac{1}{m} \sum_{i=1}^m u(X_i^Q) + J_m(u) \right \},
\end{equation}
for a DNN class $\mathcal{H}_\sigma(L_m, N_m, B_m, F_m)$.
In the second step, we proceed as in (\ref{reweighted_estimator}), but here we replace $r$ with its estimator above. 
Hence, the pre-training reweighted SPDNN estimator of $h_0$ is defined by,
\begin{equation}\label{def_h_hat_r_hat}
    \widehat{h}_{\widehat{r }_{\mathbb{S}},\mathbb{D}} \in \argmin_{h \in \mathcal{H}_{\sigma}(L_n, N_n, B_n, F_n)} \left\{  \dfrac{1}{n} \sum_{i = 1}^{n} \widehat{r }_{\mathbb{S}} (X_i)  \ell_\tau (Y_i - h(X_i)) + J_n(h) \right\}.
\end{equation}
We assume the common hypothesis (see also \cite{duan2022convergence}, \cite{feng2024deep}) that the unlabeled data $\mathbb{S}_{\mathbb{P}}$ and $\mathbb{S}_{\mathbb{Q}}$ used for the estimation of $r$, are independent of the training sample $\mathbb{D}$.

 \medskip
 In many practical problems, the density ratio $r$ may be unbounded or very takes very large values.
 This is the case, for example, if $q_{X_0}$ is heavy-tailed and $p_{X_0}$ is light-tailed.
 In such cases, estimating $r$ can become difficult, since we deal with a class of bounded DNNs.
 In order to overcome this problem, we set a threshold level $\eta >0$ (as in \cite{feng2024deep}) that the values of $r$ must not exceed. 
 So, the truncated estimator of $r$, that takes into account this threshold is defined as,
\begin{equation}
\label{rho_estimator_tronqued}
\widehat{r}_{\eta, \mathbb{S}} \in \argmin_{u \in T_\eta \mathcal{H}_\sigma(L_m, N_m, B_m, F)}  \left\{ \dfrac{1}{2m} \sum_{i=1}^m u(X_i^P)^2 - \frac{1}{m} \sum_{i=1}^m u(X_i^Q) + J_m(u) \right \},
\end{equation}
where $T_\eta \mathcal{H}_\sigma(L_m, N_m, B_m, F) = \left\{ T_\eta u : u \in \mathcal{H}_\sigma(L_m, N_m, B_m, F) \right\}$ denotes a truncated set of the DNN functions and $T_\eta$ is defined in (\ref{def_T_eta}).
As above, the pre-training reweighted SPDNN estimator of $h_0$, based on this truncated estimator $\widehat{r }_{\eta, \mathbb{S}}$ is given by,
\begin{equation}\label{def_h_hat_r_hat_eta_S}
    \widehat{h}_{\widehat{r }_{\eta, \mathbb{S}},\mathbb{D}} \in \argmin_{h \in \mathcal{H}_{\sigma}(L_n, N_n, B_n, F)} \left\{  \dfrac{1}{n} \sum_{i = 1}^{n} \widehat{r }_{\eta, \mathbb{S}} (X_i)  \ell_\tau (Y_i - h(X_i)) + J_n(h) \right\}.
\end{equation}

\subsection{Error bounds}
In this subsection, we establish non-asymptotic error bounds of the estimators of $r$ and that of the reweighted pre-training estimators of $h_0$, as defined above.
The following proposition provides a $L_2(\mathbb{P}_{X_0})$ error bound of the estimator $\widehat{r}_{\mathbb{S}}$ on the class of H\"older functions.
\begin{proposition}
\label{ratio_estimator_result}
Assume that \textbf{(A1)}, \textbf{(A2)} and \textbf{(A4)} hold,  and that $r\in \mathcal{C}^{s}(\mathcal{X}, \Gamma)$ for some $s >  0$.
 Consider the DNN class $\mathcal{H}_{\sigma,m} := \mathcal{H}_\sigma(L_m, N_m, B_m, F_m)$ where the network architecture parameters $(L_m, N_m, B_m, F_m)$ satisfy
\begin{equation}\label{cond_arch_H_m}
L_m \asymp \log \Big( \varphi (m) \Big), N_m \gtrsim \Big( \varphi (m) \Big)^{\frac{d}{2 s +d}}, B_m \gtrsim \Big( \varphi (m) \Big)^{\frac{4(\alpha+d)}{2 s +d}}, F_m = F \geq \Gamma + 1 .
\end{equation}
Consider the DNN estimator of $r$, $\widehat{r}_{\mathbb{S}}$ defined in (\ref{rho_estimator}) where the tuning parameters $\lambda_m$ and $\tau_m$ of the penalty term $J_m$ satisfy,
\begin{equation}\label{cond_tuning_param}
\lambda_m \asymp \frac{\left(\log \Big( \varphi (m) \Big) \right)^{v_0}}{ \varphi (m) } ~ \text{ for some } \nu_0 >2  ~ \text{ and } ~ \tau_m \leq \frac{1}{32(F+1)(L_m + 1)((N_m + 1)B_m)^{L_m+1} \varphi (m) }.
\end{equation}
Then, we have
\[ \mathbb{E}_{\mathbb{S}} \left[ \norm{\widehat{r}_{\mathbb{S}} - r }^2_{2,\mathbb{P}_{X_0}} \right]  \lesssim   \dfrac{\Big(\log \varphi(m) \Big)^{\nu}}{\Big( \varphi (m) \Big)^{\frac{2s}{2 s  + d}}}, \]
for all $\nu > 3$.
\end{proposition}
\noindent
As the comments in Subsection \ref{sub_sect_unwei_rewei_theoretical} above, Proposition \ref{ratio_estimator_result}  shows that the estimator $\widehat{r}_{\mathbb{S}} $ can be minimax optimal (up to a logarithmic factor) for many useful processes.
 Based on this proposition, the next theorem establishes a $L_2(\mathbb{Q}_{X_0})$ error bound of the pre-training reweighted SPDNN estimator $\widehat{h}_{\widehat{r}_\mathbb{S}, \mathbb{D} }$.

\begin{theorem}
\label{pretraining_reweighted_estimator_result}
 Consider model \eqref{regression_model2} as either the Huber regression with $\tau > 6 \left( \mathcal{K} \lor  F \lor  C_{\xi} \right)$ or the quantile regression.  Assume that \textbf{(A1)}, \textbf{(A2)}, \textbf{(A4)} and \textbf{(A5)} hold, and that $r \in \mathcal{C}^{s_1}(\mathcal{X},\Gamma) $, $h_0 \in \mathcal{C}^{s_2}(\mathcal{X}, \mathcal{K})$ for some $s_1, s_2, \mathcal{K} > 0$.  
 Consider the estimator $\widehat{r}_{\mathbb{S}}$ defined (\ref{rho_estimator}) based on the DNN class $\mathcal{H}_{\sigma,m} := \mathcal{H}_\sigma(L_m, N_m, B_m, F_m)$ where the network architecture parameters $(L_m, N_m, B_m, F_m)$ and the tuning parameters $\lambda_m$ and $\tau_m$ fulfill (\ref{cond_arch_H_m}) and (\ref{cond_tuning_param}).
Consider the estimator $\widehat{h}_{\widehat{r}_\mathbb{S}, \mathbb{D} }$ defined in (\ref{def_h_hat_r_hat}) based on the DNN class $\mathcal{H}_\sigma(L_n, N_n, B_n, F_n)$ where the network architecture $(L_m, N_m, B_m, F_m)$ and the tuning parameters $\lambda_n, \tau_n$ satisfy (\ref{cond_network_architecture}) and (\ref{cond_tuning_par}) with $\nu_4 = 1$.
Then, it holds that 
\begin{equation}\label{pretraining_reweighted_estimator_result_bound}
    \mathbb{E}_{\mathbb{S}, \mathbb{D}} \left[ \norm{\widehat{h}_{\widehat{r}_\mathbb{S}, \mathbb{D} } - h_0 }^2_{2, \mathbb{Q}_{X_0}} \right]\lesssim \dfrac{\left( \log \varphi (m) \right)^{\nu_1}}{\left( \varphi (m) \right)^{\frac{2 s_1}{2 s_1 + d}}} +  \dfrac{\left( \log \varphi (n) \right)^{\nu_2}}{\left( \varphi (n) \right)^{\frac{2s_2}{2s_2 + d}}},
\end{equation} 
for all $\nu_1, \nu_2 >3$.
\end{theorem}
\noindent
Theorem \ref{pretraining_reweighted_estimator_result} shows the effect of the first step (estimation of the density ratio $r$) on the error bound of the pre-training reweighted estimator $\widehat{h}_{\widehat{r}_\mathbb{S}, \mathbb{D} }$.
This bound is the error of the estimation of $r$ and that of the estimation of $h_0$ when $r$ is known.

\begin{remark}\label{remark_pretraining_reweighted_estimator}
The bound in (\ref{pretraining_reweighted_estimator_result_bound}) is of order $\mathcal{O}\Bigg(   \dfrac{\left( \log \varphi (m) \right)^{3}}{\left( \varphi (m) \right)^{\frac{2 s_1}{2 s_1 + d}}} +  \dfrac{\left( \log \varphi (n) \right)^{3}}{\left( \varphi (n) \right)^{\frac{2s_2}{2s_2 + d}}}  \Bigg)$.
We have 
\begin{equation}\label{remark_pretraining_reweighted_estimator_eq}
\frac{\left( \log \varphi (m) \right)^{3}}{\left( \varphi (m) \right)^{\frac{2 s_1}{2 s_1 + d}}} \leq \frac{\left( \log \varphi (n) \right)^{3}}{\left( \varphi (n) \right)^{\frac{2s_2}{2s_2 + d}}}    \Longleftrightarrow  \frac{ \varphi (m) }{ \big( \log \varphi (m) \big)^{3(2s_1 +d)/2s_1 } }  \geq  \frac{ \big( \varphi (n) \big)^{s_2(2s_1 + d)/s_1(2s_2 +d)}  }{ \big( \log \varphi (n) \big)^{3(2s_1 +d)/2s_1 }  }.  
\end{equation}  
Let $n \geq 1$ fixed.
The function $m \mapsto \frac{ \varphi (m) }{ \big( \log \varphi (m) \big)^{3(2s_1 +d)/2s_1 } } $ increases to $\infty$. 
Hence, there exists $m_0 \geq 1$ depending $n, d, s_1, s_2$ such that the inequalities in (\ref{remark_pretraining_reweighted_estimator_eq}) holds for all $m \geq m_0$. 
Therefore, for sufficiently large $m$, the convergence rate in (\ref{remark_pretraining_reweighted_estimator}) is of order
$\mathcal{O}\Bigg(  \dfrac{\left( \log \varphi (n) \right)^{3}}{\left( \varphi (n) \right)^{\frac{2s_2}{2s_2 + d}}}  \Bigg)$.
In this case,  the estimator $\widehat{h}_{\widehat{r}_\mathbb{S}, \mathbb{D} } $ achieves (up to a logarithmic factor) the minimax optimal rate for many examples of useful processes.
\end{remark}

Let us consider the setting where the density ratio $r$ is unknown and may be unbounded.
In this case, the truncated estimator $\widehat{r}_{\eta, \mathbb{S}}$ is performed and the following proposition establishes its $L_2(\mathbb{P}_{X_0})$ error bound. 

\begin{proposition}
\label{truncated_ratio_result}
Assume that \textbf{(A1)}, \textbf{(A3)} with $\mu = 2+ \delta$  for some $\delta >0$ and  \textbf{(A4)}  hold.
Consider the estimator $\widehat{r}_{\eta, \mathbb{S}}$ define in (\ref{rho_estimator_tronqued}), based on the DNN class $\mathcal{H}_{\sigma,m} := \mathcal{H}_\sigma(L_m, N_m, B_m, F_m)$, where the network architecture $L_m, N_m, B_m$ satisfy (\ref{cond_arch_H_m}) and $F_m =\eta +1$ with $\eta \asymp \Big( \varphi(m) \Big)^{\frac{2 s}{(1 + \delta)(2 s +d)}} $.
Also, assume that the tuning parameters $\lambda_m$, $\tau_m$ fulfill 
\begin{equation}\label{cond_tuning_param_truncate}
\lambda_m \asymp \frac{\left(\log \Big( \varphi (m) \Big) \right)^{v_0}}{\Big( \varphi (m) \Big)^{\frac{d}{2s + d}}} ~ \text{ for some } \nu_0 >2 , s >0 ~ \text{ and } ~ \tau_m \leq \frac{1}{32(\eta +1)(L_m + 1)((N_m + 1)B_m)^{L_m+1} \varphi (m) }.
\end{equation}
If $T_{\eta} r \in \mathcal{C}^{s} ( \mathcal{X}, \eta)$, then,
\[  \mathbb{E}_{\mathbb{S}} \left[ \norm{\widehat{r}_{\eta, \mathbb{S}} - r}_{2,\mathbb{P}_{X_0}}^2 \right] \lesssim \dfrac{\left( \log \varphi (m) \right)^{\nu}}{\left( \varphi (m) \right)^{\frac{2 s}{2 s + d} \times\frac{\delta}{1+ \delta}}}, \]
for all $\nu > 3$.
\end{proposition}
\noindent
Therefore, as above, if the ratio $r$ has moments of any order, that is $\delta \to \infty$ (i.e. $\mu \to +\infty$ in \textbf{(A3)}), then the truncated estimator $\widehat{r}_{\eta, \mathbb{S}} $ is optimal (up to a logarithmic factor) in the minimax sense, for several class of useful processes.
This proposition is used to derive a $L_2(\mathbb{Q}_{X_0})$ error bound of the pre-training reweighted SPDNN estimator $\widehat{h}_{\widehat{r}_{\eta, \mathbb{S}}, \mathbb{D} }$.
\begin{theorem} \label{pretraining_reweighted_truncated_ratio_estimator_result}
 Consider model \eqref{regression_model2} as either the Huber regression with $\tau > 6 \left( \mathcal{K} \lor  F \lor  C_{\xi} \right)$ or the quantile regression.  Assume that \textbf{(A1)}, \textbf{(A3)} with $\mu = 2+ \delta$  for some $\delta >0$, \textbf{(A4)}  and \textbf{(A5)} hold, and that $h_0 \in \mathcal{C}^{s_2}(\mathcal{X}, \mathcal{K})$ for some $s_2, \mathcal{K} > 0$.
  Consider the estimator $\widehat{r}_{\mathbb{S}}$ defined in (\ref{rho_estimator}) based on the DNN class $\mathcal{H}_{\sigma,m} := \mathcal{H}_\sigma(L_m, N_m, B_m, F_m)$ where the network architecture parameters $(L_m, N_m, B_m, F_m)$, the tuning parameters $\lambda_m$, $\tau_m$, and the truncation level $\eta$ fulfill the conditions in Proposition \ref{truncated_ratio_result}, and $T_{\eta} r \in \mathcal{C}^{s_1} ( \mathcal{X}, \eta)$ for some $s_1 >0$.
Consider the estimator $\widehat{h}_{\widehat{r}_{\eta, \mathbb{S}}, \mathbb{D} }$ defined in (\ref{def_h_hat_r_hat_eta_S}) based on the DNN class $\mathcal{H}_\sigma(L_n, N_n, B_n, F_n)$ where the network architecture $(L_m, N_m, B_m, F_m)$ and the tuning parameters $\lambda_n, \tau_n$ satisfy (\ref{cond_network_architecture}) and (\ref{cond_tuning_par}) with $\nu_4 = \frac{d}{2s_2+d}$.
Then, it holds that 
\begin{equation}\label{pretraining_reweighted_truncated_eq}
  \mathbb{E}_{\mathbb{S}, \mathbb{D}} \left[ \norm{\widehat{h}_{\widehat{r}_{\eta, \mathbb{S}}, \mathbb{D} } - h_0 }^2_{2, \mathbb{Q}_{X_0}} \right]  \lesssim  \dfrac{\left( \log \varphi (m) \right)^{\nu_1}}{\left( \varphi (m) \right)^{\frac{2 s_1}{2 s_1 + d} \times\frac{\delta}{1+ \delta}}}  + \dfrac{\big(\log \varphi(n) \big)^{\nu_2}  \left( \varphi(m) \right) ^{\frac{2 s_1}{(1+ \delta)(2 s_1 + d)}}}{\big( \varphi (n) \big)^{\frac{2s_2}{2s_2 + d}}}.    
\end{equation} 
for all $\nu_1, \nu_2 >3$.
\end{theorem}

\noindent
As in Remark \ref{remark_pretraining_reweighted_estimator},  for sufficiently large $m$, the bound in (\ref{pretraining_reweighted_truncated_eq}) is of order $\mathcal{O}\Bigg( \dfrac{\big(\log \varphi(n) \big)^{3}}{\big( \varphi (n) \big)^{ \frac{\delta}{1 + \delta} \times \frac{2s}{2s + d}}} \Bigg) $.
In such cases, even if the ratio $r$ is unbounded but has moments of any order (that is $\delta \to \infty$), the truncated based pre-training reweighted estimator $\widehat{h}_{\widehat{r}_{\eta, \mathbb{S}}, \mathbb{D} }$ achieves (up to a logarithmic factor) the minimax optimal rate for several examples of useful processes.

\section{ Some applications and examples}\label{sect_exampl}

Consider a probability space $(\Omega, \mathcal{B}, P)$, and a stationary process $Z = \{Z_t\}_{t\in \mathbb{Z}}$ on this space. For any $-\infty \leq k \leq + \infty$, denote by $\sigma_{-\infty}^k := \sigma(Z_j, j \leq k)$ and $\sigma^{+\infty}_k := \sigma(Z_j, j \geq k) $ the $\sigma$-field generated by the whole past and future at time $k$.
Recall that $\mathcal{Z} = \mathcal{X} \times \mathcal{Y} \subset \mathbb{R}^d \times \mathbb{R}$.
Consider a stationary and ergodic $Z=\{Z_t = (X_t, Y_t), t \in \mathbb{Z}\}$ with values in $\mathcal{Z}$ and solution of (\ref{regression_model2}).

\begin{enumerate}
    \item \textbf{I.I.d. processes.} Assume that the process $Z = \{Z_t\}_{t\in \mathbb{Z}}$ is i.i.d. By fixing $n_0 = 1$, $C = 1$, $c_\gamma = 2$, $c_A = \dfrac{2}{3}$, and $\varphi(n) = n$ in assumption \textbf{(A1)}, we find the classical Bernstein inequality.
    \item \textbf{$\phi$-mixing processes.} The $\mathcal{Z}$-valued process $Z = \{Z_t\}_{t\in \mathbb{Z}} $ is  $\phi$-mixing, if its $\phi$-mixing coefficient, $\phi ( k)$ with $k \in \mathbb{Z}$ satisfies,
    \[ \phi( k ) := \underset{A \in \sigma_{-\infty}^0, B \in\sigma_k^{\infty}}{\sup} \{ |P(B) - P(B|A|\}   \underset{ k \to \infty}{\longrightarrow} 0.
    \]
    From \cite{samson2000concentration}, we find that $\phi$-mixing processes satisfy assumption \textbf{(A1)} with $n_0 = 1$, $C = 1$, $c_\gamma = 32 \sum_{k \geq 1} \sqrt{\phi(k)}$, $c_A = 8\sum_{k \geq 1} \sqrt{\phi(k)}$, and $\varphi(n) = n$. 
    
     \item \textbf{Exponential strong mixing processes.} The $\mathcal{Z}$-valued process $Z = \{Z_t\}_{t\in \mathbb{Z}} $ is said to be $\alpha$-mixing or strongly mixing,  if its $\alpha$-mixing coefficient, $\alpha ( k)$ with $k \in \mathbb{Z},$ satisfies,
        \[ \alpha (k) :=  \underset{A \in \sigma_{-\infty}^0, B \in\sigma_k^{\infty}}{\sup} \{ |P(A \cap B) - P(A)P(B) |\}  \underset{ k \to \infty}{\longrightarrow} 0 .
        \]
        In addition, $Z$ is exponential strongly mixing if,
        \begin{equation*}
        \alpha(k) \leq c\exp \big(-bk^{\varrho}\big),  ~~ \text{ for all } k \ge 1,  \end{equation*}
        for some constants $b > 0$, $c \geq 0$, and $\varrho \ge 1$.
        \medskip
        
        \noindent
        It follows from \cite{hang2016learning}, \cite{merlevede2009bernstein} that an exponential $\alpha$-mixing process satisfies assumption \textbf{(A1)} for some $n_0 \geq 1$, $C, c_\gamma, c_A \geq 0$ with $\varphi(n) = n/\big(\log n \big)^2$ for all $n \geq n_0$.
    \item \textbf{Geometrically $\mathcal{C}$-mixing processes.}  Recall that $(\Omega, \mathcal{B}, P)$ denotes a probability space. For any $p \geq 1$, let $L_p(P) := \{f : \Omega \to \mathbb{R}\text{ measurable: } \|f\|_p < \infty\}$, where $\|f\|_p = \big( \int_\Omega |f(x)| dP(x)  \big)^{1/p} $. Moreover, if $\mathcal{B}' \subset \mathcal{B}$ is a sub-$\sigma$-algebra, we denote by $L_p(\mathcal{B'}, P) := \{f : \Omega \to \mathbb{R},  ~\mathcal{B'}-\text{measurable: } \|f\|_{p} < \infty \}$.
The $\mathcal{C}$-norm is defined by $\norm{f}_{\mathcal{C}} = \norm{f} + \norm{f}_{\infty}$ where  $\| \cdot\|$ is a a semi-norm on the vector space of bounded measurable functions $f : \mathcal{Z} \rightarrow \mathbb{R}$ and $\|\cdot\|_{\infty}$ is the sup-norm. Therefore, $\mathcal{C}(\mathcal{Z}) := \{f : \mathcal{Z} \to \mathbb{R}: \|f\|_{\mathcal{C}} < \infty\}$ denotes the set of all bounded $\mathcal{C}$-functions.

\medskip
\noindent
The $\mathcal{Z}$-valued process $Z=\{Z_t\}_{t\in \mathbb{Z}} $ is said to be $\mathcal{C}$-mixing (see \cite{maume2006exponential}, \cite{hang2017bernstein}),  if its $\mathcal{C}$-mixing coefficient, $\phi_{\mathcal{C}}(k)$ with $k \in \mathbb{Z}$,  satisfies,
\[
\phi_{\mathcal{C}}(k) := \sup_{j\geq 1} \Big\{ \big| \mathbb{E}\big( Z f(Z_{j+k}) \big) - \mathbb{E}(Z)\mathbb{E} \big(f(Z_{j+k}) \big) \big|:  Z \in  L_1(\sigma_{1}^ j, P), \|Z\|_1 \leq 1, f \in \mathcal{C}(\mathcal{Z}), \|f\|_{\mathcal{C}} \leq 1 \Big\} \underset{k \rightarrow}{\longrightarrow}  0,
\] 
Furthermore, $Z$ is  geometrically $\mathcal{C}$-mixing if,
\begin{equation*}
\phi_{\mathcal{C}}(k) \leq c\exp\big(-bk^{\varrho}\big),  ~~ \text{ for all } k\ge 1,   
\end{equation*}
for some constants $c > 0$, $b > 0$, and $\varrho > 0$.
From \cite{hang2016learning}, we find that geometrically $\mathcal{C}$-mixing processes satisfy assumption \textbf{(A1)} for some $n_0 \geq 1$, and with $C=2, c_\gamma = 8, c_A=8/3$, $\varphi(n) = n/\big(\log n \big)^{2/\varrho}$ for all $n \geq n_0$.
\end{enumerate}
 The mixing conditions above include many useful autoregressive processes, such as, ARX, TARX, ARMAX among other, and some discrete dynamical systems, see for instance \cite{Doukhan1994}, \cite{maume2006exponential} for more details.
 
 \medskip
 For the classes of dependence structures above, Table \ref{table_of_some_examples} displays the ``effective number of observations'' $\varphi(n)$, the $L_2(\mathbb{Q}_{X_0})$ error bounds of the unweighted, reweighted and pre-training reweighted SPDNN estimators for both the cases where the density ratio is uniformly bounded and may be unbounded.
More precisely, this table provides the $L_2(\mathbb{Q}_{X_0})$ error bounds of the unweighted, the reweighted, the reweighted with the truncated density ratio, the pre-training reweighted, and the pre-training reweighted with the truncated density ratio estimators on the class of H\"older smooth functions.
As previously pointed out, these estimators can adaptively achieve the minimax optimal rate.

\newpage 
% \newgeometry{left = 3cm, bottom = 0.5cm, top = 1cm, right = 1cm}

\begin{landscape}

    {  

\begin{table}[h!]
    \centering
    \tiny
    
  \caption{Convergence rates of the $L_2(\mathbb{Q}_{X_0})$ error bounds of the unweighted, reweighted and pre-training estimators. 
  It is assumed that $h_0 \in C^{s} \left( \mathcal{X},\mathcal{K}  \right)$
  for the unweighted and reweighted estimators $\widehat{h}_{\mathbb{D}} $, $\widehat{h}_{r,\mathbb{D}}$ and $\widehat{h}_{T_{\eta} r, \mathbb{D}} $ (with existence of (1+$\delta$)-order moment of $r$), and $h_0 \in C^{s_2} \left( \mathcal{X},\mathcal{K}  \right)$ in the case of the pre-training reweighted estimators $\widehat{h}_{\widehat{r}_{\mathbb{S}}, \mathbb{D} }$ (with  $r \in C^{s_1} \left( \mathcal{X},\mathcal{K}  \right)$) and $\widehat{h}_{\widehat{r}_{\eta, \mathbb{S}}, \mathbb{D} } $ (with  $T_\eta r \in C^{s_1} \left( \mathcal{X},\mathcal{K}  \right)$ and existence of (2+$\delta$)-order moment of $r$).
  }
  
  \medskip

  \medskip
  \label{table_of_some_examples}
\begin{tabular}{|c| >$c<$ | >$c<$ | >$c<$ | >$c<$ | >$c<$ | >$c<$ |}
\cline{3-7}
   \multicolumn{2}{c|}{} & \multicolumn{5}{|c|}{\parbox[c][1cm][c]{1cm}{\normalsize \text{Convergence rate}}} \\\hline 
     \shortstack{ \\ Dependence \\ structure} & \varphi (n) &  \mathbb{E}_{\mathbb{D}} \left[ \norm{\widehat{h}_{\mathbb{D}} - h_0 }^2_{2,\mathbb{Q}_{X_0}} \right] & \mathbb{E}_{\mathbb{D}} \left[ \norm{\widehat{h}_{r,\mathbb{D}} - h_0 }^2_{2,\mathbb{Q}_{X_0}} \right] & \mathbb{E}_{\mathbb{D}} \left[ \norm{\widehat{h}_{T_{\eta} r, \mathbb{D}} - h_0 }^2_{2,\mathbb{Q}_{X_0}} \right] & \mathbb{E}_{\mathbb{S}, \mathbb{D}} \left[ \norm{\widehat{h}_{\widehat{r}_{\mathbb{S}}, \mathbb{D} } - h_0 }^2_{2, \mathbb{Q}_{X_0}} \right] &   \mathbb{E}_{\mathbb{S}, \mathbb{D}} \left[ \norm{\widehat{h}_{\widehat{r}_{\eta, \mathbb{S}}, \mathbb{D} } - h_0 }^2_{2, \mathbb{Q}_{X_0}} \right] \\[0.3cm] \hline 
    % La ligne N°2 
     iid & n &   \mathcal{O} \left( n^{-\frac{2s_2}{2s_2 + d}} ( \log n)^3 \right) & \mathcal{O} \left( n^{-\frac{2s_2}{2s_2 + d}} ( \log n)^3 \right) & \mathcal{O} \left( n^{-\frac{\delta}{1+ \delta} \times \frac{2s_2}{2s_2 + d}} ( \log n)^3 \right) & \mathcal{O} \left(  \dfrac{\left( \log m \right)^{3}}{m^{\frac{2 s_1}{2 s_1 + d}}} +  \dfrac{\left( \log n \right)^{3}}{n^{\frac{2s_2}{2s_2 + d}}} \right) & \mathcal{O} \left(   \dfrac{\left( \log m \right)^{3}}{m^{\frac{2 s_1}{2 s_1 + d} \times\frac{\delta}{1+ \delta}}}  + \dfrac{\Big(\log n \Big)^{3} \times  m^{\frac{2 s_1}{(1+ \delta)(2 s_1 + d)}}}{n^{\frac{2s_2}{2s_2 + d}}} \right)  \\[0.3cm] \hline 
    % la ligne N°3 
    \shortstack{\\ $\phi$-mixing }  & n &   \mathcal{O} \left( n^{-\frac{2s_2}{2s_2 + d}} ( \log n)^3 \right) & \mathcal{O} \left( n^{-\frac{2s_2}{2s_2 + d}} ( \log n)^3 \right) & \mathcal{O} \left( n^{-\frac{\delta}{1+ \delta} \times \frac{2s_2}{2s_2 + d}} ( \log n)^3 \right) & \mathcal{O} \left(  \dfrac{\left( \log m \right)^{3}}{m^{\frac{2 s_1}{2 s_1 + d}}} +  \dfrac{\left( \log n \right)^{3}}{n^{\frac{2s_2}{2s_2 + d}}} \right) & \mathcal{O} \left(   \dfrac{\left( \log m \right)^{3}}{m^{\frac{2 s_1}{2 s_1 + d} \times\frac{\delta}{1+ \delta}}}  + \dfrac{\Big(\log n \Big)^{3} \times  m^{\frac{2 s_1}{(1+ \delta)(2 s_1 + d)}}}{n^{\frac{2s_2}{2s_2 + d}}} \right)  \\[0.3cm] \hline 
   \shortstack{ \\ Exponential \\ $\alpha$-mixing }   & \dfrac{n}{\left( \log n \right)^2} & \mathcal{O} \left( n^{-\frac{2s_2}{2s_2 + d}} ( \log n)^5 \right) & \mathcal{O} \left( n^{-\frac{2s_2}{2s_2 + d}} ( \log n)^5 \right) & \mathcal{O} \left( n^{-\frac{\delta}{1+ \delta} \times \frac{2s_2}{2s_2 + d}} ( \log n)^5 \right) & \mathcal{O} \left(  \dfrac{\left( \log m \right)^{5}}{m^{\frac{2 s_1}{2 s_1 + d}}} +  \dfrac{\left( \log n \right)^{5}}{n^{\frac{2s_2}{2s_2 + d}}} \right) & \mathcal{O} \left(   \dfrac{\left( \log m \right)^{5}}{m^{\frac{2 s_1}{2 s_1 + d} \times\frac{\delta}{1+ \delta}}}  + \dfrac{\Big(\log n \Big)^{5} \times  m^{\frac{2 s_1}{(1+ \delta)(2 s_1 + d)}}}{n^{\frac{2s_2}{2s_2 + d}}} \right) \\[0.3cm] \hline 
   \shortstack{ \\ Geoemtrically \\ $\mathcal{C}$-mixing }  & \dfrac{n}{\left( \log n \right)^{2/ \varrho}} & \mathcal{O} \left( n^{-\frac{2s_2}{2s_2 + d}} ( \log n)^{3 + \frac{2}{\varrho}} \right) & \mathcal{O} \left( n^{-\frac{2s_2}{2s_2 + d}} ( \log n)^{3 + \frac{2}{\varrho}} \right) &  \mathcal{O} \left( n^{-\frac{\delta}{1+ \delta} \times \frac{2s_2}{2s_2 + d}} ( \log n)^{3 + \frac{2}{\varrho}} \right) & \mathcal{O} \left(  \dfrac{\left( \log m \right)^{3 + \frac{2}{\varrho}}}{m^{\frac{2 s_1}{2 s_1 + d}}} +  \dfrac{\left( \log n \right)^{3 + \frac{2}{\varrho}}}{n^{\frac{2s_2}{2s_2 + d}}} \right) & \mathcal{O} \left(   \dfrac{\left( \log m \right)^{3 + \frac{2}{\varrho}}}{m^{\frac{2 s_1}{2 s_1 + d} \times\frac{\delta}{1+ \delta}}}  + \dfrac{\Big(\log n \Big)^{3 + \frac{2}{\varrho}} \times  m^{\frac{2 s_1}{(1+ \delta)(2 s_1 + d)}}}{n^{\frac{2s_2}{2s_2 + d}}} \right) \\[0.3cm] \hline 
\end{tabular} 
\end{table}
} % La fin avec la police de footnotescriptsize 
\end{landscape}

\section{Conclusion and outlook}\label{sec_conclusion}
In this work, we consider Huber and quantile nonparametric from dependent observations and under covariate shift, i.e. when the training and test datasets do not follow the same distribution.
For this purpose, the density ratio $r$ is used to quantify the shift between the distributions of the covariate over the source and the target datasets.
We perform both cases where this ratio is known and unknown.
Also, $r$ may be bounded or unbounded. In the latter case, a finite moment condition is considered.
We propose SPDNN unweighted, reweighted and two stages pre-training reweighted estimators and establish their non-asymptotic $L_2(\mathbb{Q}_{X_0})$ error bounds on the class of H\"older smooth functions. 
 For several classical models, including $\phi$-mixing, strong mixing, and $\mathcal{C}$-mixing processes, these estimators can adaptively achieve  (up to a logarithmic factor) the minimax optimal rate.
 
\medskip

This work can be extended in several directions; examples:
\begin{enumerate}
    \item We have performed nonparametric Huber and quantile regression.
    An interesting question is to consider a general deep learning framework that includes regression and classification under  covariate shift.
    This issue could be carried out through a general loss function, see for instance \cite{kengne2025robust}, \cite{kengne2025deep}, \cite{alquier2025minimax}.
   \item We have assumed that the target function $h_0$ belongs to the class of H\"older smooth functions.
   Another extension is to consider the classes of composition H\"older functions (see \cite{schmidt2020nonparametric}), and non-smooth functions (for instance, piecewise smooth functions, see \cite{petersen2018optimal}, \cite{imaizumi2019deep}, \cite{imaizumi2022advantage}).
   \item  One can increase the predictive performance of the proposed estimators by considering transfer learning under covariate shift. In this  extension, we deal with two samples: a large sample $\{(X_i, Y_i)\}_{i=1}^n$ with $(X_0, Y_0) \sim \mathbb{P}_{X_0,Y_0}$ (the source distribution), and a second sample $\{(X_i', Y_i')\}_{i=1}^k$ with $(X_0', Y_0') \sim \mathbb{Q}_{X_0,Y_0}$ (the target distribution), where $k \ll n$.
   The estimators above will be constructed from these two samples in order to minimize the $L_2(\mathbb{Q}_{X_0})$ error.
   The case examined in the present study corresponds to $k = 0$. 
   Such issue has been addressed for regression (see for instance \cite{schmidt2024local}, \cite{zamolodtchikov2026minimax}) and for classification (see \cite{reeve2021adaptive}, \cite{jiao2024deep}) from i.i.d. data. 
   Such an approach with $k > 0$ could be performed in the framework considered here, including dependent observations and adaptive estimators. 
\end{enumerate}

\section{Proofs of the main results}
\label{sec_proofs}

\subsection{Proof of Lemma \ref{lemma_decomposition_estimator}}
    Firstly, let us deal with the case of the Huber regression. 
Let $\tau >0$.
Recall that, the Huber loss $\ell_\tau (u) = \frac{1}{2}u^2 \ind(|u| \leq \tau) + \big( \tau |u| - \frac{1}{2} \tau^2 \big) \ind(|u| > \tau)  $
is differentiable and the derivative is given for all $x \in \R$ by,
\[ \ell'_\tau(X_0) = \sgn(x) (|x| \wedge \tau)  .\]
This score function is Lipschitz continuous and has a derivative for almost everywhere $x \in \R$ given by,
\[  \ell''_\tau(X_0) = \ind(|x| \leq \tau) .\]
So, on can write (from the fundamental theorem of calculus) for any $u, v \in \R$,
\begin{equation}
\label{fundamental_calculus_huber}
    \ell_\tau(u+v) - \ell_\tau(u) = \ell'_\tau(u) v + \int_0^v \ell''_\tau(u+z)(v-z) dz . 
\end{equation}
Let $h \in \mathcal{H}_{\sigma,n}$. By taking $u=Y_0 - h_{\mathcal{H}_{\sigma,n}}(X_0)$ and $v= h_{\mathcal{H}_{\sigma,n}}(X_0) - h(X_0)$, we get,
\begin{align}\label{proof_lem_unwei_est_ell_tau_int}
\nonumber &\ell_\tau \big( Y_0 - h(X_0)  \big) - \ell_\tau \big( Y_0 - h_{\mathcal{H}_{\sigma,n}}(X_0) \big) = \ell'_\tau\big( Y_0 - h_{\mathcal{H}_{\sigma,n}}(X_0) \big)  \big( h_{\mathcal{H}_{\sigma,n}}(X_0) - h(X_0) \big) \\
\nonumber &+ \int_0^{h_{\mathcal{H}_{\sigma,n}}(X_0) - h(X_0)} \ell''_\tau \big( Y_0 - h_{\mathcal{H}_{\sigma,n}}(X_0) +  z \big) \big( h_{\mathcal{H}_{\sigma,n}}(X_0) - h(X_0) -z \big) dz \\
\nonumber & =  \big( h_{\mathcal{H}_{\sigma,n}}(X_0) - h(X_0) \big)  \ell'_\tau\big( Y_0 - h_{0}(X_0) \big) +  \big( h_{\mathcal{H}_{\sigma,n}}(X_0) - h(X_0) \big)   \big[ \ell'_\tau\big( Y_0 - h_{\mathcal{H}_{\sigma,n}}(X_0) \big) - \ell'_\tau\big( Y_0 - h_{0}(X_0) \big)\big] \\
\nonumber &+ \int_0^{h_{\mathcal{H}_{\sigma,n}}(X_0) - h(X_0)} \ell''_\tau \big( Y_0 - h_{\mathcal{H}_{\sigma,n}}(X_0) +  z \big) \big( h_{\mathcal{H}_{\sigma,n}}(X_0) - h(X_0) -z \big) dz  \\
\nonumber & =  \big( h_{\mathcal{H}_{\sigma,n}}(X_0) - h(X_0) \big)  \ell'_\tau\big( \xi_0 \big) +  \big( h_{\mathcal{H}_{\sigma,n}}(X_0) - h(X_0) \big)   \big[ \ell'_\tau\big( Y_0 - h_{\mathcal{H}_{\sigma,n}}(X_0) \big) - \ell'_\tau\big( Y_0 - h_{0}(X_0) \big)\big] \\
  &+ \int_0^{h_{\mathcal{H}_{\sigma,n}}(X_0) - h(X_0)} \ell''_\tau \big( Y_0 - h_{\mathcal{H}_{\sigma,n}}(X_0) +  z \big) \big( h_{\mathcal{H}_{\sigma,n}}(X_0) - h(X_0) -z \big) dz .   
\end{align}
From the proof of Proposition 3.1 in \cite{fan2024noise}, we have for all $x \in \mx$, $\E[\ell'_\tau(\xi_0) | X_0=x] =0$. Hence, $\E\big[  \big( h_{\mathcal{H}_{\sigma,n}}(X_0) - h(X_0) \big)  \ell'_\tau\big( \xi_0 \big)  \big] =0  $.
Moreover, the function $\ell'_\tau$ is 1-Lipschitzian. So,
$ \big| \ell'_\tau\big( Y_0 - h_{\mathcal{H}_{\sigma,n}}(X_0) \big) - \ell'_\tau\big( Y_0 - h_{0}(X_0) \big) \big| \leq \big| h_{\mathcal{H}_{\sigma,n}}(X_0) - h_{0}(X_0) \big|$.
Therefore,
\begin{multline}\label{proof_lem_unwei_est_ep_geq}
\E_{(X_0,Y_0) \sim P_{X_0, Y_0}} \Big[ \big( h_{\mathcal{H}_{\sigma,n}}(X_0) - h(X_0) \big)   \big[ \ell'_\tau\big( Y_0 - h_{\mathcal{H}_{\sigma,n}}(X_0) \big) - \ell'_\tau\big( Y_0 - h_{0}(X_0) \big)\big] \Big] \\
  \geq - \E_{X_0 \sim \mathbb{P}_{X_0}} \big[ \big| h_{\mathcal{H}_{\sigma,n}}(X_0) - h(X_0) \big| \big| h_{\mathcal{H}_{\sigma,n}}(X_0) - h_0(X_0) \big|  \big].
\end{multline}
Now, let $x \in \mx$. We have,
\begin{align*}
&\E\Big[ \int_0^{h_{\mathcal{H}_{\sigma,n}}(X_0) - h(X_0)} \ell''_\tau \big( Y_0 - h_{\mathcal{H}_{\sigma,n}}(X_0) +  z \big) \big( h_{\mathcal{H}_{\sigma,n}}(X_0) - h(X_0) -z \big) dz  \big| X_0=x\Big] \\
&= \E\Big[ \int_0^{h_{\mathcal{H}_{\sigma,n}}(X_0) - h(X_0)} \ind\big(|Y_0 - h_{\mathcal{H}_{\sigma,n}}(X_0) +  z | \leq \tau \big) \big( h_{\mathcal{H}_{\sigma,n}}(X_0) - h(X_0) -z \big) dz  \big| X_0=x\Big] \\
&= \E\Big[ \int_0^{h_{\mathcal{H}_{\sigma,n}}(X_0) - h(X_0)} \Big( 1 - \ind\big(|Y_0 - h_{\mathcal{H}_{\sigma,n}}(X_0) +  z | > \tau \big) \Big) \big( h_{\mathcal{H}_{\sigma,n}}(X_0) - h(X_0) -z \big) dz  \big| X_0=x\Big] \\
&= \E\Big[ \int_0^{h_{\mathcal{H}_{\sigma,n}}(X_0) - h(X_0)} \Big( 1 - \ind\big(|\xi_0 + h_0(X_0)- h_{\mathcal{H}_{\sigma,n}}(X_0) +  z | > \tau \big) \Big) \big( h_{\mathcal{H}_{\sigma,n}}(X_0) - h(X_0) -z \big) dz  \big| X_0=x\Big] \\
&= \E\Big[ \int_0^{h_{\mathcal{H}_{\sigma,n}}(x) - h(x)} \Big( 1 - \ind\big(|\xi_0 + h_0(x)- h_{\mathcal{H}_{\sigma,n}}(x) +  z | > \tau \big) \Big) \big( h_{\mathcal{H}_{\sigma,n}}(x) - h(x) -z \big) dz  \big| X_0=x\Big] \\
&= \int_0^{h_{\mathcal{H}_{\sigma,n}}(x) - h(x)}  \big( h_{\mathcal{H}_{\sigma,n}}(x) - h(x) -z \big) dz \\ 
& - \E\Big[ \int_0^{h_{\mathcal{H}_{\sigma,n}}(x) - h(x)}  \ind\big(|\xi_0 + h_0(x) - h_{\mathcal{H}_{\sigma,n}}(x) +  z | > \tau \big)  \big( h_{\mathcal{H}_{\sigma,n}}(x) - h(x) -z \big) dz  \big| X_0=x\Big] 
\end{align*}
\begin{align}
 \label{proof_lem_unwei_est_ep_geq_square2} &= \dfrac{1}{2} \big( h_{\mathcal{H}_{\sigma,n}}(x) - h(x) \big)^2 - \E\Big[ \int_0^{h_{\mathcal{H}_{\sigma,n}}(x) - h(x)}  \ind\big(|\xi_0 + h_0(x) - h_{\mathcal{H}_{\sigma,n}}(x) +  z | > \tau \big)  \big( h_{\mathcal{H}_{\sigma,n}}(x) - h(x) -z \big) dz  \big| X_0=x\Big]  \\
\nonumber &\geq \dfrac{1}{2} \big( h_{\mathcal{H}_{\sigma,n}}(x) - h(x) \big)^2  \\
\nonumber & - \E\Big[ \int_0^{h_{\mathcal{H}_{\sigma,n}}(x) - h(x)} \Big( \ind\big(|\xi_0| > \tau/3 \big) + \ind\big( | h_0(x) - h_{\mathcal{H}_{\sigma,n}}(x) | > \tau/3 \big) +  \ind\big( |h_{\mathcal{H}_{\sigma,n}}(x) - h(x)| > \tau/3 \big) \Big) \\
\nonumber  & \hspace{12cm} \times \big( h_{\mathcal{H}_{\sigma,n}}(x) - h(x) -z \big) dz  \big| X_0=x\Big] \\
\nonumber &= \dfrac{1}{2} \big( h_{\mathcal{H}_{\sigma,n}}(x) - h(x) \big)^2  - \Big( P\big( |\xi_0| > \tau/3 | X_0=x \big) +  \ind\big( | h_0(x) - h_{\mathcal{H}_{\sigma,n}}(x) | > \tau/3 \big) +  \ind\big( |h_{\mathcal{H}_{\sigma,n}}(x) - h(x)| > \tau/3 \big)  \Big) \\
\nonumber & \hspace{10.5cm}  \times  \int_0^{h_{\mathcal{H}_{\sigma,n}}(x) - h(x)} \big( h_{\mathcal{H}_{\sigma,n}}(x) - h(x) -z \big) dz \\ 
\nonumber &= \dfrac{1}{2} \big( h_{\mathcal{H}_{\sigma,n}}(x) - h(x) \big)^2 \Big(1 - P\big( |\xi_0| > \tau/3 | X_0=x \big) - \ind\big( | h_0(x) - h_{\mathcal{H}_{\sigma,n}}(x) | > \tau/3 \big) -  \ind\big( |h_{\mathcal{H}_{\sigma,n}}(x) - h(x)| > \tau/3 \big) \Big) \\
\label{proof_lem_unwei_est_ep_geq_square} &=\dfrac{1}{2} \big( h_{\mathcal{H}_{\sigma,n}}(x) - h(x) \big)^2 \big[1 - P\big( |\xi_0| > \tau/3 | X_0=x \big) \big],
\end{align}
where the last equality holds from $ | h_0(x) - h_{\mathcal{H}_{\sigma,n}}(x) | \leq \| h_0 \|_{\infty, \mathcal{X}} + \| h_{\mathcal{H}_{\sigma,n}} \|_{\infty, \mathcal{X}} \leq \mk + F \leq \tau/3$ and 
$ | h_{\mathcal{H}_{\sigma,n}}(x) - h(x)| \leq  \| h_{\mathcal{H}_{\sigma,n}} \|_{\infty, \mathcal{X}} + \| h \|_{\infty, \mathcal{X}} \leq 2F \leq \tau/3$ since $\tau \geq 6 (\mk \vee F) $.
We get from the Markov's inequality,
\[ P\big( |\xi_0| > \tau/3 | X_0=x \big) \leq \dfrac{ \E(|\xi_0| | X_0=x )}{ \tau /3} \leq 3 C_\xi /\tau .  \]
 Hence, (\ref{proof_lem_unwei_est_ep_geq_square}) gives,
 \begin{equation}\label{proof_lem_unwei_est_ep_geq_square2}
 \E\Big[ \int_0^{h_{\mathcal{H}_{\sigma,n}}(X_0) - h(X_0)} \ell''_\tau \big( Y_0 - h_{\mathcal{H}_{\sigma,n}}(X_0) +  z \big) \big( h_{\mathcal{H}_{\sigma,n}}(X_0) - h(X_0) -z \big) dz  \big| X_0=x\Big] \geq \dfrac{1}{2} \big(1 - \dfrac{3 C_\xi}{\tau} \big) \big( h_{\mathcal{H}_{\sigma,n}}(x) - h(x) \big)^2. 
 \end{equation}
Recall that $1 - \dfrac{3 C_\xi}{\tau} > 0$ since $\tau > 3 C_\xi$.   
Thus, in addition to (\ref{proof_lem_unwei_est_ell_tau_int}), (\ref{proof_lem_unwei_est_ep_geq}) and from the Cauchy-Schwarz inequality, we get for all $\beta >0$,
\begin{align*}
& \E_{(X_0,Y_0) \sim P_{X_0, Y_0}} \big[ \ell_\tau \big(Y_0 - h(X_0) \big) - \ell_\tau \big(Y_0 - h_{\mathcal{H}_{\sigma,n}}(X_0) \big)  \big] \\
 &\geq  - \E_{X_0 \sim \mathbb{P}_{X_0}} \big[ \big| h_{\mathcal{H}_{\sigma,n}}(X_0) - h(X_0) \big| \big| h_{\mathcal{H}_{\sigma,n}}(X_0) - h_0(X_0) \big|  \big] 
 + \dfrac{1}{2} \big(1 - \dfrac{3 C_\xi}{\tau} \big) \E_{X_0 \sim \mathbb{P}_{X_0}} \big[ \big( h_{\mathcal{H}_{\sigma,n}}(X_0) - h(X_0) \big)^2 \big] \\
 & \geq  -\sqrt{\E_{X_0 \sim \mathbb{P}_{X_0}} \big[ \big( h_{\mathcal{H}_{\sigma,n}}(X_0) - h(X_0) \big)^2  \big] }
 \sqrt{\E_{X_0 \sim \mathbb{P}_{X_0}} \big[ \big( h_{\mathcal{H}_{\sigma,n}}(X_0) - h_0(X_0) \big)^2  \big] } \\
 & \hspace{10cm} + \dfrac{1}{2} \big(1 - \dfrac{3 C_\xi}{\tau} \big) \E_{X_0 \sim \mathbb{P}_{X_0}} \big[ \big( h_{\mathcal{H}_{\sigma,n}}(X_0) - h(X_0) \big)^2 \big] \\
  & \geq  -\dfrac{1}{2 \beta } \E_{X_0 \sim \mathbb{P}_{X_0}} \big[ \big( h_{\mathcal{H}_{\sigma,n}}(X_0) - h(X_0) \big)^2  \big]   
- \dfrac{\beta}{2}\E_{X_0 \sim \mathbb{P}_{X_0}} \big[ \big( h_{\mathcal{H}_{\sigma,n}}(X_0) - h_0(X_0) \big)^2  \big]  \\
 & \hspace{10cm} + \dfrac{1}{2} \big(1 - \dfrac{3 C_\xi}{\tau} \big) \E_{X_0 \sim \mathbb{P}_{X_0}} \big[ \big( h_{\mathcal{H}_{\sigma,n}}(X_0) - h(X_0) \big)^2 \big] \\
  & =  \dfrac{1}{2 } \Big( \big(1 - \dfrac{3 C_\xi}{\tau} \big) - \dfrac{1}{\beta} \Big) \|  h_{\mathcal{H}_{\sigma,n}} - h  \|_{2,\mathbb{P}_{X_0}}^2     
- \dfrac{\beta}{2} \|  h_{\mathcal{H}_{\sigma,n}} - h_0  \|_{2,\mathbb{P}_{X_0}}^2. 
\end{align*}
Hence,
\[
\dfrac{1}{2 } \Big( \big(1 - \dfrac{3 C_\xi}{\tau} \big) - \dfrac{1}{\beta} \Big)  \|  h_{\mathcal{H}_{\sigma,n}} - h  \|_{2,\mathbb{P}_{X_0}}^2   
\leq \E_{(X_0,Y_0) \sim P_{X_0, Y_0}} \big[ \ell_\tau \big(Y_0 - h(X_0) \big) - \ell_\tau \big(Y_0 - h_{\mathcal{H}_{\sigma,n}}(X_0) \big)  \big]  
  +  \dfrac{\beta}{2} \|  h_{\mathcal{H}_{\sigma,n}} - h_0  \|_{2,\mathbb{P}_{X_0}}^2. 
\]
So,
\begin{align}\label{proof_lem_unwei_est_norm_L2}
\nonumber & \| h - h_0  \|_{2,\mathbb{P}_{X_0}}^2  \leq 2  \|  h_{\mathcal{H}_{\sigma,n}} - h  \|_{2,\mathbb{P}_{X_0}}^2 + 2 \|  h_{\mathcal{H}_{\sigma,n}} - h_0  \|_{2,\mathbb{P}_{X_0}}^2 \\
\nonumber &\leq \dfrac{4}{   \big(1 - \dfrac{3 C_\xi}{\tau} \big) - \dfrac{1}{\beta}  } \Big(  \E_{(X_0,Y_0) \sim P_{X_0, Y_0}} \big[ \ell_\tau \big(Y_0 - h(X_0) \big) - \ell_\tau \big(Y_0 - h_{\mathcal{H}_{\sigma,n}}(X_0) \big)  \big] + \dfrac{\beta}{2} \|  h_{\mathcal{H}_{\sigma,n}} - h_0  \|_{2,\mathbb{P}_{X_0}}^2  \Big) + 2 \|  h_{\mathcal{H}_{\sigma,n}} - h_0  \|_{2,\mathbb{P}_{X_0}}^2 \\
&=  \dfrac{4}{   \big(1 - \dfrac{3 C_\xi}{\tau} \big) - \dfrac{1}{\beta}  }   \E_{(X_0,Y_0) \sim P_{X_0, Y_0}} \big[ \ell_\tau \big(Y_0 - h(X_0) \big) - \ell_\tau \big(Y_0 - h_{\mathcal{H}_{\sigma,n}}(X_0) \big)  \big]
 + \Bigg( \dfrac{2 \beta }{   \big(1 - \dfrac{3 C_\xi}{\tau} \big) - \dfrac{1}{\beta}  } + 2 \Bigg)  \|  h_{\mathcal{H}_{\sigma,n}} - h_0  \|_{2,\mathbb{P}_{X_0}}^2 .     
\end{align}
This inequality holds for all $\beta >0$ and $h \in \mathcal{H}_{\sigma,n}$.
 So, set 
 \[ \beta = 2, ~ C_1 = \dfrac{4}{ \frac{1}{2}  - \dfrac{3 C_\xi}{\tau}  } ~ \text{ and } C_2 = C_1 +2 .\]  
We have  $C_1 >0$, since $\tau > 6 C_\xi$. It holds from (\ref{proof_lem_unwei_est_norm_L2}) that,
\begin{align}\label{proof_lem_unwei_est_norm_L2_inf}
 \nonumber \| h- h_0  \|_{2,\mathbb{P}_{X_0}}^2 
 & \leq C_1 \E_{(X_0,Y_0) \sim P_{X_0, Y_0}} \big[ \ell_\tau \big(Y_0 - h(X_0) \big) - \ell_\tau \big(Y_0 - h_{\mathcal{H}_{\sigma,n}}(X_0) \big)  \big] + C_2 \|  h_{\mathcal{H}_{\sigma,n}} - h_0  \|_{2,\mathbb{P}_{X_0}}^2 \\
\nonumber & \leq C_1 \E_{(X_0,Y_0) \sim P_{X_0, Y_0}} \big[ \ell_\tau \big(Y_0 - h(X_0) \big) - \ell_\tau \big(Y_0 - h_{\mathcal{H}_{\sigma,n}}(X_0) \big)  \big] + C_2 \|  h_{\mathcal{H}_{\sigma,n}} - h_0  \|_{\infty, \mx}^2 \\
&= C_1 \E_{(X_0,Y_0) \sim P_{X_0, Y_0}} \big[ \ell_\tau \big(Y_0 - h(X_0) \big) - \ell_\tau \big(Y_0 - h_{\mathcal{H}_{\sigma,n}}(X_0) \big)  \big] +C_2 \inf_{h \in \mathcal{H}_{\sigma,n}} \| h - h_0\|^2_{\infty, \mx}. 
\end{align}
The case of the quantile regression can be performed as in Lemma 4 in \cite{feng2024deep}. 
This completes the proof of the lemma. 
 \qed

 \subsection{Proof of Lemma \ref{lemma_decomposition_estimator_reweighted}}
We will perform the proof for the Huber regression.
The case of the quantile regression can be obtained in the same way as in the proof of Lemma 21 in \cite{feng2024deep}.\\ 
\noindent
Let $h \in \mathcal{H}_{\sigma,n}$. 
We get from (\ref{proof_lem_unwei_est_ell_tau_int}),
\begin{align}\label{proof_lem_unwei_rX0_delta_ell_tau}
  \nonumber   &r(X_0) \left( \ell_\tau \big( Y_0 - h(X_0)  \big) - \ell_\tau \big( Y_0 - h_{\mathcal{H}_{\sigma,n}}(X_0) \big) \right) \\ 
 \nonumber    &= \left( r(X_0)\big( h_{\mathcal{H}_{\sigma,n}}(X_0) - h(X_0) \big)  \ell'_\tau\big( \xi_0 \big) \right)     +    r(X_0) \big( h_{\mathcal{H}_{\sigma,n}}(X_0) - h(X_0) \big)  \big[ \ell'_\tau\big( Y_0 - h_{\mathcal{H}_{\sigma,n}}(X_0) \big) - \ell'_\tau\big( Y_0 - h_{0}(X_0) \big)\big] \\
 &  \hspace{3cm} + r(X_0)\int_0^{h_{\mathcal{H}_{\sigma,n}}(X_0) - h(X_0)} \ell''_\tau \big( Y_0 - h_{\mathcal{H}_{\sigma,n}}(X_0) +  z \big) \big( h_{\mathcal{H}_{\sigma,n}}(X_0) - h(X_0) -z \big) dz. 
\end{align}
From the same arguments as in the proof of Lemma \ref{lemma_decomposition_estimator}, we get, 
$\mathbb{E} \left[\left( r(X_0)\big( h_{\mathcal{H}_{\sigma,n}}(X_0) - h(X_0) \big)  \ell'_\tau\big( \xi_0 \big) \right) \right] =0$, for all $\beta >0$,
\begin{align} 
  \nonumber & \mathbb{E}_{(X_0, Y_0) \sim \mathbb{P}_{X_0, Y_0}}    \left[ \left( r(X_0) \big( h_{\mathcal{H}_{\sigma,n}}(X_0) - h(X_0) \big) \right)  \big[ \ell'_\tau\big( Y_0 - h_{\mathcal{H}_{\sigma,n}}(X_0) \big) - \ell'_\tau\big( Y_0 - h_{0}(X_0) \big)\big] \right] \\
  \label{proof_lem_unwei_est_ep_geq_Q_X0}  & \geq - \mathbb{E}_{X_0 \sim \mathbb{P}_{X_0}} \left[ r(X_0)   \lvert h_{\mathcal{H}_{\sigma,n}}(X_0) - h(X_0)  \rvert \times \lvert h_{\mathcal{H}_{\sigma,n}}(X_0) - h_0(X_0)  \rvert  \right] \\
 \label{proof_lem_unwei_est_ep_X0_Cauchy_Schwarz}  & \geq - \sqrt{\mathbb{E}_{X_0 \sim \mathbb{P}_{X_0}} \left[ r(X_0)   \big( h_{\mathcal{H}_{\sigma,n}}(X_0) - h(X_0) \big)^2 \right] }
   \sqrt{\mathbb{E}_{X_0 \sim \mathbb{P}_{X_0}}  \left[ r(X_0) \big( h_{\mathcal{H}_{\sigma,n}}(X_0) - h_0(X_0)  \big)^2  \right] } \\
   \label{proof_lem_unwei_est_ep_geq_Q_X0_beta} & \geq - \frac{1}{2 \beta} \mathbb{E}_{X_0 \sim \mathbb{P}_{X_0}} \left[ r(X_0)   \big( h_{\mathcal{H}_{\sigma,n}}(X_0) - h(X_0) \big)^2 \right]  
   - \frac{\beta}{2} \mathbb{E}_{X_0 \sim \mathbb{P}_{X_0}}  \left[ r(X_0) \big( h_{\mathcal{H}_{\sigma,n}}(X_0) - h_0(X_0)  \big)^2  \right]  
\end{align}
and for all $x \in \mx$,
\begin{align}\label{proof_lem_unwei_est_ep_geq_Q_X0_2}
 \nonumber   \mathbb{E} & \left[ r(X_0)\int_0^{h_{\mathcal{H}_{\sigma,n}}(X_0) - h(X_0)} \ell''_\tau \big( Y_0 - h_{\mathcal{H}_{\sigma,n}}(X_0) +  z \big) \big( h_{\mathcal{H}_{\sigma,n}}(X_0) - h(X_0) -z \big) dz \Big | X_0 = x \right] \\
    &\geq \dfrac{1}{2} \left(1 - \dfrac{3 C_{\xi}}{\tau} \right) \left( h_{\mathcal{H}_{\sigma,n}}(x) - h(x) \right)^2 r(x).
\end{align}
The inequalities in (\ref{proof_lem_unwei_est_ep_geq_Q_X0}), (\ref{proof_lem_unwei_est_ep_X0_Cauchy_Schwarz}) and (\ref{proof_lem_unwei_est_ep_geq_Q_X0_2}) hold by going as in (\ref{proof_lem_unwei_est_ep_geq}), (\ref{proof_lem_unwei_est_ep_geq_square2}) and by using the Cauchy-Schwarz inequality.
Hence, according to (\ref{proof_lem_unwei_rX0_delta_ell_tau}), (\ref{proof_lem_unwei_est_ep_geq_Q_X0_beta}) and (\ref{proof_lem_unwei_est_ep_geq_Q_X0_2}), we have for all $\beta >0$,
\begin{align}\label{proof_lem_unwei_rX0_delta_ell_ineq}
  \nonumber & \mathbb{E}_{(X_0, Y_0) \sim \mathbb{P}_{X_0, Y_0}}  \left[ r(X_0) \left( \ell_\tau \big( Y_0 - h(X_0)  \big) - \ell_\tau \big( Y_0 - h_{\mathcal{H}_{\sigma,n}}(X_0) \big) \right) \right] \\
  \nonumber  & \geq \dfrac{1}{2} \left( \left(1 - \dfrac{3 C_{\xi}}{\tau} \right) - \dfrac{1}{\beta} \right) \mathbb{E}_{X_0 \sim \mathbb{P}_{X_0}} \left[  r(X_0) \left( h_{\mathcal{H}_{\sigma,n}}(X_0) - h(X_0) \right)^2 \right] - \dfrac{\beta}{2} \mathbb{E}_{X_0 \sim \mathbb{P}_{X_0}} \left[ r(X_0)  \left( h_{\mathcal{H}_{\sigma,n}}(X_0) - h_0(X_0)\right)^2 \right] \\
  & \geq \dfrac{1}{2} \left( \left(1 - \dfrac{3 C_{\xi}}{\tau} \right) - \dfrac{1}{\beta} \right) \norm{h_{\mathcal{H}_{\sigma,n}} - h}_{2, \mathbb{Q}_{X_0}}^2 - \dfrac{\beta}{2} \norm{h_{\mathcal{H}_{\sigma,n}} - h_0}_{2, \mathbb{Q}_{X_0}}^2,  
\end{align}
where the inequality (\ref{proof_lem_unwei_rX0_delta_ell_ineq}) holds since
\begin{multline*}
     \mathbb{E}_{X_0 \sim \mathbb{P}_{X_0}} \left[  r(X_0) \left( h_{\mathcal{H}_{\sigma,n}}(X_0) - h(X_0) \right)^2 \right] =  \norm{h_{\mathcal{H}_{\sigma,n}} - h}_{2, \mathbb{Q}_{X_0}}^2 \;  \text{and}  \\
      \mathbb{E}_{X_0 \sim \mathbb{P}_{X_0}} \left[  r(X_0) \left( h_{\mathcal{H}_{\sigma,n}}(X_0) - h_0(X_0) \right)^2 \right] =  \norm{h_{\mathcal{H}_{\sigma,n}} - h_0}_{2, \mathbb{Q}_{X_0}}^2.
\end{multline*}
By taking $\beta = 2$ and set $C_1 = \dfrac{4}{1/2 - 3C_{\xi}/\tau}, \; C_2 = C_1 + 2$ in (\ref{proof_lem_unwei_rX0_delta_ell_ineq}) and by going as in (\ref{proof_lem_unwei_est_norm_L2}), we get,
\begin{align*}
 \norm{h - h_0}_{2, \mathbb{Q}_{X_0}}^2  & \leq 2 \norm{h_{\mathcal{H} _{\sigma,n}} - h }_{2, \mathbb{Q}_{X_0}}^2 + 2 \norm{h_{\mathcal{H} _{\sigma,n}} - h_0 }_{2, \mathbb{Q}_{X_0}}^2  \\
 & \leq C_1 \mathbb{E}_{(X_0, Y_0) \sim \mathbb{P}_{X_0, Y_0}}  \left[ r(X_0) \left( \ell_\tau \big( Y_0 - h(X_0)  \big) - \ell_\tau \big( Y_0 - h_{\mathcal{H}_{\sigma,n}}(X_0) \big) \right) \right] +    C_2 \norm{h_{\mathcal{H} _{\sigma,n}} - h_0 }_{2, \mathbb{Q}_{X_0}}^2. 
\end{align*}
Hence, it holds as in (\ref{proof_lem_unwei_est_norm_L2_inf}) that,
\[ \norm{h- h_0}_{2, \mathbb{Q}_{X_0}}^2 
 \leq C_1  \mathbb{E}_{(X_0, Y_0) \sim \mathbb{P}_{X_0, Y_0}}  \left[ r(X_0) \left( \ell_\tau \big( Y_0 - h(X_0)  \big) - \ell_\tau \big( Y_0 - h_{\mathcal{H}_{\sigma,n}}(X_0) \big) \right) \right]  + C_2 \inf_{h \in \mathcal{H}_{\sigma,n}} \| h - h_0\|^2_{\infty, \mx}, \]
which completes the proof.
\qed

\subsection{Proof of Theorem \ref{unweighted_SPDNN_result}}
In this proof, we denote 

\[
 R_{\mathbb{P}} (h) = \mathbb{E}_{(X_0,Y_0) \sim \mathbb{P}_{X_0,Y_0}} \left[ \ell_\tau \left( h(X_0) , Y_0 \right) \right]. 
\]
\subsubsection*{1. Under \textbf{(A2)}}
Let $\nu >3$. On the one hand, by applying Lemma \ref{lemma_decomposition_estimator}, we have 
\begin{equation}\label{proof_unweighted_L2_error}
      \mathbb{E}_{\mathbb{D}} \left[ \norm{\widehat{h}_{\mathbb{D}} - h_0 }^2_{2,\mathbb{P}_{X_0}} \right] \lesssim  \mathbb{E}_{\mathbb{D}} \left\{  \mathbb{E}_{(X_0, Y_0) \sim \mathbb{P}_{X_0,Y_0}} \left[ \ell_{\tau} \left( Y_0 - \widehat{h}_\mathbb{D} (X_0) \right) - \ell_{\tau} \left( Y_0 - h_{\mathcal{H}_{\sigma,n}} (X_0) \right)\right] \right\} + \inf_{h \in \mathcal{H}_{\sigma,n}} \norm{h - h_0}^2_{\infty, \mathcal{X}}, 
\end{equation}
where $h_{\mathcal{H}_{\sigma,n}}$ is given in (\ref{def_h_H_sigma}). 
Note that, $h_0$ is a target function, see (\ref{h0_target_function}) and we have for all $h \in \mathcal{H}_{\sigma,n}$, 
$ R_{\mathbb{P}} (h) - R_{\mathbb{P}} (h_{\mathcal{H}_{\sigma,n}}) \leq R_{\mathbb{P}} (h) -   R_{\mathbb{P}} (h_0)$.
Hence,
\begin{align}\label{proof_unweighted_estimation_error}
  \nonumber &  \mathbb{E}_{\mathbb{D}} \left\{ \mathbb{E}_{(X_0, Y_0) \sim \mathbb{P}_{X_0,Y_0}} \left[ \ell_{\tau} \left( Y_0 - \widehat{h}_\mathbb{D} (X_0) \right) - \ell_{\tau} \left( Y_0 - h_{\mathcal{H}_{\sigma,n}} (X_0) \right)\right] \right\} \\
  &= \mathbb{E}_{\mathbb{D}} \left[ R_{\mathbb{P}} (\widehat{h}_{\mathbb{D}}) - R_{\mathbb{P}} (h_{\mathcal{H}_{\sigma,n}}) \right] 
  \leq \mathbb{E}_{\mathbb{D}} \left[ R_{\mathbb{P}} (\widehat{h}_{\mathbb{D}}) - R_{\mathbb{P}} (h_0) \right]  \lesssim \dfrac{\Big(\log \varphi(n) \Big)^{\nu}}{\Big( \varphi (n) \Big)^{\frac{2s}{2s + d}}}, 
\end{align}
where the last inequality above holds from Corollary 3.4 in \cite{kengne2025general}.
 According to  Theorem 3.2 in \cite{kengne2025excess}, we have
    \begin{equation}\label{proof_approx_DNN}
       % \label{unweighted_estimation_approximation}
        \inf_{h \in \mathcal{H}_{\sigma,n}} \norm{h - h_0}^2_{\infty, \mathcal{X}} \leq \dfrac{1}{\Big( \varphi (n) \Big)^{\frac{2s}{2s + d}}}.
    \end{equation}
In addition to (\ref{proof_unweighted_L2_error}) and (\ref{proof_unweighted_estimation_error}), we get 
 \begin{equation}\label{proof_unweighted_L2_error_PX0_rate}
 \mathbb{E}_{\mathbb{D}} \left[ \norm{\widehat{h}_{\mathbb{D}} - h_0 }^2_{2,\mathbb{P}_{X_0}} \right] \lesssim \dfrac{1}{\Big( \varphi (n) \Big)^{\frac{2s}{2s + d}}} + \dfrac{\Big(\log \varphi(n) \Big)^{\nu}}{\Big( \varphi (n) \Big)^{\frac{2s}{2s + d}}} \lesssim \dfrac{\Big(\log \varphi(n) \Big)^{\nu}}{\Big( \varphi (n) \Big)^{\frac{2s}{2s + d}}}. 
\end{equation}
On the other hand, we have for all $h \in \mathcal{H}_{\sigma,n}$,  
\begin{equation}\label{proof_unweighted_L2_error_Q_P}
    \| h - h_0 \|^{2}_{2, \mathbb{Q}_{X_0}} = \mathbb{E}_{X_0 \sim \mathbb{Q}_{X_0}} \left[ \left( h(X_0) - h_0(X_0) \right)^2 \right] = \mathbb{E}_{X_0 \sim\mathbb{P}_{X_0}} \left[ r(X_0) \left( h(X_0) - h_0(X_0) \right)^2 \right] \leq \Gamma \| h - h_0 \|_{2,\mathbb{P}_{X_0}}^2.
\end{equation}
Then, we have 
 \[ \mathbb{E}_{\mathbb{D}} \left[ \| \widehat{h}_{\mathbb{D}} - h_0 \|^{2}_{2, \mathbb{Q}_{X_0}} \right]  \leq \Gamma  \mathbb{E}_{\mathbb{D}} \left[ \|  \widehat{h}_{\mathbb{D}} - h_0 \|^{2}_{2,\mathbb{P}_{X_0}} \right].  \]
Hence, it follows from (\ref{proof_unweighted_L2_error_PX0_rate}) that,
\[ \mathbb{E}_{\mathbb{D}} \left[ \| \widehat{h}_{\mathbb{D}} - h_0 \|^{2}_{2, \mathbb{Q}_{X_0}} \right]  \leq \Gamma  \mathbb{E}_{\mathbb{D}} \left[ \|  \widehat{h}_{\mathbb{D}} - h_0 \|^{2}_{2,\mathbb{P}_{X_0}} \right]  \lesssim  \dfrac{\Big(\log \varphi(n) \Big)^{\nu}}{\Big( \varphi (n) \Big)^{\frac{2s}{2s + d}}}.
\]

\subsubsection*{2. Under \textbf{(A3)} with $\mu=2$}  
 Let $\nu >3$ and $h \in \mathcal{H}_{\sigma,n}$. 
    From (\ref{proof_unweighted_L2_error_Q_P}) and the Cauchy-Schwarz inequality, we have
    \begin{align}\label{proof_unweighted_L2_error_finite_moment}
    \nonumber    \| h - h_0  \|_{2, \mathbb{Q}_{X_0}}^{2} &= \mathbb{E}_{X_0 \sim\mathbb{P}_{X_0}} \left[ r(X_0) \left( h(X_0) - h_0(X_0) \right)^2 \right] \\
       \nonumber  & \leq \left\{ \mathbb{E}_{X_0 \sim\mathbb{P}_{X_0}} \left[ r^2(X_0)  \right] \right\}^{\frac{1}{2}} \times  \left\{ \mathbb{E}_{X_0 \sim\mathbb{P}_{X_0}} \left[ \left( h(X_0) - h_0 (X_0) \right)^4 \right] \right\}^{\frac{1}{2}} \\
     \nonumber    & \leq \left\{ \mathbb{E}_{X_0 \sim\mathbb{P}_{X_0}} \left[ r^2(X_0)  \right] \right\}^{\frac{1}{2}} \times  \left\{ \mathbb{E}_{X_0 \sim\mathbb{P}_{X_0}} \left[ \left( h(X_0) - h_0 (X_0) \right)^2 \left( h(X_0) - h_0 (X_0) \right)^2  \right] \right\}^{\frac{1}{2}} \\
        & \lesssim  \| h - h_0 \|_{2,\mathbb{P}_{X_0}}.
    \end{align}
The inequality in (\ref{proof_unweighted_L2_error_finite_moment}) holds since $U_2 = \mathbb{E}_{X_0 \sim\mathbb{P}_{X_0}} \left[ r^2 (X_0) \right] < \infty$,  $\| h_0 \|_{\infty} \leq \mathcal{K}$ and $\| h \|_{\infty} \leq F$. 
From (\ref{proof_unweighted_L2_error_finite_moment}) and (\ref{proof_unweighted_L2_error_PX0_rate}), we have
\begin{align*}
 \mathbb{E}_{\mathbb{D}} \left[ \norm{\widehat{h}_{\mathbb{D}} - h_0 }^2_{2,\mathbb{Q}_{X_0}} \right] \lesssim \mathbb{E}_{\mathbb{D}} \left[ \| \widehat{h}_{\mathbb{D}} - h_0 \|_{2,\mathbb{P}_{X_0}} \right]
  \leq \left( \mathbb{E}_{\mathbb{D}} \left[ \| \widehat{h}_{\mathbb{D}} - h_0  \|_{2,\mathbb{P}_{X_0}}^2 \right]  \right)^{\frac{1}{2}} \lesssim \dfrac{\Big(\log \varphi(n) \Big)^{\nu/2}}{\Big( \varphi (n) \Big)^{\frac{s}{2s + d}}}.
\end{align*}
Thus the theorem follows.
\qed

\subsection{Proof of Proposition \ref{argmin_h0}}
 It suffices to show that $\mathcal{E}_{\mathbb{P},r} (h) \geq 0$ for all $h \in \mathcal{F}$.
\subsubsection*{1. Huber regression}
Let $h \in \mathcal{F}$.
From the proof of Lemma \ref{lemma_decomposition_estimator}, we have 
\begin{align*}
     \ell_\tau \big( Y_0 - h(X_0)  \big) - \ell_\tau \big( Y_0 - h_{0}(X_0) \big) &= \ell'_\tau\big( Y_0 - h_{0}(X_0) \big)  \big( h_{0}(X_0) - h(X_0) \big) \\
\nonumber &+ \int_0^{h_{0}(X_0) - h(X_0)} \ell''_\tau \big( Y_0 - h_{0}(X_0) +  z \big) \big( h_{0}(X_0) - h(X_0) -z \big) dz.
\end{align*}
So,
\begin{align*}
    r(X_0) \left(  \ell_\tau \big( Y_0 - h(X_0)  \big) - \ell_\tau \big( Y_0 - h_{0}(X_0) \big) \right) &= r(X_0)  \ell'_\tau\big( \xi_0 \big)  \big( h_{0}(X_0) - h(X_0) \big) \\
\nonumber &+ r(X_0)\int_0^{h_{0}(X_0) - h(X_0)} \ell''_\tau \big( Y_0 - h_{0}(X_0) +  z \big) \big( h_{0}(X_0) - h(X_0) -z \big) dz. 
\end{align*}
Since $\mathbb{E} \left[ \ell'_{\tau} (\xi_0) | X_0 = x \right] = 0 $ for all $x \in \mx$, we have $\mathbb{E}_{X_0 \sim \mathbb{P}_{X_0}} \big[ r(X_0)  \ell'_\tau\big( \xi_0 \big)  \big( h_{0}(X_0) - h(X_0) \big) \big] = 0 $.
 Furthermore,  from the proof of Lemma \ref{lemma_decomposition_estimator}, we have 
\[
 \mathbb{E} \left[ r(X_0)\int_0^{h_{0}(X_0) - h(X_0)} \ell''_\tau \big( Y_0 - h_{0}(X_0) +  z \big) \big( h_{0}(X_0) - h(X_0) -z \big) dz\Big | X_0 = x\right] \geq \dfrac{1}{2} \big(1 - \dfrac{3 C_\xi}{\tau} \big) \big( h_{0}(x) - h(X_0) \big)^2 r(x) \geq 0. 
\]
Then, we have 
\[
 \mathcal{E}_{\mathbb{P},r} (h) \geq \mathbb{E}_{X_0 \sim \mathbb{P}_{X_0}} \left[  \dfrac{1}{2} \big(1 - \dfrac{3 C_\xi}{\tau} \big) \big( h_{0}(X_0) - h(X_0) \big)^2 r(X_0) \right] \geq 0.
\] 
Hence, 
\[
 h_0 \in \argmin_{h \in \mathcal{F}} R_{\mathbb{P},r} (h).  
\]
\subsubsection*{2. Quantile regression}
 From the equation (B.3) in \cite{belloni2011l1}, we have for any $u, v \in \R$,
\[ \ell_\tau (u - v) - \ell_\tau (u) = - v \left( \tau - \ind_{\{u \leq 0 \}} \right) + \int_{0}^v  \left( \ind_{\{u \leq z \}} - \ind_{\{u \leq 0 \}} \right) dz. \]
Let $h \in \mathcal{F}$.  
By taking $u = Y_0 - h_0 (X_0)$ and $v = h(X_0) - h_0 (X_0) $, we have 
\begin{align*}
    r(X_0) \left( \ell_\tau (Y_0 - h(X_0) - \ell_\tau (Y_0 - h_0(X_0) \right) & = - r(X_0) \left( h(X_0) - h_0 (X_0)  \right) \left( \tau - \ind_{\{ Y_0 \leq h_0(X_0) \}} \right) \\
    & + r(X_0) \int_{0}^{h(X_0) - h_0 (X_0) } \left( \ind_{\{  Y_0 \leq h_0 (X_0) + z \}} - \ind_{\{  Y_0 \leq h_0 (X_0) \}} \right) dz. 
\end{align*}
Since $\mathbb{E} \left[  \tau - \ind_{\{ Y_0 \leq h_0(X_0) \}} |X_0 \right] = \tau - \tau = 0$, so
\begin{align*}
    \mathcal{E}_{\mathbb{P},r } (h) &= \mathbb{E}_{X_0 \sim \mathbb{P}_{X_0}} \left[ r(X_0) \int_{0}^{h(X_0) - h_0 (X_0) } \left( \mathbb{E}_{Y_0|X_0 \sim \mathbb{P}_{Y_0|X_0}} \left(\ind_{\{ Y_0 \leq h_0(X_0) + z\}} | X_0 \right) - \mathbb{E}_{Y_0|X_0 \sim \mathbb{P}_{Y_0|X_0}} \left(\ind_{\{ Y_0 \leq h_0(X_0) \}} | X_0\right) \right) dz  \right] \\ 
    &= \mathbb{E}_{X_0 \sim \mathbb{P}_{X_0}} \left[ r(X_0) \int_{0}^{h(X_0) - h_0 (X_0) } \left( F_{Y_0|X_0} (h_0 (X_0) + z) - F_{Y_0|X_0} (h_0 (X_0) )  \right) dz  \right].
\end{align*}
From Assumption \textbf{(A4)} and Lemma 13 in \cite{madrid2022risk}, one can find a constant $C>0$ such that,
\[\int_{0}^{h(X_0) - h_0 (X_0) } \left(  F_{Y_0|X_0} (h_0 (X_0) + z) - F_{Y_0|X_0} (h_0 (X_0) ) \right)  dz  \geq C \min \big(  |h(X_0) - h_0(X_0)|, |h(X_0) - h_0(X_0)|^2 \big)  \geq 0 .\]
Hence, since $r(X_0) \geq 0$ for all $x \in \mx$, we get $\mathcal{E}_{\mathbb{P},r} (h) \geq 0$. 
This completes the proof of the proposition. 
\qed 

\subsection{Proof of Theorem \ref{oracle_reweighted}}
\phantom{contenu} \newline 
We denote 
\begin{equation}
\label{start}
\mathcal{E}_{\mathbb{P},r} (\widehat{h}_{r,\mathbb{D}}):= \mathbb{E}[B_{1,n}] + \mathbb{E}[B_{2,n}].
\end{equation}

\noindent where,

\begin{multline}\label{proof_def_B1n}
B_{1,n} = \mathbb{E}_{(X_0,Y_0) \sim \mathbb{P}_{X_0,Y_0}} \left(  r(X_0)  \left[ \ell_\tau(Y_0 - \widehat{h}_{r,\mathbb{D}} (X_0) ) - \ell_\tau (Y_0 - h_0 (X_0))\right] \right)- \dfrac{2}{n} \sum_{i = 1}^n r(X_i)  \left[ \ell_\tau(Y_i - \widehat{h}_{r,\mathbb{D}}(X_i) ) - \ell_\tau (Y_i - h_0 (X_i))\right] \\
   - 2J_n(\widehat{h}_{r,\mathbb{D}});    
\end{multline}
and
\begin{equation}\label{proof_def_B2n}
    B_{2,n} = \dfrac{2}{n} \sum_{i = 1}^n r(X_i)  \left[ \ell_\tau(Y_i - \widehat{h}_{r,\mathbb{D}}(X_i) ) - \ell_\tau (Y_i - h_0 (X_i))\right] + 2J_n(\widehat{h}_{r,\mathbb{D}}).
\end{equation}
\noindent It suffices to derive bounds of $\mathbb{E}[B_{1,n}]$ and $\mathbb{E}[B_{2,n}]$.

\medskip
Let $\rho > 1/\varphi(n)$ and $L_n, N_n, B_n, F_n = F > 0$ satisfying the condition of the theorem. 
Set, 
\begin{equation*}
\mathcal{H}_{n,j,\rho} := \left\{h \in \mathcal{H}_{\sigma,n} : 2^{j-1}\mathbf{1}_{\{j \neq 0\}}\rho \leq J_n(h) \leq 2^j \rho \right\},
\end{equation*}
where $\mathcal{H}_{\sigma,n} := \mathcal{H}_{\sigma}(L_n, N_n, B_n, F),$.
\noindent Let

\begin{equation*}
G(h,Z_0) := r(X_0) \left[ \ell_\tau(Y_0 - h(X_0)) - \ell_\tau(Y_0- h_0(X_0)) \right] , \quad \text{with } Z_0 := (X_0, Y_0).
\end{equation*}

\noindent As in the proof of Theorem 4.1 in \cite{kengne2025deep} have for all $\rho > 1/\varphi(n)$,
\begin{align}
\nonumber & P(B_{1,n} > \rho) = P\left(\mathbb{E}[ r(X_0)\ell_\tau(Y_0 - \widehat{h}_{r,\mathbb{D}}(X_0))] - \mathbb{E}[r(X_0)\ell_\tau(Y_0 - h_0(X_0))]\right. \\
\nonumber & \hspace{7cm} \left.- \frac{2}{n}\sum_{i=1}^n r(X_i) [\ell_\tau(Y_i - \widehat{h}_{r,\mathbb{D}}(X_i)) - \ell_\tau(Y_i - h_0(X_i))] - 2J_n(\widehat{h}_{r,\mathbb{D}}) > \rho\right) \\
\nonumber &\leq P\left(\exists h \in \mathcal{H}_{\sigma,n} : \mathbb{E}[r(X_0)\ell_\tau(Y_0 - h(X_0))] - \mathbb{E}[r(X_0)\ell_\tau(Y_0 - h_0(X_0))] - \frac{1}{n}\sum_{i=1}^n r(X_i)  [\ell_\tau(Y_i - h(X_i)) - \ell_\tau(Y_i - h_0(X_i))]\right. \nonumber \\
\nonumber & \hspace{6.4cm} \left.> \frac{1}{2}\left(\mathbb{E}[ r(X_0)\ell_\tau(Y_0 - h(X_0))] - \mathbb{E}[r(X_0)\ell_\tau(Y_0 - h_0(X_0))] + 2J_n(h) + \rho\right)\right) \\
\nonumber &\leq P\left(\exists h \in \mathcal{H}_{\sigma,n} : \frac{1}{n}\sum_{i=1}^n [\mathbb{E}[G(h,Z_i)] - G(h,Z_i)] > \frac{1}{2}\left(\rho + 2J_n(h) + \mathbb{E}[G(h,Z_0)]\right)\right) \\
\nonumber &\leq P\left(\sup_{h \in \mathcal{H}_{\sigma,n}} \frac{\frac{1}{n}\sum_{i=1}^n [\mathbb{E}[G(h,Z_i)] - G(h,Z_i)]}{\rho + 2J_n(h) + \mathbb{E}[G(h,Z_0)]} > \frac{1}{2}\right) 
\leq \sum_{j=1}^{\infty} P\left(\sup_{h \in \mathcal{H}_{n,j,\rho}} \frac{\frac{1}{n}\sum_{i=1}^n [\mathbb{E}[G(h,Z_i)] - G(h,Z_i)]}{\rho + 2J_n(h) + \mathbb{E}[G(h,Z_0)]} > \frac{1}{2}\right) \\
\nonumber  &\leq \sum_{j=1}^{\infty} P\left(\sup_{h \in \mathcal{H}_{n,j,\rho}} \frac{\frac{1}{n}\sum_{i=1}^n [\mathbb{E}[G(h,Z_i)] - G(h,Z_i)]}{2^j \rho + \mathbb{E}[G(h,Z_0)]} > \frac{1}{2}\right) \\
\label{proof_B1n_sum_proba} & \leq \sum_{j=1}^{\infty} P\left(\sup_{h \in \mathcal{H}_{n,j,\rho}} \frac{\frac{1}{n}\sum_{i=1}^n [\mathbb{E}[G(h,Z_i)] - G(h,Z_i)]}{\sqrt{2^j \rho + \mathbb{E}[G(h,Z_0)]}} > \frac{1}{2}\sqrt{2^j \rho}\right) \\
 \label{proof_B1n_sum_proba2}    & = \sum_{j=1}^{\infty} P\left(\sup_{G \in \mathcal{G}_{n,j,\rho}} \frac{\frac{1}{n}\sum_{i=1}^n [\mathbb{E}[G(h,Z_i)] - G(h,Z_i)]}{\sqrt{2^j \rho + \mathbb{E}[G(h,Z_0)]}} > \frac{1}{2}\sqrt{2^j \rho}\right),
\end{align}
where 
\[\mathcal{G}_{n,j,\rho} = \{G(h, \cdot) : \mathbb{R}^d \times \mathcal{Y} \to \mathbb{R} : h \in \mathcal{H}_{n,j,\rho}\}, \]
and the inequality in (\ref{proof_B1n_sum_proba}) holds since $h \in \mathcal{H}_{n,j,\rho}, 2J_n(h) \geq 2^j \rho$ for all $j \geq 0$ and $\mathbb{E}[G(h,Z_0)] = \mathbb{E}[ r(X_0)\ell(Y_0-h(X_0))] - \mathbb{E}[r(X_0)\ell(Y_0-h_0(X_0))] \geq 0$ from Proposition \ref{argmin_h0}.
\noindent Set for any $h \in \mathcal{H}_{n,j,\rho}$ and $z = (x,y)$,
\[ g(h,z) = r(x) \left[ \ell_\tau(y-h(x)) - \ell_\tau(y-h_0(x)) \right] . \]

\noindent  
Let $h \in \mathcal{H}_{n,j,\rho}$.
Since $\|h\|_{\infty} \leq F, \|h_0\|_{\infty} \leq \mathcal{K}$ and $\ell_\tau$ is $\mk_{{\ell}_\tau}$-Lipschitzian, we have with $Z_0 = (X_0, Y_0)$,
\begin{equation*}
|g(h,Z_0)| = |G(h,Z_0)| \leq \Gamma \mathcal{K}_{{\ell}_\tau}(F + \mathcal{K}), 
\end{equation*}
 
\begin{equation*}
\text{Var}(g(h,Z_0)) \leq \mathbb{E}[g(h,Z_0)^2] \leq \Gamma  \mathcal{K}_{{\ell}_\tau}(F + K)\mathbb{E}[g(h,Z_0)],
\end{equation*}
and 
\[
 \lvert \mathbb{E} (g(h,Z_0)) - g(h,Z_0)) \rvert \leq 2 \Gamma \mk_{{\ell}_\tau}\left( F + \mathcal{K}\right). 
\]
By applying Lemma 6.1 in \cite{kengne2025general} with
\[
 C_1 = 2 \Gamma \mathcal{K}_{{\ell}_\tau} (F + \mathcal{K}), \quad C_2 = \Gamma \mathcal{K}_{{\ell}_\tau} (F + \mathcal{K}) ; \quad \varepsilon = 2^j \rho ; \quad u = \dfrac{1}{8}; \quad C_A, c_{\gamma} > 0,  
\]
we get,
\[
 P\left(\sup_{G \in \mathcal{G}_{n,j,\rho}} \frac{\frac{1}{n}\sum_{i=1}^n [\mathbb{E}[G(h,Z_i)] - G(h,Z_i)]}{\sqrt{2^j \rho + \mathbb{E}[G(h,Z_0)]}} > \frac{1}{2}\sqrt{2^j \rho}\right) \lesssim \mathcal{N} \left( \mathcal{G}_{n,j,\rho}, \dfrac{2^j \rho}{8}, \| \cdot \|_{\infty} \right)  \exp \left( - \dfrac{ 2^j \rho \, \varphi (n)}{64 \,\Gamma \mathcal{K}_{{\ell}_\tau} \left( F + \mathcal{K} \right) \left( c_{\gamma} + 2 C_A \right)} \right).  
\]
Set $\varepsilon = \frac{2^j \rho}{8}$, we find that
$$ \mathcal{N} \left( \mathcal{G}_{n,j,\rho}, \varepsilon, \| \cdot \|
_{\infty} \right) \leq \mathcal{N} \left( \mathcal{H}_{n,j,\rho}, \dfrac{\varepsilon}{\Gamma \mathcal{K}_{\ell}} , \| \cdot \|
_{\infty} \right). $$
For all $L, N, B, F, S \geq 0$, set
\[\widetilde{\mathcal{H}}_{\sigma}(L, N, B, F, S) := \{h \in \mathcal{H}_{\sigma}(L, N, B, F) : J_n(h) \leq \lambda_n S\}.\]
We have for all $j \in \mathbb{N}$ and $\rho > 0$
\[\mathcal{H}_{n,j,\rho} \subset \widetilde{\mathcal{H}}_{\sigma}(L, N, B, F, \frac{2^j \rho}{\lambda_n}).\]
From lemma E.6 in \cite{kurisu2025adaptive}, it holds that for all $\varepsilon \in \left( \Gamma \mathcal{K}_{{\ell}_\tau} \tau_n(L_n + 1)((N_n + 1)B_n)^{L_n+1} , \Gamma \mathcal{K}_{{\ell}_\tau} \right)$
\begin{align}
\mathcal{N} \left(\mathcal{G}_{n,j,\rho}, \varepsilon, \| \cdot \|_{\infty} \right) &\leq \mathcal{N} \left(\mathcal{H}_{n,j,\rho}, \frac{\varepsilon}{ \Gamma \mathcal{K}_{{\ell}_\tau}}, \| \cdot \|_{\infty}\right) \leq \mathcal{N} \left(\widetilde{\mathcal{H}}_{\sigma}(L_n, N_n, B_n, F, \frac{2^j \rho}{\lambda_n}), \frac{\varepsilon}{ \Gamma \mathcal{K}_{{\ell}_\tau}}, \| \cdot \|_{\infty} \right) \nonumber \\
&\leq \exp\left[2\frac{2^j \rho}{\lambda_n}(L_n + 1)\log\left(\frac{(L_n + 1)(N_n + 1)B_n}{\frac{\varepsilon}{\Gamma \mathcal{K}_{{\ell}_\tau}} - \tau_n(L_n + 1)((N_n + 1)B_n)^{L_n+1}}\right)\right]. 
\end{align}
\[ P(B_{1,n} > \rho) \lesssim \sum_{j=1}^{+ \infty} \exp\left[2\frac{2^j \rho}{\lambda_n}(L_n + 1)\log\left(\frac{(L_n + 1)(N_n + 1)B_n}{\frac{\varepsilon}{\Gamma \mathcal{K}_{{\ell}_\tau}} - \tau_n(L_n + 1)((N_n + 1)B_n)^{L_n+1}}\right)\right] \times \exp \left( - \dfrac{ 2^j \rho \, \varphi (n)}{64 \,\Gamma \mathcal{K}_{{\ell}_\tau} \left( F + \mathcal{K} \right) \left( c_{\gamma} + 2 C_A \right)} \right). \]
Then, we have
\begin{equation}
    P(B_{1,n} > \rho) \lesssim \sum_{j=1}^{+ \infty} \exp\left[2\frac{2^j \rho}{\lambda_n}(L_n + 1)\log\left(\frac{(L_n + 1)(N_n + 1)B_n}{\frac{\varepsilon}{\Gamma \mathcal{K}_{{\ell}_\tau}} - \tau_n(L_n + 1)((N_n + 1)B_n)^{L_n+1}}\right) - \dfrac{ 2^j \rho \, \varphi (n)}{64 \,\Gamma \mathcal{K}_{{\ell}_\tau} \left( F + \mathcal{K} \right) \left( c_{\gamma} + 2 C_A \right)}\right].
\end{equation}
The rest of the proof is obtained by following the same steps as in the proof of Theorem 4.1 in \cite{kengne2025deep} dealing with $\varphi(n)$ instead of $n^{(\alpha)}$.
This leads to, for sufficiently large n,
\begin{equation*}
    \mathbb{E}[B_{1,n}] \leq \int_0^{\infty} P(B_{1,n} > \rho)d\rho  \lesssim \dfrac{\Gamma}{\varphi(n)},
\end{equation*}
and
\[
\mathbb{E}[B_{2,n}] \leq 2J_n(h_{r, \mathbb{D}}^{\diamond}) + 2\mathcal{E}_{\mathbb{P}, r}(h_{r, \mathbb{D}}^{\diamond}) \leq 2 \inf_{h \in \mathcal{H}_{\sigma,n}} \left[\mathcal{E}_{\mathbb{P}, r}(h) + J_n(h)\right] + \frac{1}{\varphi (n)}. 
\]
Hence, the theorem follows.
\qed

\subsection{Proof of Corollary \ref{excess_risk_reweighted}}
Firstly, remark that for any predictor $h$, we have:
    \begin{equation}\label{Epr_L2_quantile_Huber}
     \mathcal{E}_{\mathbb{P},r} (h) \leq \frac{1}{2} (c_2 \vee 1) \Gamma \|h - h_0\|_{2,\mathbb{P}_{X_0}}^2,    
    \end{equation} 
    where $c_2$ is given in \textbf{(A4)}.
Indeed, on the one hand for quantile regression, we have from the Knight inequality, 
$$ \ell_\tau(u - v) - \ell_\tau(u) = -v(\tau - I(u \leq 0)) +
 \int_0^{v} (I(u \leq t) - I(u \leq 0)) dt, $$
and by taking $u=Y_0 - h_0(X_0)$ and $v= h(X_0)-h_0(X_0)$, we get,
\begin{align*}
\ell_\tau(Y_0 - h(X_0)) - \ell_\tau(Y_0 - h_0(X_0)) &= -(h(X_0) - h_0(X_0))(\tau - I(Y_0 \leq h_0(X_0))) \\
&\quad + \int_0^{h(X_0)-h_0(X_0)} (\ind(Y_0 \leq h_0(X_0) + t) - I(Y_0 \leq h_0(X_0))) dt.
\end{align*}
Taking the expectation and using Fubini's theorem, we obtain that
\begin{align*}
 \mathcal{E}_{\mathbb{P},r} (h) &= \mathbb{E}_{(X_0,Y_0) \sim \mathbb{P}_{X_0,Y_0}}[ r(X_0) \left(\ell_\tau(Y_0 - h(X_0)) - \ell_\tau(Y_0 - h_0(X_0)) \right)] \\
 &= -\mathbb{E}_{X_0 \sim\mathbb{P}_{X_0}}[ r(X_0) (h(X_0) - h_0(X_0)) \mathbb{E}_{Y_0|X_0}((\tau - I(Y_0 \leq h_0(X_0)))] \\
& \hspace{3cm} + \mathbb{E}_{X_0 \sim \mathbb{P}_{X_0}}\left[ r(X_0) \int_0^{h(X_0)-h_0(X_0)} [\mathbb{E}_{Y_0|X_0} (I(Y_0 \leq h_0(X_0) + t)) - \mathbb{E}_{Y_0|X_0} (I(Y_0 \leq h_0(X_0)))] dt  \right].
\end{align*}
Since 
$  \mathbb{E}_{Y_0|X_0} \left[ \tau - I(Y_0 \leq h_0(X_0))|X\right]  = \tau - \mathbb{P} \left[ Y_0 \leq h_0(X_0))|X_0  \right] = \tau - \tau = 0$, 
we get
\begin{align*}
 \mathcal{E}_{\mathbb{P},r} (h) = &\mathbb{E}_{X_0 \sim \mathbb{P}_{X_0}}\left[ r(X_0) \int_0^{h(X_0)-h_0(X_0)} [\mathbb{E}_{Y_0|X_0} (I(Y_0 \leq h_0(X_0) + t)) - \mathbb{E}_{Y_0|X_0} (I(Y_0 \leq h_0(X_0)))] dt  \right] \\
&= \mathbb{E}_{X_0 \sim \mathbb{P}_{X_0}}\left[ r(X_0) \int_0^{h(X_0)-h_0(X_0)} [ F_{Y_0|X_0} (h_0(X_0) + t) - F_{Y_0|X_0}  (h_0(X_0))] dt\right] \\
&\leq c_2 \mathbb{E}_{X_0 \sim \mathbb{P}_{X_0}}\left[ r(X_0) \int_0^{|h(X_0)-h_0(X_0)|} t dt\right] \leq  \frac{c_2 \Gamma}{2} \|h - h_0\|_{2,\mathbb{P}_{X_0}}^2,
\end{align*}
where the first inequality holds from the mean value theorem and $c_2$ is given in \textbf{(A4)}. Hence, (\ref{Epr_L2_quantile_Huber}) is satisfied.

\medskip
\noindent
On the other hand for Huber regression, by taking $u= Y_0 - h_0(X_0)$ and $v= h_0(X_0) - h(X_0)$ in \eqref{fundamental_calculus_huber}, we get,
\begin{multline*}
   r(X_0) \left(  \ell_\tau \big( Y_0 - h(X_0)  \big) - \ell_\tau \big( Y_0 - h_0(X_0)\big) \right) = r(X_0) \ell'_\tau\big( \xi_0 \big)  \big( h_0(X_0) - h(X_0) \big) \\ 
   + r(X_0)\int_0^{h_0(X_0) - h(X_0)} \ell''_\tau \big( \xi_0 +  z \big) \big( h_0(X_0) - h(X_0) -z \big) dz.
\end{multline*}
From the proof of Proposition 3.1 in \cite{fan2024noise}, we have for all $x \in \mx$, $\E[ r(X_0)\ell'_\tau(\xi_0) | X_0=x] =0$. Hence, $\E\big[  \big( h_0(X_0) - h(X_0) \big)  \ell'_\tau\big( \xi_0 \big)  \big] =0  $.
Now, let $x \in \mx$. 
By going in the same way as in (\ref{proof_lem_unwei_est_ep_geq_square}), we get,
\begin{align}\label{proof_lem_unwei_est_ell_tau_int2}
\nonumber &\E\Big[ \int_0^{h_0(X_0) - h(X_0)} \ell''_\tau \big( \xi_0 +  z \big) \big( h_0(X_0) - h(X_0) -z \big) dz  \big| X_0=x\Big]\\ 
 &= \dfrac{1}{2} \big( h_0(x) - h(x) \big)^2 - \E\Big[ \int_0^{h_0(x) - h(x)} \ind\big(|\xi_0 +  z | > \tau \big)  \big( h_0(x) - h(x) -z \big) dz  \big| X_0=x\Big] \leq \dfrac{1}{2} \big( h_0(x) - h(x) \big)^2, 
\end{align}
where the last inequality holds since
$\int_0^{h_0(x) - h(x)} \ind\big(|\xi_0 +  z | > \tau \big)  \big( h_0(x) - h(x) -z \big) dz \geq 0$.
So, (\ref{proof_lem_unwei_est_ell_tau_int2}) gives,

\[  \E\Big[  r(X_0)\int_0^{h_0(X_0) - h(X_0)} \ell''_\tau \big( \xi_0 +  z \big) \big( h_0(X_0) - h(X_0) -z \big) dz  \big| X_0=x\Big] \leq  \dfrac{1}{2} r(x)  \big( h_0(x) - h(x) \big)^2.  \]
Hence,
\[  \mathcal{E}_{\mathbb{P},r} (h) \leq \mathbb{E}_{X_0 \sim \mathbb{P}_{X_0}} \left[ \dfrac{1}{2} r(X_0)  \big( h_0(X_0) - h(X_0) \big)^2 \right] \leq \dfrac{\Gamma}{2} \norm{h - h_0}^2_{2,\mathbb{P}_{X_0}}.\] 
Thus, (\ref{Epr_L2_quantile_Huber}) holds.

\medskip
Now, let $L_n, N_n, B_n, F > 0$ satisfying the conditions in Corollary \ref{excess_risk_reweighted} and $h_0 \in C^s(\mathcal{X}, K)$ with $s, K > 0$. Set
\[
\varepsilon_n = \frac{1}{\varphi (n)^{\frac{s}{2 s+d}}}
\]
From Theorem 3.2 in \cite{kengne2025excess}, there exist positive constants $L_0, N_0,$ $B_0, S_0 > 0$ such that with $\tilde{L}_n = \frac{sL_0}{2 s + d} \log \varphi( n), \tilde{N}_n = N_0 \varphi(n)^{\frac{s}{2 s+d}}, \tilde{S}_n = \frac{sS_0}{2 s+d} \varphi(n)^{\frac{d}{2 s+d}} \log \varphi (n),$ and $\tilde{B}_n = B_0 \varphi(n)^{\frac{4(d+s)}{2 s+d}}$, there exists a neural network $\tilde{h}_{r,\mathbb{D}} \in \tilde{\mathcal{H}}_{\sigma,n} := \mathcal{H}_\sigma(\tilde{L}_n, \tilde{N}_n, \tilde{B}_n, F, \tilde{S}_n)$ satisfying,

\begin{equation}\label{proof_cor_approx_DNN}
\|\tilde{h}_{r,\mathbb{D}} - h_0\|_{\infty,\mathcal{X}} \leq \varepsilon_n = \frac{1}{\left( \varphi (n) \right)^{\frac{s}{2 s+d}}}. 
\end{equation}
Set
\[
\tilde{\mathcal{H}}_{\sigma,n,\varepsilon_n} := \{h \in \tilde{\mathcal{H}}_{\sigma,n}, \|h - h_0\|_{\infty,\mathcal{X}} \leq \varepsilon_n\}.
\]
\medskip
Furthermore, from the assumption on the penalty term, we have for any $h \in \tilde{\mathcal{H}}_{\sigma,n}$, $J_n(h) \lesssim \lambda_n\tilde{S}_n$ where $\lambda_n \asymp \frac{(\log \varphi (n))^{\nu_3}}{\varphi (n)}$ with $\nu_3 > 2$.
Since $\widetilde{\mathcal{H}}_{\sigma,n,\varepsilon_n} \subset \widetilde{\mathcal{H}}_{\sigma,n} \subset \mathcal{H}_\sigma(L_n, N_n, B_n, F, \widetilde{S}_n)\subset \mathcal{H}_\sigma(L_n, N_n, B_n, F) $), in addition to Theorem \ref{oracle_reweighted} and (\ref{Epr_L2_quantile_Huber}), we get:
\begin{align}
   \nonumber   \mathbb{E}[\mathcal{E}_{\mathbb{P},r}(\widehat{h}_{r,\mathbb{D}})] & \lesssim \inf_{h \in \mathcal{H}_\sigma(L_n, N_n, B_n, F)} \left[\mathcal{E}_{\mathbb{P},r}(h) + J_n(h)\right] + \dfrac{1}{\varphi (n)}   \lesssim \inf_{h \in \mathcal{H}_\sigma(\tilde{L}_n, \tilde{N}_n, \tilde{B}_n, F, \tilde{S}_n)} \left[\mathcal{E}_{\mathbb{P},r}(h) + J_n(h)\right] + \dfrac{1}{\varphi (n)} \\
   \nonumber   & \lesssim \inf_{h \in \tilde{\mathcal{H}}_{\sigma,n}} \left[\mathcal{E}_{\mathbb{P},r}(h) + \lambda_n\tilde{S}_n\right] + \dfrac{1}{\varphi (n)}   \lesssim \inf_{h \in \tilde{\mathcal{H}}_{\sigma,n,\varepsilon_n}} \left[\mathcal{E}_{\mathbb{P},r}(h) + \lambda_n\tilde{S}_n\right] + \dfrac{1}{\varphi (n)} \\ 
 \nonumber   & \lesssim \inf_{h \in \tilde{\mathcal{H}}_{\sigma,n,\varepsilon_n}} \Gamma \|h - h_0\|_{2,\mathbb{P}_{X_0}}^{2} + \lambda_n\tilde{S}_n + \dfrac{1}{\varphi (n)}   \lesssim \inf_{h \in \tilde{\mathcal{H}}_{\sigma,n,\varepsilon_n}} \Gamma  \|h - h_0\|_{\infty,\mathcal{X}}^{2} + \lambda_n\tilde{S}_n + \dfrac{1}{\varphi (n)} \\
  \label{max_gamma_equation}  & \lesssim \Gamma  \left( \varphi (n) \right)^{-\frac{2 s}{2 s+d}} + \left( \varphi (n) \right)^{\frac{d}{2 s+d}-1} \left( \log \varphi (n) \right)^{\nu_3+1} + \dfrac{\Gamma}{\varphi (n)} \lesssim \Gamma\dfrac{\left( \log n \right)^{\nu_3 + 1}}{\left( \varphi (n) \right)^{\frac{2 s}{2 s+d}}}.
\end{align}
Hence,
\[
\mathbb{E}[\mathcal{E}_{\mathbb{P},r}(\widehat{h}_{r,\mathbb{D}})] \lesssim \dfrac{\left( \log \varphi(n) \right)^{\nu}}{\left( \varphi (n) \right)^{\frac{2 s}{2 s+d}}},
\]
for all $v > 3$.
\qed

\subsection{Proof of Theorem \ref{reweighted_estimator_result}}
Let $\nu>3$.
From Lemma \ref{lemma_decomposition_estimator_reweighted}, we have 
\begin{equation}\label{proof_theo_rewei_L2_error_QX0}
    \E_{\mathbb{D}}\left[ \norm{\widehat{h}_{r, \mathbb{D}} - h_0}_{2, \mathbb{Q}_{X_0}}^2  \right] \lesssim  \E_{\mathbb{D}} \left\{ \mathbb{E}_{(X_0, Y_0) \sim \mathbb{P}_{X_0, Y_0}}  \left[ r(X_0) \left( \ell_\tau \big( Y_0 - \widehat{h}_{r, \mathbb{D}}(X_0)  \big) - \ell_\tau \big( Y_0 - h_{\mathcal{H}_{\sigma,n}}(X_0) \big) \right) \right] \right\} + \inf_{h \in \mathcal{H}_{\sigma,n}} \| h - h_0\|^2_{\infty, \mx}.
\end{equation}
Since $h_0$ is a target function with respect to the reweighted risk (see Proposition \ref{argmin_h0}), and in addition to Corollary \ref{excess_risk_reweighted}, we have
\begin{equation}\label{proof_theo_rewei_delta_re_risk}
   \E_{\mathbb{D}} \left\{ \mathbb{E}_{(X_0, Y_0) \sim \mathbb{P}_{X_0, Y_0}}  \left[ r(X_0) \left( \ell_\tau \big( Y_0 - \widehat{h}_{r, \mathbb{D}}(X_0)  \big) - \ell_\tau \big( Y_0 - h_{\mathcal{H}_{\sigma,n}}(X_0) \big) \right) \right] \right\}  \leq \E_{\mathbb{D}}[\mathcal{E}_{\mathbb{P},r}(\widehat{h}_{r,\mathbb{D}})] \lesssim   \dfrac{\Big(\log \varphi(n) \Big)^{\nu}}{\Big( \varphi (n) \Big)^{\frac{2s}{2s + d}}}. 
\end{equation}
In addition to (\ref{proof_approx_DNN}), it holds from (\ref{proof_theo_rewei_L2_error_QX0}) that,
    \[
 \E_{\mathbb{D}}\left[ \norm{\widehat{h}_{r, \mathbb{D}} - h_0}_{2, \mathbb{Q}_{X_0}}^2  \right] \lesssim  \dfrac{\Big(\log \varphi(n) \Big)^{\nu}}{\Big( \varphi (n) \Big)^{\frac{2s}{2s + d}}} +    \dfrac{1}{\Big( \varphi (n) \Big)^{\frac{2s}{2s + d}}} \lesssim \dfrac{\Big(\log \varphi(n) \Big)^{\nu}}{\Big( \varphi (n) \Big)^{\frac{2s}{2s + d}}},  
\]
 which completes the proof. 
\qed

\subsection{Proof of Theorem \ref{reweighted_estimator_truncated_ratio_result}}
Let $h \in \mathcal{H}_{\sigma,n}$. 
Set,
 \begin{equation*}
 \widetilde{\mathcal{E}}_{\mathbb{P}, r} (h) := \mathbb{E}_{(X_0,Y_0) \sim \mathbb{P}_{X_0,Y_0}} \left[ r(X_0) \Big( \ell_{\tau} (Y_0 - h(X_0)) - \ell_{\tau} (Y - h_{\mathcal{H}_{\sigma,n}}(X)) \Big) \right] ,  
 \end{equation*}
 and
    \begin{equation*}
 \widetilde{\mathcal{E}}_{\mathbb{P}, T_\eta r} (h)   := \mathbb{E}_{(X_0,Y_0) \sim \mathbb{P}_{X_0,Y_0}} \left[ T_\eta r(X_0) \Big( \ell_{\tau} (Y_0 - h(X_0)) - \ell_{\tau} (Y - h_{\mathcal{H}_{\sigma,n}}(X)) \Big) \right],
\end{equation*}
where $h_{\mathcal{H}_{\sigma,n}}$ is given in (\ref{def_h_H_sigma}).
Furthermore, one can easily see that,
\[ |T_\eta r(X_0) - r(X_0)| \leq r(X_0)\ind (r(X_0) \geq  \eta) .    \]
Hence, according to the Lipschitz condition on $\ell_\tau$ and \textbf{(A3)} with $\mu=1+\delta$, we get
\begin{align}\label{proof_rewei_est_truncated_delta_EP_tilde}
 \nonumber   \widetilde{\mathcal{E}}_{\mathbb{P}, r} (h) - \widetilde{\mathcal{E}}_{\mathbb{P}, T_\eta r} (h)  & = \mathbb{E}_{(X_0,Y_0) \sim \mathbb{P}_{X_0,Y_0}} \left[ \left( T_\eta r(X_0) - r(X_0)  \right)  \Big( \ell_{\tau} (Y_0 - h(X_0)) - \ell_{\tau} (Y - h_{\mathcal{H}_{\sigma,n}}(X_0)) \Big) \right] \\ 
   &\leq F \mathcal{K}_{\ell_{\tau}} \mathbb{E}_{X_0\sim \mathbb{P}_{X_0}} \left[ r(X_0)\ind (r(X_0) \geq  \eta) \right] \leq F \mathcal{K}_{\ell_{\tau}} \mathbb{E}_{X_0\sim \mathbb{P}_{X_0}} \left[ r(X_0) \dfrac{r^{\delta} (X_0)}{\eta^{\delta} } \right]     \leq \dfrac{ F \mathcal{K}_{\ell_{\tau}} U_{\mu} }{\eta^\delta}.
\end{align}
Therefore, 
 \begin{equation*}
 \widetilde{\mathcal{E}}_{\mathbb{P}, r} (h)  \leq  \widetilde{\mathcal{E}}_{\mathbb{P}, T_\eta r} (h) + \dfrac{ F \mathcal{K}_{\ell_{\tau}} U_\mu }{\eta^\delta}. 
 \end{equation*}
In addition to Lemma \ref{lemma_decomposition_estimator_reweighted}, we have,
\begin{equation}\label{proof_theo_T_eta_1} 
\E_{\mathbb{D}}\left[ \norm{\widehat{h}_{T_\eta r, \mathbb{D}} - h_0}_{2, \mathbb{Q}_{X_0}}^2  \right] \leq C_1 \E_{\mathbb{D}}  \left[  \widetilde{\mathcal{E}}_{\mathbb{P}, T_\eta r} (\widehat{h}_{T_\eta r, \mathbb{D}})   \right] +  C_1 \dfrac{ F \mathcal{K}_{\ell_{\tau}} U_\mu }{\eta^\delta} +C_2 \inf_{h \in \mathcal{H}_{\sigma,n}} \| h - h_0\|^2_{\infty, \mx},  
\end{equation}
for some $C_1, C_2 >0$.
Recall that $\eta \asymp \Big( \varphi(n) \Big)^{\frac{2 s}{(1 + \delta)(2 s +d)}} \asymp \left( \varphi(n) \right)^{\frac{1 - \nu_4}{1+ \delta}}$ with $ \nu_4 = \frac{d}{2s + d}  $. 
Since $T_\eta r(X_0) \leq \eta$, it holds from \eqref{max_gamma_equation} that,
\begin{equation}\label{proof_theo_T_eta_2} 
 \E_{\mathbb{D}}  \left[  \widetilde{\mathcal{E}}_{\mathbb{P}, T_\eta r} (\widehat{h}_{T_\eta r, \mathbb{D}})   \right] \lesssim \eta \dfrac{\Big(\log \varphi(n) \Big)^{\nu}}{\Big( \varphi (n) \Big)^{\frac{2s}{2s + d}}} \lesssim 
   \dfrac{\Big(\log \varphi(n) \Big)^{\nu}}{\Big( \varphi (n) \Big)^{\frac{\delta}{ 1+ \delta} \times \frac{2s}{2s + d}}},
\end{equation}
for all $\nu >3$.
So, from (\ref{proof_theo_T_eta_1} ), (\ref{proof_theo_T_eta_2}) and (\ref{proof_cor_approx_DNN}), we get for all $\nu >3$,
\[
 \E_{\mathbb{D}}\left[ \norm{\widehat{h}_{T_\eta r, \mathbb{D}} - h_0}_{2, \mathbb{Q}_{X_0}}^2  \right]  \lesssim \dfrac{\Big(\log \varphi(n) \Big)^{\nu}}{\Big( \varphi (n) \Big)^{\frac{\delta}{ 1+ \delta} \times \frac{2s}{2s + d}}} + \dfrac{1}{\Big( \varphi (n) \Big)^{\frac{\delta}{ 1+ \delta} \times \frac{2s}{2s + d}}} + \dfrac{1}{\Big( \varphi (n) \Big)^{\frac{2s}{2s + d}}} \lesssim \dfrac{\Big(\log \varphi(n) \Big)^{\nu}}{\Big( \varphi (n) \Big)^{\frac{\delta}{ 1+ \delta} \times \frac{2s}{2s + d}}}. 
\]
Hence, the theorem follows.
\qed

\subsection{Proof of Proposition \ref{ratio_estimator_result}}
Consider the function $\ell$ defined by: 
    $$ \ell (u; x) = \dfrac{1}{2} \big( u^2(x_p) - r^2(x_p) \big) - \big( u(x_q) - r(x_q) \big) ~
    \text{ for all }  ~ x = (x_p, x_q) \in \mx^2 \text{ and } u \in \mathcal{H}_{\sigma,m}. $$
$\ell$ is Lipchitz continuous over $u$; that is, for all  $x = (x_p,x_q) \in \mx^2$ and $u_1, u_2 \in \mathcal{H}_{\sigma,n}$, one can easily get,
\[
 \Big \lvert \ell (u_1;x) - \ell (u_2;x) \Big \rvert \leq (F+1)\Big( \lvert u_1(x_p) - u_2(x_p)\rvert + \lvert u_1(x_q) - u_2(x_q)\rvert \Big) \leq 2(F+1) \|u_1 - u_2\|_\infty. 
\]
Let $X_0^P$ and $X_0^Q$ be two random variables with values in $\mx$ and distribution $\mathbb{P}_{X_0}$ and $\mathbb{Q}_{X_0}$ respectively. Set $\widetilde{X}_0 = (X_0^P, X_0^Q)$.
Define the contrast function,
\begin{equation*}
    C(u) = \E[\ell(u; \widetilde{X}_0)], ~ \text{ for all } u \in \mathcal{H}_{\sigma,m}. 
\end{equation*}
From the extra unlabelled samples, $S_{\mathbb{P}} := \{X_i^P\}_{i=1}^m$ and $S_{\mathbb{Q}} := \{X_i^Q\}_{i=1}^m$, set  $\widetilde{X}_i = (X_i^P, X_i^Q)$ for all $i = 1, \cdots m$.
Define the empirical version of the contrast $C$ by,
\begin{equation*}
   \widehat{C}_n (u) = \frac{1}{m} \sum_{i=1}^m \ell(u; \widetilde{X}_i), ~ \text{ for all } u \in \mathcal{H}_{\sigma,m}. 
\end{equation*}
The density ratio $r$ and the estimator $\widehat{r}_{\mathbb{S}} $ satisfy,
\[ r \in \argmin_{u \in \mf} C(u) ~ \text{ and } ~  \widehat{r}_{\mathbb{S}} \in \argmin_{u \in \mathcal{H}_{\sigma,m}}\big[ \widehat{C}_m(u)+ J_m(u)  \big] .  \]
Let $u \in \mathcal{H}_{\sigma,m}$. By using the equality $\E[r(X_0^P) u(X_0^P)] = \E[ u(X_0^Q)]$, we get,
\[ C(u) - C(r) = C(u) = \E\big[   \dfrac{1}{2} \big( u^2(X_0^P) - r^2(X_0^P) \big) - \big( u(X_0^Q) - r(X_0^Q) \big) \big] = \frac{1}{2}\E\big[ \big( u(X_0^P) - r(X_0^P)  \big)^2 \big] = \frac{1}{2} \|u-r\|^2_{2, \mathbb{P}_{X_0}}  .\]
Therefore,  since $ r \in \mathcal{C}^{s}( \mathcal{X}, \Gamma) $ and in addition to (\ref{cond_arch_H_m}) and (\ref{cond_tuning_param}), one can follow the same steps as in the proof of Theorem 4.1 and Corollary 4.3 in \cite{kengne2025deep} (with $\varphi(m)$ instead of $n^{(\alpha)}$ and with $\kappa=2$) to obtain the result of the proposition.
\qed

\subsection{Proof of Theorem \ref{pretraining_reweighted_estimator_result}}
 From Lemma \ref{lemma_decomposition_estimator_reweighted}, one can find $C_1, C_2 >0$ such that,
\begin{align}\label{proof_E_SD_h_h0}
   \nonumber &    \mathbb{E}_{\mathbb{S}, \mathbb{D}} \left[ \norm{\widehat{h}_{\widehat{r}_\mathbb{S}, \mathbb{D} } - h_0 }^2_{2, \mathbb{Q}_{X_0}} \right] \\
       \nonumber  &\leq C_1 \E_{\mathbb{S,D}} \mathbb{E}_{(X_0, Y_0) \sim \mathbb{P}_{X_0, Y_0}}  \left[ \Big( r(X_0) - \widehat{r}_{\mathbb{S}} (X_0) + \widehat{r}_{\mathbb{S}} (X_0)\Big) \left( \ell_\tau \big( Y_0 - \widehat{h}_{\widehat{r}_\mathbb{S}, \mathbb{D}}(X_0)  \big) - \ell_\tau \big( Y_0 - h_{\mathcal{H}_{\sigma,n}}(X_0) \big) \right) \right] \\ 
     \nonumber & \hspace{12.5cm} +C_2 \inf_{h \in \mathcal{H}_{\sigma,n}} \| h - h_0\|^2_{\infty, \mx} \\ 
  \nonumber   &\leq C_1 \E_{\mathbb{S,D}} \mathbb{E}_{(X_0, Y_0) \sim \mathbb{P}_{X_0, Y_0}}  \left[ \Big( r(X_0) - \widehat{r}_{\mathbb{S}} (X_0) \Big) \left( \ell_\tau \big( Y_0 - \widehat{h}_{\widehat{r}_\mathbb{S}, \mathbb{D}}(X_0)  \big) - \ell_\tau \big( Y_0 - h_{\mathcal{H}_{\sigma,n}}(X_0) \big) \right) \right] \\
     & +  C_1 \E_{\mathbb{S,D}} \mathbb{E}_{(X_0, Y_0) \sim \mathbb{P}_{X_0, Y_0}}  \left[ \widehat{r}_{\mathbb{S}} (X_0) \left( \ell_\tau \big( Y_0 - \widehat{h}_{\widehat{r}_\mathbb{S}, \mathbb{D}}(X_0)  \big) - \ell_\tau \big( Y_0 - h_{\mathcal{H}_{\sigma,n}}(X_0) \big) \right) \right] +C_2 \inf_{h \in \mathcal{H}_{\sigma,n}} \| h - h_0\|^2_{\infty, \mx}. 
\end{align}
We have for all $\beta > 0$, 
\begin{align}\label{proof_def_Anm}
  \nonumber  A_{n,m} &:= \E_{\mathbb{S,D}} \mathbb{E}_{(X_0, Y_0) \sim \mathbb{P}_{X_0, Y_0}}  \left[ \Big( r(X_0) - \widehat{r}_{\mathbb{S}} (X_0) \Big) \left( \ell_\tau \big( Y_0 - \widehat{h}_{\widehat{r}_\mathbb{S}, \mathbb{D}}(X_0)  \big) - \ell_\tau \big( Y_0 - h_{\mathcal{H}_{\sigma,n}}(X_0) \big) \right) \right] \\
    \leq & \dfrac{1}{2 \beta} \mathbb{E}_{\mathbb{S}} \left[  \mathbb{E}_{X_0 \sim \mathbb{P}_{X_0}} \big(  r(X_0) - \widehat{r}_{\mathbb{S}} (X_0) \big)^2 \right]  + \dfrac{\beta}{2} \mathbb{E}_{\mathbb{S},\mathbb{D}} \mathbb{E}_{X_0,Y_0 \sim \mathbb{P}_{X_0,Y_0}} \left[ \Big ( \ell_\tau \big( Y_0 - \widehat{h}_{\widehat{r}_\mathbb{S}, \mathbb{D}}(X_0)  \big) - \ell_\tau \big( Y_0 - h_{\mathcal{H}_{\sigma,n}}(X_0) \big)  \Big )^2 \right] .
\end{align}
It follows from the assumption \textbf{(A5)} that,
\begin{equation*} %\label{proof_norm_L2_Upsilon}
 \mathbb{E}_{X_0 \sim \mathbb{Q}_{X_0}} \left[\right( \widehat{h}_{\widehat{r}_\mathbb{S}, \mathbb{D}} (X_0)  - h_{\mathcal{H}_{\sigma,n}} (X_0) \left)^2 \right] = \mathbb{E}_{X_0 \sim\mathbb{P}_{X_0}} \left[ r(X_0)  \left( \widehat{h}_{\widehat{r}_\mathbb{S}, \mathbb{D}} (X_0)  - h_{\mathcal{H}_{\sigma,n}} (X_0)  \right)^2 \right] \geq \Upsilon  \norm{\widehat{h}_{\widehat{r}_\mathbb{S}, \mathbb{D}} - h_{\mathcal{H}_{\sigma,n}}}_{2,\mathbb{P}_{X_0}}^2. 
\end{equation*}
In addition, according to the Lipschitz condition on $\ell_\tau$, we get
 \begin{align}\label{proof_exp_SD_L2_Upsilon}
 \nonumber \mathbb{E}_{\mathbb{S},\mathbb{D}} \mathbb{E}_{X_0,Y_0 \sim \mathbb{P}_{X_0,Y_0}} \left[  \right( \ell_\tau \big( Y_0 - \widehat{h}_{\widehat{r}_\mathbb{S}, \mathbb{D}}(X_0)  \big) - \ell_\tau \big( Y_0 - h_{\mathcal{H}_{\sigma,n}}(X_0) \big)   \left)^2 \right] & \leq \mk_{\ell_\tau}^2 \mathbb{E}_{\mathbb{S},\mathbb{D}} \left[ \norm{\widehat{h}_{\widehat{r}_\mathbb{S}, \mathbb{D}} - h_{\mathcal{H}_{\sigma,n}}}_{2, \mathbb{P}_{X_0}}^2 \right] \\
 & \leq  \dfrac{ \mk_{\ell_\tau}^2 }{\Upsilon} \mathbb{E}_{\mathbb{S,D}} \left[ \norm{\widehat{h}_{\widehat{r}_\mathbb{S}, \mathbb{D}} - h_{\mathcal{H}_{\sigma,n}}}_{2, \mathbb{Q}_{X_0}}^2 \right].    
 \end{align}
 (\ref{proof_def_Anm}) and (\ref{proof_exp_SD_L2_Upsilon}) give
 \begin{equation}\label{proof_def_Anm_eq_beta_L2}
      A_{n,m} \leq \dfrac{1}{2 \beta} \mathbb{E}_{\mathbb{S}} \left[  \norm{\widehat{r}_{\mathbb{S}} - r}_{2,\mathbb{P}_{X_0}}^2 \right] + \dfrac{\beta \mk_{\ell_\tau}^2}{\Upsilon} \mathbb{E}_{\mathbb{S,D}} \left[ \norm{\widehat{h}_{\widehat{r}_\mathbb{S}, \mathbb{D}} - h_{0}}_{2, \mathbb{Q}_{X_0}}^2 \right] + \dfrac{\beta \mk_{\ell_\tau}^2 }{\Upsilon} \norm{h_{\mathcal{H}_{\sigma,n}} - h_{0}}_{2, \mathbb{Q}_{X_0}}^2. 
 \end{equation}
Hence, it holds from (\ref{proof_E_SD_h_h0}), (\ref{proof_def_Anm}) and (\ref{proof_def_Anm_eq_beta_L2}) that,
\begin{align}\label{proof_E_SD_h_h0_eq2}
     \nonumber   \mathbb{E}_{\mathbb{S}, \mathbb{D}} \left[ \norm{\widehat{h}_{\widehat{r}_\mathbb{S}, \mathbb{D} } - h_0 }^2_{2, \mathbb{Q}_{X_0}} \right] 
     &\leq \dfrac{C_1}{2 \beta} \mathbb{E}_{\mathbb{S}} \left[  \norm{\widehat{r}_{\mathbb{S}} - r}_{2,\mathbb{P}_{X_0}}^2 \right] + \dfrac{C_1 \beta \mk_{\ell_\tau}^2 }{\Upsilon} \mathbb{E}_{\mathbb{S,D}} \left[ \norm{\widehat{h}_{\widehat{r}_\mathbb{S}, \mathbb{D}} - h_{0}}_{2, \mathbb{Q}_{X_0}}^2 \right] + \dfrac{C_1 \beta \mk_{\ell_\tau}^2 }{\Upsilon} \norm{h_{\mathcal{H}_{\sigma,n}} - h_{0}}_{\infty, \mathcal{X}}^2 \\
   \nonumber  & +  C_1 \E_{\mathbb{D}} \mathbb{E}_{(X_0, Y_0) \sim \mathbb{P}_{X_0, Y_0}}  \left[ \widehat{r}_{\mathbb{S}} (X_0) \left( \ell_\tau \big( Y_0 - \widehat{h}_{\widehat{r}_\mathbb{S}, \mathbb{D}}(X_0)  \big) - \ell_\tau \big( Y_0 - h_{\mathcal{H}_{\sigma,n}}(X_0) \big) \right) \right] \\ 
     & +  C_2 \inf_{h \in \mathcal{H}_{\sigma,n}} \| h - h_0\|^2_{\infty, \mx}.
\end{align}
By choosing $\beta$ sufficiently small, such as $ \beta < \Upsilon / C_1 \mk_{\ell_\tau}^2$, we get from (\ref{proof_E_SD_h_h0_eq2}), 
\begin{align*}
  \nonumber  \Big( 1 - \dfrac{C_1 \beta \mk_{\ell_\tau}^2 }{\Upsilon} \Big)\mathbb{E}_{\mathbb{S}, \mathbb{D}} \left[ \norm{\widehat{h}_{\widehat{r}_\mathbb{S}, \mathbb{D} } - h_0 }^2_{2, \mathbb{Q}_{X_0}} \right] & \leq \dfrac{C_1 }{2 \beta} \mathbb{E}_{\mathbb{S}} \left[  \norm{\widehat{r}_{\mathbb{S}} - r}_{2,\mathbb{P}_{X_0}}^2 \right] \\ 
\nonumber & +  C_1 \E_{\mathbb{S,D}} \mathbb{E}_{(X_0, Y_0) \sim \mathbb{P}_{X_0, Y_0}}  \left[ \widehat{r}_{\mathbb{S}} (X_0) \left( \ell_\tau \big( Y_0 - \widehat{h}_{\widehat{r}_\mathbb{S}, \mathbb{D}}(X_0)  \big) - \ell_\tau \big( Y_0 - h_{\mathcal{H}_{\sigma,n}}(X_0) \big) \right) \right] \\ 
 & + \Big( C_2 + \dfrac{C_1 \beta \mk_{\ell_\tau}^2 }{\Upsilon} \Big) \inf_{h \in \mathcal{H}_{\sigma,n}} \| h - h_0\|^2_{\infty, \mx}.
\end{align*}
Hence,
\begin{align}\label{proof_E_SD_h_h0_eq2}
   \nonumber \mathbb{E}_{\mathbb{S}, \mathbb{D}} \left[ \norm{\widehat{h}_{\widehat{r}_\mathbb{S}, \mathbb{D} } - h_0 }^2_{2, \mathbb{Q}_{X_0}} \right] &\lesssim \mathbb{E}_{\mathbb{S}} \left[  \norm{\widehat{r}_{\mathbb{S}} - r}_{2,\mathbb{P}_{X_0}}^2 \right] \\ 
\nonumber & +  \E_{\mathbb{S,D}} \mathbb{E}_{(X_0, Y_0) \sim \mathbb{P}_{X_0, Y_0}}  \left[ \widehat{r}_{\mathbb{S}} (X_0) \left( \ell_\tau \big( Y_0 - \widehat{h}_{\widehat{r}_\mathbb{S}, \mathbb{D}}(X_0)  \big) - \ell_\tau \big( Y_0 - h_{\mathcal{H}_{\sigma,n}}(X_0) \big) \right) \right] \\ 
 & +  \inf_{h \in \mathcal{H}_{\sigma,n}} \| h - h_0\|^2_{\infty, \mx}.
\end{align}
Since $\norm{\widehat{r}_{\mathbb{S}}}_{\infty} \leq \Gamma$, by employing similar arguments as in the proof of Theorem \ref{reweighted_estimator_result}, we get
\begin{equation}
    \label{bound_expectation_over_S_D_of_pretained_reweighted}
    \E_{\mathbb{S,D}} \mathbb{E}_{(X_0, Y_0) \sim \mathbb{P}_{X_0, Y_0}}  \left[ \widehat{r}_{\mathbb{S}} (X_0) \left( \ell_\tau \big( Y_0 - \widehat{h}_{\widehat{r}_\mathbb{S}, \mathbb{D}}(X_0)  \big) - \ell_\tau \big( Y_0 - h_{\mathcal{H}_{\sigma,n}}(X_0) \big) \right) \right] \lesssim \dfrac{\left( \log \varphi (n) \right)^{\nu_2}}{\left( \varphi (n) \right)^{\frac{2s_2}{2s_2 + d}}}, \; \nu_2 > 3.
\end{equation}
So, (\ref{proof_E_SD_h_h0_eq2}), (\ref{proof_cor_approx_DNN}), (\ref{bound_expectation_over_S_D_of_pretained_reweighted}) and Proposition \ref{ratio_estimator_result} we have
\begin{align*}
    \mathbb{E}_{\mathbb{S}, \mathbb{D}} \left[ \norm{\widehat{h}_{\widehat{r}_\mathbb{S}, \mathbb{D} } - h_0 }^2_{2, \mathbb{Q}_{X_0}} \right] & \lesssim  \dfrac{\left( \log \varphi (m) \right)^{\nu_1}}{\left( \varphi (m) \right)^{\frac{2s_1}{2s_1 + d}}} +  \dfrac{\left( \log \varphi (n) \right)^{\nu_2}}{\left( \varphi (n) \right)^{\frac{2s_2}{2s_2 + d}}} + \dfrac{1}{\left( \varphi (n) \right)^{\frac{2s_2}{2s_2 + d}}} \\
    & \lesssim \dfrac{\left( \log \varphi (m) \right)^{\nu_1}}{\left( \varphi (m) \right)^{\frac{2 s_1}{2 s_1 + d}}} + \dfrac{\left( \log \varphi (n) \right)^{\nu_2}}{\left( \varphi (n) \right)^{\frac{2s_2}{2s_2 + d}}},
\end{align*}
for some $\nu_1, \nu_2 >3$.
 This completes the proof.
\qed

\subsection{Proof of Proposition \ref{truncated_ratio_result}}
We have,
    \begin{equation}\label{proof_Prop_truncated_ratio_result1}
     \mathbb{E}_{\mathbb{S}} \left[ \norm{\widehat{r}_{\eta, \mathbb{S}} - r}_{2,\mathbb{P}_{X_0}}^2 \right] \lesssim  \mathbb{E}_{\mathbb{S}} \left[\norm{\widehat{r}_{\eta, \mathbb{S}} - T_{\eta}r}_{2,\mathbb{P}_{X_0}}^2 \right] +  \norm{T_{\eta}r - r}_{2,\mathbb{P}_{X_0}}^2.    
    \end{equation}

Since $\norm{\widehat{r}_{\eta,\mathbb{S}}}_{\infty} \leq \eta $ and $T_{\eta}r \in \mathcal{C}^{s}(\mathcal{X}, \eta)$,
 by using the approach in the proof of Proposition \ref{ratio_estimator_result}, and by following the steps of the proof of Theorem 4.1 and Corollary 4.3 in \cite{kengne2025deep} we get (as in (\ref{max_gamma_equation})) for some constant $C>0$ independent of $\eta$, 
\begin{equation}\label{proof_Prop_truncated_ratio_result2}
    \mathbb{E}_{\mathbb{S}} \left[\norm{\widehat{r}_{\eta, \mathbb{S}} - T_{\eta}r}_{2,\mathbb{P}_{X_0}}^2 \right] \leq C \eta \dfrac{\left( \log \varphi (m) \right)^{\nu_1}}{\left( \varphi (m) \right)^{\frac{2 s}{2 s + d}}} \lesssim \eta \dfrac{\left( \log \varphi (m) \right)^{\nu_1}}{\left( \varphi (m) \right)^{\frac{2 s}{2 s + d}}}, 
\end{equation}
for all $ \nu >3$.
Furthermore, we have as in the proof of Theorem \ref{reweighted_estimator_truncated_ratio_result} above,  
\begin{align*}
\|T_\eta r - r\|^2_{2 ,\mathbb{P}_{X_0}} &\leq \|r \ind(r > \eta)\|^2_{2 ,\mathbb{P}_{X_0}} = \mathbb{E}_{X_0 \sim \mathbb{P}_{X_0}} \left[ r^2(X_0) \ind\big(r(X_0) \big) > \eta  \right] \\
&\leq \mathbb{E}_{X_0 \sim \mathbb{P}_{X_0}} \left[ r^2(X_0) \frac{r^\delta(X_0)}{\eta^\delta} \right]   \leq \frac{U_{\mu}}{\eta^\delta}, \text{ with }  \mu = 2+\delta \text{ for some } \delta.
\end{align*}
In addition to (\ref{proof_Prop_truncated_ratio_result1}) and (\ref{proof_Prop_truncated_ratio_result2}), we get with $\eta \asymp \Big( \varphi(m) \Big)^{\frac{2 s}{(1 + \delta)(2 s +d)}} $,
\[
 \mathbb{E}_{\mathbb{S}} \left[ \norm{\widehat{r}_{\eta, \mathbb{S}} - r}_{2,\mathbb{P}_{X_0}}^2 \right] \lesssim  \eta \dfrac{\left( \log \varphi (m) \right)^{\nu}}{\left( \varphi (m) \right)^{\frac{2 s}{2 s + d}}}+\frac{U_\mu}{\eta^\delta} \lesssim  \dfrac{\left( \log \varphi (m) \right)^{\nu}}{\left( \varphi (m) \right)^{\frac{2 s}{2 s + d} \times\frac{\delta}{1+ \delta}}},
\]
for all $\nu >3$. Hence, the proposition holds.
\qed

\subsection{Proof of Theorem \ref{pretraining_reweighted_truncated_ratio_estimator_result}}
By going along similar lines as in (\ref{proof_E_SD_h_h0_eq2}), we get,
\begin{align}\label{proof_theo_truncature_L2_bound}
  \nonumber  \mathbb{E}_{\mathbb{S}, \mathbb{D}} \left[ \norm{\widehat{h}_{\widehat{r}_{\eta, \mathbb{S}}, \mathbb{D} } - h_0 }^2_{2, \mathbb{Q}_{X_0}} \right] &\lesssim \mathbb{E}_{\mathbb{S}} \left[  \norm{\widehat{r}_{\eta, \mathbb{S}} - r}_{2,\mathbb{P}_{X_0}}^2 \right] \\ 
 \nonumber & +  \E_{\mathbb{S,D}} \mathbb{E}_{(X_0, Y_0) \sim \mathbb{P}_{X_0, Y_0}}  \left[ \widehat{r}_{\eta, \mathbb{S}} (X_0) \left( \ell_\tau \big( Y_0 - \widehat{h}_{\widehat{r}_{\eta, \mathbb{S}}, \mathbb{D}}(X_0)  \big) - \ell_\tau \big( Y_0 - h_{\mathcal{H}_{\sigma,n}}(X_0) \big) \right) \right] \\ 
 & +  \inf_{h \in \mathcal{H}_{\sigma,n}} \| h - h_0\|^2_{\infty, \mx}.
\end{align}
It holds from Proposition \ref{truncated_ratio_result} that,
\begin{equation}
\label{proof_theo_truncature_r_eta_bound}
    \mathbb{E}_{\mathbb{S}} \left[  \norm{\widehat{r}_{\eta, \mathbb{S}} - r}_{2,\mathbb{P}_{X_0}}^2 \right] \lesssim \dfrac{\left( \log \varphi (m) \right)^{\nu_1}}{\left( \varphi (m) \right)^{\frac{2 \alpha}{2 \alpha + d} \times\frac{\delta}{1+ \delta}}},
\end{equation}
for all $\nu_1 >3$.
Since $\norm{\widehat{r}_{\eta, \mathbb{S}}}_{\infty} \leq \eta $, from similar arguments as in (\ref{max_gamma_equation}) and (\ref{proof_theo_rewei_delta_re_risk}), one can find that
\begin{equation}
\label{proof_theo_truncature_r_eta_statistical_bound}
     \E_{\mathbb{S,D}} \mathbb{E}_{(X_0, Y_0) \sim \mathbb{P}_{X_0, Y_0}}  \left[ \widehat{r}_{\eta, \mathbb{S}} (X_0) \left( \ell_\tau \big( Y_0 - \widehat{h}_{\widehat{r}_{\eta, \mathbb{S}}, \mathbb{D}}(X_0)  \big) - \ell_\tau \big( Y_0 - h_{\mathcal{H}_{\sigma,n}}(X_0) \big) \right) \right] \lesssim \eta \dfrac{\Big(\log \varphi(n) \Big)^{\nu_2}}{\Big( \varphi (n) \Big)^{\frac{2s}{2s + d}}},
\end{equation}
for all $\nu_2 >3$.
Hence, with $\eta \asymp \Big( \varphi(n) \Big)^{\frac{2 s}{(1 + \delta)(2 s +d)}} $, the result of the theorem follows from (\ref{proof_theo_truncature_L2_bound}), (\ref{proof_theo_truncature_r_eta_bound}), (\ref{proof_theo_truncature_r_eta_statistical_bound}) and (\ref{proof_cor_approx_DNN}).
\qed

\end{document}